\documentclass[5pt]{article}

\usepackage{graphicx}
\usepackage{multicol}
\usepackage[utf8]{inputenc}
\usepackage{siunitx} 
\usepackage[toc,page]{appendix}
\usepackage{amsmath,amssymb}
\usepackage{color}
\definecolor{Gray}{gray}{.25}
\usepackage{cite}
\usepackage{multirow}
\usepackage{nameref,hyperref}
\usepackage[right]{lineno}
\usepackage{microtype}
\DisableLigatures[f]{encoding = *, family = * }
\usepackage{tikz,graphics,float,epsf,caption,subcaption}
\usepackage[aboveskip=1pt,labelfont=bf,labelsep=period,singlelinecheck=off]{caption}

\usepackage{amssymb}
\usepackage{pifont}
\newcommand{\cmark}{\ding{51}}%
\newcommand{\xmark}{\ding{55}}%

\usepackage{changepage}
\usepackage{lastpage,fancyhdr}
\usepackage{epstopdf}
\usepackage{sidecap}
\usepackage{wrapfig}
\usepackage[pscoord]{eso-pic}
\usepackage[fulladjust]{marginnote}

\usepackage[top=0.85in,left=1in,footskip=0.75in,marginparwidth=1in]{geometry}
\def\true{true}
\def\false{false}
\def\linum{false}
\def\multcol{true}

\ifx\true\false
\raggedright
\makeatletter
\renewcommand{\@biblabel}[1]{\quad#1.}
\makeatother
\fancyheadoffset[L]{2.25in}
\fancyfootoffset[L]{2.25in}
\reversemarginpar
\fi
\begin{document}

\begin{flushleft}
\center{\huge
\textbf\newline{Psychophysical evaluation of individual low-level feature influences on visual attention}
}
\newline
\\
David Berga\textsuperscript{1,*},
Xos\'e R. Fdez-Vidal\textsuperscript{2},
Xavier Otazu\textsuperscript{1},
V\'ictor Lebor\'an\textsuperscript{2},
Xos\'e M. Pardo\textsuperscript{2}
\\
\bigskip
\small
\bf{1} Computer Vision Center, Universitat Aut\`onoma de Barcelona, Spain
\\
\bf{2} Centro de Investigaci\'on en Tecnolox\'ias da Informaci\'on, Universidade Santiago de Compostela, Spain
\\
\bigskip
*corresponding author: David Berga, dberga@cvc.uab.es

\end{flushleft}

\section*{Abstract} \label{abstract}

In this study we provide the analysis of eye movement behavior elicited by low-level feature distinctiveness with a dataset of synthetically-generated image patterns. Design of visual stimuli was inspired by the ones used in previous psychophysical experiments, namely in free-viewing and visual searching tasks, to provide a total of 15 types of stimuli, divided according to the task and feature to be analyzed. Our interest is to analyze the influences of low-level feature contrast between a salient region and the rest of distractors, providing fixation localization characteristics and reaction time of landing inside the salient region. Eye-tracking data was collected from 34 participants during the viewing of a 230 images dataset. Results show that saliency is predominantly and distinctively influenced by: 1. feature type, 2. feature contrast, 3. temporality of fixations, 4. task difficulty and 5. center bias. This experimentation proposes a new psychophysical basis for saliency model evaluation using synthetic images.

%
%
%
%
%

\vspace{2mm}
\begin{center} 
\textbf{Keywords:} visual attention, psychophysics, saliency, task, context, contrast, center bias, low-level, synthetic, dataset
\end{center}

\ifx\linum\true
\linenumbers
\else
\nolinenumbers
\fi

\ifx\multcol\true
\begin{multicols}{2}
\fi


\section{Introduction} \label{introduction}

Visual attention is the cognitive capacity of efficiently selecting relevant visual information from a scene.  Researchers record eye movements in psychophysical experiments using eye-tracking technology as means of  identifying overt attentional cues around fixation points, \cite{Kowler2011}. Registered data of different subjects show different patterns of eye movement depending on reflexive, goal-directed or contextually-specific influences \cite{Rothkopf2016}. This suggests the existence of two types of general influences in the Human Visual System (HVS), combining both bottom-up and top-down processing \cite{Desimone1995}\cite{Lamme2000}\cite{Corbetta2002}\cite{FECTEAU2006}\cite{WhiteMunoz2011}. Bottom-up processing of low-level visual features takes place in the early stages of the HVS, namely, when the nervous system efficiently extracts the basic information of the scene and processes it in the visual cortex. When higher areas of the brain are involved is when the top-down processing occurs, by taking into account internal state of the subject (task, mental state, experiences, etc.). 

"Visual salience is the distinct subjective perceptual quality which makes some items in the world stand out from their neighbors and immediately grab our attention" \cite{Itti2007}. According to the saliency map hypothesis, the HVS processes visual information of basic features such as orientation, color and scale, combining them in order to guide our attention \cite{Treisman1980}\cite{Wolfe1989}. These visual properties can be represented as distinct feature maps, being processed and then integrated in what is called saliency map (representing the pre-attentive guidance as a master map), which precedes the deployment of attention, mainly processed by bottom-up neuronal mechanisms \cite{Engel1977}\cite{Findlay2003}\cite{Borji2013a}. When the HVS combines both top-down (relevance) and bottom-up (saliency) mechanisms in order to select specific target locations of eye movements (which the superior colliculus [SC] is responsible \cite{SchillerTehovnik2001}\cite{WhiteMunoz2011}), the resulting map is termed priority map \cite{Egeth1997}. 

\cite{Koch1987} outlined a common framework in which feature maps are processed in parallel, opening up a way to computationally extract and integrate these distinct features captured by the HVS for visual saliency prediction \cite{Itti1998}\cite{Bruce2005}. Given the basis of this models, a myriad of new computational models, with biological, mathematical and physical inspiration \cite{Judd2012}\cite{Borji2013c}\cite{Zhang2013}\cite{Riche2016a}, have been able to  predict, to some extent, visual fixations while observing still images \cite{Borji2013b}\cite{Bylinskii2015}\cite{Riche2013a}\cite{Borji2013d}\cite{Bruce2015}. The limits on the prediction capability of these models arise as a consequence of the evaluation from previous datasets, that do not account contextual, perceptual, temporal and task-related biases.
%
%
%



\subsection{Contextual Relevance}\label{contextual_relevance}

One of the properties that guide visual attention is the contextual relevance of the observed scene \cite{Palmer1975}\cite{Chun1998}\cite{Henderson1999}\cite{Parkhurst2003}\cite{Vo2011}\cite{Hwang2011}\cite{deGroot2016}. Semantically-relevant content or specific high-level features can generate endogenous attentional guidance. For instance, looking at a website promotes specific eye movement patterns that differ from looking at a nature scene image; different scanpath patterns can also be found in eye-tracking experiments while humans observe indoor, outdoor and synthetic images. In most datasets for saliency modeling, observers perform free-viewing tasks with real images labeled in specific scene context categories (either faces, cars...), without taking full account of the top-down priors influenced by the context of the image with respect to feature contrast \cite{Winkler2013}\cite{Borji2013b}, which could bias both feature localization and discrimination difficulty \cite{Palmer2000}\cite{Verghese2001}\cite{Wolfe2004}.
%
%
%
\subsection{Contrast Relevance}\label{contrast_relevance}
Eye movement behavior is influenced not only by content and stimulus context, but also by the human perceptual capabilities for distinct contrast adaptation and discrimination \cite{Nothdurft2000}\cite{Pashler2004}\cite{Pestilli2007}\cite{McDermott2010}\cite{Hart2013}. Other perceptually-relevant factors could also be related either to the lighting conditions used in each experiment, the starting point of view when perceiving stimuli, etc. The evaluation of relative distinctiveness between features at distinct regions of the image is needed to be done in order to analyze each image according to its spatial properties and feature specificities. This suggests that each image promotes distinct saliency.
%
%
%

\subsection{Temporal Relevance}\label{temporal_relevance} 
Eye movements have been shown to have temporal influences, varying its behavior upon viewing time or number of fixations (e.g. showing decreasing saccade amplitude, increasing fixation duration \cite{Follet2011}\cite{Antes1974} or higher inter-participant differences \cite{Tatler2005}\cite{Rothkegel2017}), suggesting the idea that saliency influences more early saccades than late viewing saccades \cite{Parkhurst2002}\cite{Tatler2005}\cite{Zhaoping2012}\cite{Zhaoping2015}. Most saliency predictions based on eye tracking data do not evaluate the temporal relevance in relation to the saliency elicited by the scene, being for most cases, evaluated spatially across all fixations.

\subsection{Task Relevance}\label{task_relevance}
Alfred Yarbus' seminal work revealed differences in eye movement patterns \cite{Yarbus1967} caused by certain top-down influences such as previous experience, motivation and other endogenous factors. Distinctive studies have also concluded that task priors are decisive in that respect 
\cite{buswell1935people}\cite{Navalpakkam2005}\cite{Tatler2006}\cite{Castelhano2009}
\cite{Greene2012}\cite{Borji2014}. Goal-directed tasks proved to be able to condition eye movement behavior enhancing visual attention processing \cite{Posner1980}\cite{Jonides1981}\cite{Huk2000}\cite{Kurtz2017}. That might suggest that visual search tasks could minimize such eye-movement patterns produced by endogenous top-down mechanisms \cite{Horowitz1998}\cite{Wolfe1998}, by increasing induced attention towards salient targets (combining both saliency and relevance to influence eye guidance towards these regions). Thus, for all tasks, there is an induced top-down processing that tune overall visual priority when recording eye-movements \cite{Henderson1999}  \cite{Jonides1981}\cite{Desimone1998}, given both exogenous and endogenous influences. Such design puts forward that there could be a better computational estimation of saliency if such task-related influences were focused uniquely on the regions that pop-out on the scene.

\subsection{Center bias}\label{biases_relevance} 
Eye movement datasets built for the assessment of saliency models tend to be center biased, not only because of scene framing (photographies tend to focus the salient region in the center of view) but also because of the specific task and stimuli, whereof top-down modulatory constraints are enough to prevent attentional shift, giving a trend to promote center biases \cite{Borji2016}\cite{Mannan1996}\cite{tatler2008}\cite{Clarke2014}\cite{Rothkegel2017}, not only in oculomotor terms but also in tendencies in experimentation of eye movement behavior. As aforementioned, bottom-up and top-down processing of the stimuli will depend on the feature characteristics from the scene. If these are simpler, the contextual influence will be lower, making the indicators of saliency easier to analyze \cite{Tatler2007}\cite{tatler2008}. 
There will be an endogenous top-down attentional modulation whether the stimulus is cued or uncued. For concrete salient stimuli, facilitating attentional guidance by inducing specific endogenous cues could enable the selection of specific regions of interest in order to prevent the aforementioned factors that generate these center biases.

\subsection*{Objectives}

Acknowledging the aforementioned problems on capturing bottom-up visual saliency, we have decided to create a dataset with synthetic images, lacking the presence of high-level features, promoting saliency uniquely elicited from low-level features (providing as well a synthetic image generator code). An alternative evaluation of saliency proposed, by measuring eye movements upon low-level feature distinctiveness and their temporality. Fixations and saccades will be evaluated individually with the corresponding stimuli on free-viewing and visual search tasks, with different feature types and distinct target-distractor feature contrasts. 

In order to vary the level of saliency of specific features in a scene, a parametrization of the distinctiveness between a specific item and a set of distractors or its surrounding background is needed. By parameterizing feature contrast, it is possible to analyze feature search efficiency, its accordance with the Weber Law, and the effects in which search asymmetries apply. Using synthetic images in eye-tracking experiments, the complexity of the image features is reduced by minimizing any top-down contextually-related effect, putting forward an easier and more accurate evaluation of eye movement behavior. By modeling stimulus areas of interest for selected pop-out targets, it it possible to test participants performance on landing inside salient regions and their eye movement patterns (in the extent of fixation duration and saccade amplitude) for distinct feature contrasts, and their temporal evolution. This will allow us to observe whether low-level features influence visual attention in a distinct manner. 


Previous experiments that perform psychophysical tests evaluating human visual performance on distinct low-level features (iLab USC \cite{Itti2000}, UCL \cite{Zhaoping2007}, VAL Harvard \cite{Wolfe2010b} and ADA KCL \cite{Spratling2012}) show that the distinctiveness between a specific region and the rest of distinct regions of an image progressively increases the level of saliency in relation to feature contrast. However, the presence of much less relevant features distorts the overall distinctiveness of a specific region with respect to the rest, thus, affecting to the bottom-up visual guidance towards the salient region. With the aforementioned datasets (\hyperref[table:datasets]{Table \ref*{table:datasets}}), feature contrast and stimulus conditions has been parametrized with search tasks (using the button trigger for calculating search reaction times) but no eye tracking experimentation has been done. 


\begin{table}[H]
\caption{Characteristics of datasets with synthetic/psychophysical pattern images}
\begin{tabular}{ |c|c|c|c|c| } 
\hline
Dataset & Task & \# SS & \# PP & PM \\ 
\hline
MIT* & FV & 3 & 15 & \xmark  \\ 
CAT2000* & FV & 100 & 18 & \xmark\\ 
iLab USC & - & \textasciitilde 540 & - & \cmark \\ 
UCL & VS \& SG & 2784 & 5 & \cmark\\
VAL Harvard & VS & 4000 & 30 & \cmark \\ 
ADA KCL & - & \textasciitilde 430 & - & \cmark \\ 
SID4VAM (Ours)* & FV \& VS & 230 & 34 & \cmark \\ 
\hline
\end{tabular}
\small SS: synthetic stimuli, PP: participants, PM: Parametrization, FV: Free-Viewing, VS: Visual Search, SG: visual segmentation
*: Fixation data is available online 
\label{table:datasets}
\end{table}

For few eye movement datasets that contain synthetic images (MIT\cite{Judd2009} and CAT2000 \cite{CAT2000}), no parametrization of feature type or contrast was done. Contrary to other saliency datasets \cite{mit-saliency-benchmark}, in this study it is possible to evaluate each of these factors individually and exclusively eye movement data is being used for calculating search performance for better accuracy. We will test the following hypotheses:
\begin{enumerate}
%
\item Performance on salient region localization could show differences upon varying the type of features present in our stimuli.
\item If feature contrast is the main factor that contributes to saliency, performance on localization of salient regions should correlate with feature contrast, specially for stimuli that require a serial ‘binding’ step.
\item Acknowledging that saliency is usually evaluated across all fixations on eye tracking experimentation, if a temporal bias exists and is increasing, it is highly possible that the first fixations show higher saliency index than the late ones.
\item If performance on salient region localization with free viewing tasks is lower for stimuli with higher contrast compared to visual search tasks, it will mean that fixations on free viewing tasks are highly guided by endogenous attention.
\item Previous datasets used for saliency prediction do not show how their center biases affects saliency. We will show how eye movement patterns influence the center bias for this dataset and if the bias increases or decreases across viewing time and feature contrast.

%

\end{enumerate}

Our objective is to allow computer vision researchers to reproduce these influences when modeling eye-movement prediction algorithms. Here we present a dataset in which we evaluate through free-viewing tasks the influence of the features that affect the spatial properties of an image (from the perception of Corners, Segments, Contours and Grouping) and how the relative distinctiveness from a search target is more salient with respect to a set of distractors that differ from specific low-level features (color, orientation, size...). Analyzing low-level features individually would allow us to see which features generate more agreement on saliency measures and are localized faster, in that manner, to allow their modeling according to their distinct neuronal mechanisms. This study can be used for a more plausible and specific saliency modeling given the presented eye-movement patterns, also extrapolable to the analysis of the interactions between these features or to study specific cases of high-level features in future studies.

\section{Materials and Methods}

\subsection{Participants} \label{participants}

Thirty four subjects (11 female and 23 male) with normal or corrected-to-normal vision took part in this experiment. Most participants were postdoc scholars and PhD students (aged 21–47 years) from non-related fields of study. No economic compensation for the experiments was given. Participants were allowed to wait until they were comfortable with the eye tracking experimental setup in case they had any kind of visual discomfort in between sessions, and they were allowed to adjust the chair while laying on the chin-rest before the experiment. Participants had to sign a consent form allowing the anonymous usage of the data captured during the experiment. 

\subsection{Apparatus} \label{apparatus}

The set of stimulus was presented on a LCD monitor (Samsung SyncMaster HMAQ935729) of screen size $340$x\SI{270}{\milli\meter}, a resolution of $1280$x\SI{1080}{px} and a refresh rate of \SI{60}{\Hz}. A color calibrator was used (Xrite i1 Display Pro) in order to set a specific luminance for the monitor of \SI{160}{\candela\per\metre\squared}, achieving the CIE Illuminant D65 according to the ISO 3664:2000 standard condition (and recommended by Adobe RGB 1998 CIE and ITU-R BT.500-11) with the whitepoint at x=$0.313$, y=$0.329$ and a gamma value of $2.2$. The light conditions of the room were set using non-direct adjustable light, measured at \SI{30}{\lux} using a luxmeter (TES1332).

\hfill

We have used a SMI RED binocular eye tracker with a tracking resolution of \textless \SI{0.1}{\deg}, gaze position accuracy of \textless \SI{0.5}{\deg} and a sampling rate of \SI{50}{\Hz}, set at a distance of \SI{600}{\milli\meter} towards the chin-rest (about $40$ pixels per degree of visual angle) and vertically equidistant with respect to the monitor, forming a slope of \SI{19}{\deg} from the horizontal axis. The monitor's screen was at a vertical distance of \SI{195}{\milli\meter} from the table and the observer's point of view was adjusted to be centered towards the screen. Fixation and saccade detection was based on SMI iView X Event Detector software, capturing fixations at a minimum duration time of \SI{80}{\milli\second} and maximum dispersion threshold value of \SI{2}{\deg} and saccades at a peak velocity threshold of \SI{75}{\deg /\second} \cite{Salvucci2000}\cite[p. 243-247]{iviewx2009}.

\subsection{Procedure} \label{procedure}

The experiment was divided in one training and two full sessions. During the training session each participant performed a visual search task with 4 types of stimulus with feature and conjunctive search, combined with present and/or absent search targets, hence to ensure their good performance in the next sessions. The first session had a duration of about 20 minutes and was divided in 8 blocks, each one corresponding to a free-viewing or visual search task. The second session had a duration of about 25 minutes and was divided in 10 blocks, similar to the first session. Each task in a block correspond to a distinct stimulus type (shown in \hyperref[stimuli]{Section \ref*{stimuli}}) that was presented in a random order. Stimulus order was also randomized across blocks (to avoid any stimulus-related priming \cite{Kristjnsson2008}), and the location of target distractors was distinct for each case in order to prevent oculomotor biases.
%
%
%

Participants performed two types of tasks: free-viewing and visual search (\hyperref[fig:figproc]{Figure \ref*{fig:figproc}}). During free-viewing tasks, they were instructed to freely look at the stimuli during \SI{5000}{\milli\second}. For the visual search task, they were instructed to look for a specific target  previously shown in an  instruction slide. In case they could find the target, they had to steer their gaze towards it during a dwell time of \SI{1000}{\milli\second} (the area of interest was based on the target area with an horizontal and vertical spacing of \SI{1}{\deg}). In case they could not identify the target, they were instructed to press a specific key. Considering that context was distinct for each block (replicating stimulus characteristics from previous studies), we decided to do a template target search task instead of an odd-one-out type of task \cite{Bacon1994}\cite{Theeuwes2004}. Participants had unlimited time for the visual search tasks, in this case for reporting target identification or absence. Transitions between stimuli had a duration of \SI{2000}{\milli\second} (blank transition without the presence of an onset cue) with a luminance equal to the stimuli in order to preserve participant's luminance and chromatic adaptation. 

\begin{figure}[H]
 \centering
\begin{subfigure}[b]{0.45\linewidth}
\caption*{\centering \textbf{Free-viewing}}
\includegraphics[width=1\linewidth,height=4cm]{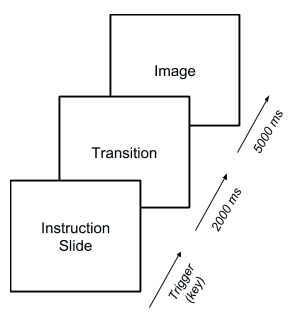} 
\end{subfigure}
\begin{subfigure}[b]{0.45\linewidth}
\caption*{\centering \textbf{Visual Search}}
\includegraphics[width=1\linewidth,height=4cm]{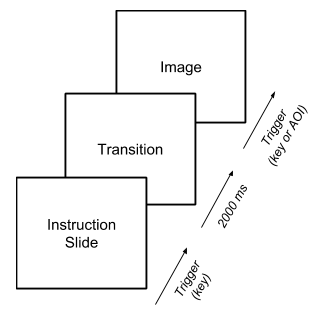} 
\end{subfigure}
\caption{Procedure for the presentation of the stimuli for each task type. } 
\label{fig:figproc}
\end{figure}

Mean pupil size was recorded to be \SI{2.98}{\milli\meter} diameter for all samples and there was a standard error of \SI{0.18}{\milli\meter} mm between stimulus type, being almost constant throughout the experiment with no significant stimulus-related luminance imbalance.

A 12 point calibration procedure was performed before each session, in which participants were instructed to gaze a red dot moving along different directions. The calibration showed mean deviations for all sessions of $\sigma_x$=$0.57$, $\sigma_y$=\SI{0.75}{\deg} for the left eye and $\sigma_x$=$0.56$, $\sigma_y$=\SI{0.82}{\deg} for the right eye. Deviations for each participant's fixation and saccade data were computed using the data of the participant's eye that presented minimum deviation  from  calibration points in each session. We did a pilot experiment with 4 participants in order to correctly design the visual search procedure, thus, to test the final experimental design for trigger timing and the difficulty of the tasks. That allowed us also to correctly parametrize the variables corresponding to target-distractor contrast where the target was too hard to identify, this parametrization will be shown in the next section.

\subsection{Stimuli} \label{stimuli}

A total of 33 types of stimuli were generated, corresponding to 15 distinct feature evaluations (5 of them using free-viewing tasks and 10 for visual search tasks) at distinct conditions. During free-viewing experiments, we evaluated how spatial properties influence saliency, namely, the capabilities of humans for detecting corners, segmenting and detecting contours as well as localizing groups of objects according to their similarity and spatial distribution (\hyperref[table:stimuli_fv]{Table \ref*{table:stimuli_fv}}). This will give some insight of how rapidly humans reflexively perceive and bind spatial properties from the features of an image. In visual search tasks, we evaluated the speed in detecting specific features and the amount of saliency produced by target-distractor feature contrast characteristics. In that aspect, stimulus were generated with features that pop-out based on their dissimilarities in orientation, color and size. Besides, we also analyzed influences of the guidance prompt from the amount of distractors on the scene, their configuration as well as the influence of background lightness, color and roughness (\hyperref[table:stimuli_vs]{Table \ref*{table:stimuli_vs}}).

Stimulus design was was inspired by Spratling's experiments \cite{Spratling2012}, by generating synthetic images similar to the ones from Li and May's psychophysical experiments \cite{Zhaoping2007}. Most stimuli items had a size of \SI{1.5}{\deg}, occupying a region of \SI{2.5}{\deg} including the spacing between distractors. In that manner, stimulus had an available grid of $10\times 13$ distractors. Distractors were black $(lsY=0,0,0)$, and background was plane white $(lsY=0.6548,0.0175,1)$. We used Spratling's code and we adapted it in order to also use any distractor shape, displacement and chromatic parameterization, downloadable at \url{https://github.com/dberga/sig4vam}. 

\hfill
\begin{equation} \label{eq:psi1}
\Psi(x) = \{\frac{x-1}{N-1}  \quad | \quad   x\in N  \},
\end{equation}
\begin{equation} \label{eq:psi2}
\Psi(x,v) = \{v \cdot \Psi(x)\},
\end{equation}
\begin{equation} \label{eq:psi3}
\Psi(x,min,max) = \{\Psi(x,min) \cup \Psi(x,max) \quad | \text{x is odd}  \}.
\end{equation}

The parameters of the generated stimuli  were set according to $"N=7"$ contrast values (ranging from 0 to 1), using the Weber's law uniform fraction in order to set discrete target-distractor evaluation $"x={1...N}"$ for the psychometric function $\Psi$. For each stimulus type on our experimentation, parameters are set according to specific values of $\Psi(x)$. We have the expression in \hyperref[eq:psi1]{Equation \ref*{eq:psi1}} and 7 contrast values in order to have extreme contrast values (no contrast and maximum contrast) with "$\Psi(1)=0$" and "$\Psi(7)=1$" as well as a middle value with "$\Psi(4)=0.5$", making the difference between the second lower contrast and the second maximum contrast at the same distance from the extreme contrast values $|\Psi(2)-\Psi(1)|=|\Psi(7)-\Psi(6)|$. In that manner we provide a psychometric function with a constant slope (Weber's uniform fraction). Absolute values of contrast can be adjusted to fit a specific value of $"v"$ as $\Psi(x,v)$ (\hyperref[eq:psi2]{Equation \ref*{eq:psi2}}). For cases that we had higher and lower contrasts with respect the target and overall distractors we adjusted the values for maximum and minimum range of the psychometric function $\Psi(x,min,max)$ as the union of odd values for both sets of $\Psi(x,min)$ and $\Psi(x,max)$ in order to acquire the same set of contrast values (\hyperref[eq:psi3]{Equation \ref*{eq:psi3}}).

Acknowledging that each stimulus was distributed according to different contrasts depending on the evaluation parameter, each stimulus was categorized as easy and hard depending on the assigned contrast (half of them as easy for higher contrasts, and half of them as hard for the case of lower contrasts, with a specific case with minimum or no contrast). One of our interests was to evaluate how low-level features  modify the spatial layout between the features on a scene, therefore its spatial properties, affecting visual saliency (in this case with free-viewing experimentation). In order to accurately evaluate low-level feature distinctiveness, visual search tasks were performed, having a search target with a specific low-level contrast with respect to a set of distractors. The stimulus design corresponding distinctively to each feature and task will be explained as follows.

\subsubsection{Free-viewing task stimuli} \label{stimuli_fv}

First, we wanted to evaluate the spatial relevance of certain regions of an image. For this stimuli, visual selection cannot be focused on a unique region due to the size and/or spatial organization of the elements in the image. Humans have a limited central vision, namely, they need several fixations over the whole region in order to attend to all of the relevant regions in detail. In that aspect, each of the spatial properties will guide attention towards a single or several spots depending on the analyzed feature. With this type of stimuli it is be able to see the temporal and spatial performance of perceiving boundaries due to corner sharpness, segment angle and spacing as well as preemption and grouping \cite{Rensink1995}\cite{Rensink1998}\cite{Wolfe2011}, effects induced by distractor continuity, proximity and similarity. These preattentive effects are not equally processed in the visual system in the same way as shown for parallel visual search \cite{Wolfe1992}. Task was separate for the aforementioned perceptual phenomena with respect to searching for a specific feature, stimuli described on \hyperref[stimuli_vs]{Section \ref*{stimuli_vs}}. 

\begin{table*}[htpb!]                                    
\caption{Description of the generated stimuli for the experiment using the free-viewing task. Stimulus have been divided in "Stimulus type" according to the type of feature or effect that is analyzed and "Stimulus subtypes" for the cases that there are presented distinct conditions using the same feature contrast. The total number of elements has been  selected according to the stimulus characteristics, preserving similar spatial properties to the ones presented on the literature.}                        
\centering                         
\resizebox{\linewidth}{!}{        
\begin{tabular}{|c|c|c|c|c|}        
\hline                              
\# of stimuli & Stimulus type & Stimulus subtypes & Parametrized Feature Contrast & Total \# of elements \\                                                  
\hline
7 & \hyperref[stimuli_fv1]{Corner Angle (1)} &  & Sharpness Orientation & 1 \\                                                         
\hline
\multirow{2}*{14} & \multirow{2}*{\hyperref[stimuli_fv2]{Segmentation by Angle (2)}} & Single & \multirow{2}*{Segment Orientation} & $10\times13$ (130) \\                                                                                      
 &  & Superimposed &  & $20\times 26$ (520) \\                                                                      
\hline
7 & \hyperref[stimuli_fv3]{Segmentation by Spacing (3)} &  & Bar Length and Spacing & $10\times 13$ (130) \\                                                                                   
\hline
6 & \hyperref[stimuli_fv4]{Contour Integration (4)} &  & Bar Continuity & $10\times 13$ (130) \\           
 \hline
\multirow{2}*{14} & \multirow{2}*{ \hyperref[stimuli_fv5]{Perceptual Grouping (5)}} & Similar & \multirow{2}*{Distractor Proximity} & \multirow{2}*{$\sim$40} \\           
 &  &  Dissimilar &  &  \\      
 \hline
\end{tabular}                        
}            
\label{table:stimuli_fv}                     
\end{table*}

\paragraph{Corner Angle (1)}  \label{stimuli_fv1}

Troncoso et al's psychophysical experimentation found that corner salience was higher on sharp corners than on shallow corners or edges \cite{Troncoso2005}\cite{Troncoso2009}. This effect could be explained by ON-center receptive field behavior towards corner stimuli \cite{Rodieck1965}, being sharp corners the ones that produce higher neuronal activity. Original stimuli from Troncoso's experiment was used, generating corners with a dark-to-white gradient and an upwards angle, corresponding to corner angles of $180, 135, 105, 75, 45, 30$ and \SI{15}{\degree} (shown in \hyperref[fig:stimuli_fv1]{Figure \ref*{fig:stimuli_fv1}}). The horizontal alignment of the corner was randomized in order to prevent oculomotor anticipation.

\begin{figure}[H]
 \centering
\begin{subfigure}{0.12\linewidth} 
\caption*{\centering 15º}
\includegraphics[width=1\linewidth]{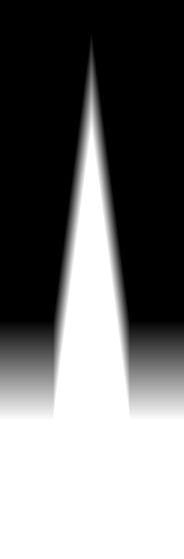}
\end{subfigure}
\begin{subfigure}{0.12\linewidth}
\caption*{\centering 30º}
\includegraphics[width=1\linewidth]{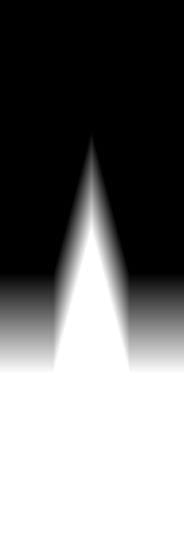}
\end{subfigure}
\begin{subfigure}{0.12\linewidth}
\caption*{\centering 45º}
\includegraphics[width=1\linewidth]{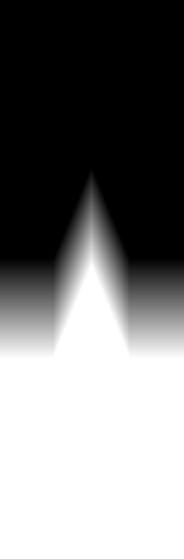} 
\end{subfigure}
\begin{subfigure}{0.12\linewidth}
\caption*{\centering 75º}
\includegraphics[width=1\linewidth]{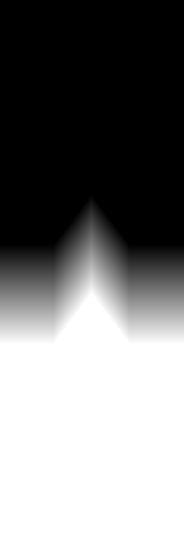} 
\end{subfigure}
 \begin{subfigure}{0.12\linewidth}
\caption*{\centering 105º}
\includegraphics[width=1\linewidth]{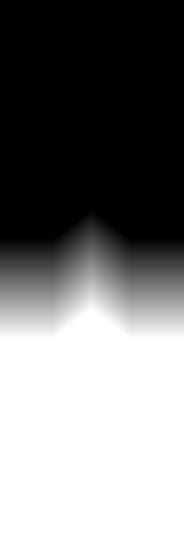}
\end{subfigure}
\begin{subfigure}{0.12\linewidth}
\caption*{\centering 135º}
\includegraphics[width=1\linewidth]{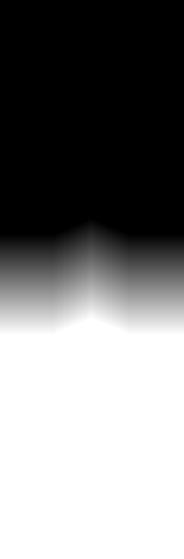}
\end{subfigure}
\begin{subfigure}{0.12\linewidth}
\caption*{\centering 180º}
\includegraphics[width=1\linewidth]{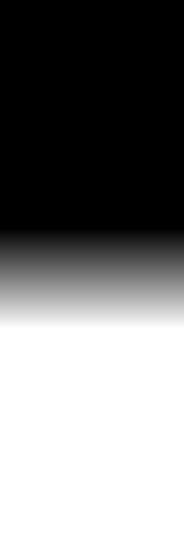}
\end{subfigure}
\caption{Examples of corner angle slopes (with the sharper at \SI{15}{\degree} and the smoother at \SI{180}{\degree}) for dark-to-bright gradient stimuli with upwards angles.}
\label{fig:stimuli_fv1}
\end{figure}

\paragraph{Visual Segmentation} \label{stimuli_fv2-3}

Distinctiveness between two homogeneous regions creates higher neural activity near region boundaries than away from them \cite{Li1999}\cite{Li2000}. In this section is described how an illusory boundary is generated by varying two segment characteristics. This effect is distinct from the concepts of edge or boundary detection (terms used as well in the image segmentation literature) or \hyperref[stimuli_fv4]{contour integration}. This phenomena proves that illusory boundaries pop-out due to the perceptual breakdown of homogeneity. Here it is studied the influence of angle contrast between these two segments (creating a salient boundary dependent on the segments angle) with an homogeneous single set of bars as well as with superimposed bars. Here is also analyzed the influence of bar spacing and length on detecting the illusory boundary between these two segments.

\subparagraph{Segmentation by Angle (2)} \label{stimuli_fv2}

Visual angle contrast between two segments can induce edge detection and therefore visual saliency towards that illusory edge \cite{Nothdurft1991}\cite{Wolfson1995}\cite{Zhaoping2007}. The resulting saliency would increase with respect to angle contrast from the two segments on the region that separates them \cite{Spratling2012}. It is a distinctive effect from orientation feature detection upon a set of distractors, that is described on \hyperref[stimuli_vs7]{Orientation Contrast (12)}, \hyperref[stimuli_vs8]{Distractor Heterogeneity (13)}, \hyperref[stimuli_vs9]{Distractor Linearity (14)} and \hyperref[stimuli_vs10]{Distractor Categorization (15)}.

\begin{equation} \label{eq:phi}
\Phi(v,a) = \{|\arcsin(\Psi(1...N,v))+a|\},
\end{equation}
\begin{equation} \label{eq:cphi}
\Delta\Phi(v,a,b) = min\{|b-\Phi(v,a)|\quad,\quad 180-|b-\Phi(v,a)|\},
\end{equation}

The psychometric values for determining angle values are defined as $\Phi(v,a)$. Here "$v$" is the incremental factor for adjusting the maximum angle for our set $\Psi(x,v)$ and $"a"$ is the starting angle value for our bar orientation (\hyperref[eq:phi]{Equation \ref*{eq:phi}}). The angle contrast between a specific angle "b" and our set of angles $\Phi(v,a)$ can be computed with $\Delta\Phi(v,a,b)$, considering that our bar orientations have upwards and downwards contrast for its comparison (due to its symmetry), contrast is calculated as the minimum from the differences from two quadrants in which these bars can be oriented (\hyperref[eq:cphi]{Equation \ref*{eq:cphi}}).

Stimuli was based on Spratling's visual segmentation, using 2 sets of bars (shown in \hyperref[fig:stimuli_fv2-3]{Figure \ref*{fig:stimuli_fv2-3}\textbf{(a,b)}}) oriented respectively using angles $\Phi(1,0)$ and $"b=90"$, forming a relative contrast of $\Delta\Phi(1,0,90)$. For the case of superimposed bars, we have created a composite of the same bars adding a bar tilted at \SI{45}{\degree} with respect to each segment. Here are accounted the contrasts between the new superimposed bars and the original segment $\Delta\Phi(1,45,90)$. The location of the vertical segment was randomized on the horizontal axis for each stimulus.

\begin{figure}[H]
 \centering
\begin{subfigure}{0.28\linewidth}
\fbox{\includegraphics[width=1\linewidth]{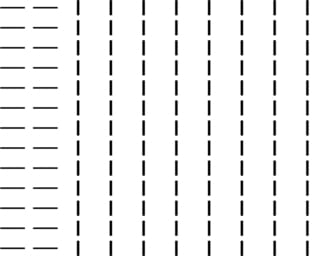}}
\caption{\centering}
\end{subfigure}
\quad
\begin{subfigure}{0.28\linewidth}
\fbox{\includegraphics[width=1\linewidth]{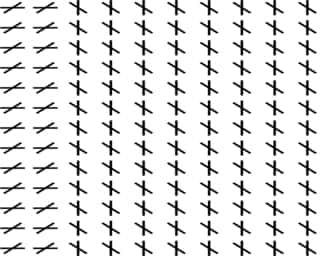}}
\caption{\centering}
\end{subfigure}
\quad
\begin{subfigure}{0.28\linewidth}
\fbox{\includegraphics[width=1\linewidth]{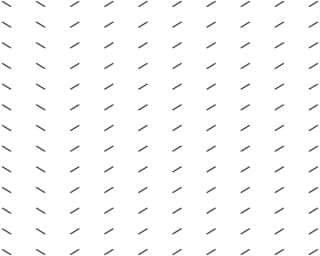}}
\caption{\centering}
\end{subfigure}
\caption{Examples for visual segmentation stimuli. \textbf{(a)} Corresponds to the segmentation by angle with a single segment and \textbf{(b)} to superimposed segments (with both cases with an orientation contrast of $\Delta\Phi$=\SI{90}{\degree}. In \textbf{(c)} segmentation is done distinctively (by changing bar length), using bars oriented at $45$ and \SI{-45}{\degree} with a bar length of 1 deg and a segment spacing of \SI{2.5}{\deg}.} 
\label{fig:stimuli_fv2-3}
\end{figure}

\subparagraph{Segmentation by Distance (3)} \label{stimuli_fv3}

Texture discrimination was shown to vary according to the spacing and length of the texture elements \cite{Nothdurft1985}\cite{Spratling2012}, making it harder as element spacing increases (as when segment elements decrease in length or size). Visual segments were modeled using 2 sets of bars, oriented respectively at \SI{45}{\degree} and \SI{-45}{\degree} (a relative angle contrast of \SI{90}{\degree}). Here we question how bar length is able to generate a specified distance at the center of the illusory segment. Segment spacing was calculated as the euclidean distance from the end of the first segment bar to the beginning of the second segment bar, with values of $0$ to \SI{2.5}{\deg} (shown in \hyperref[fig:stimuli_fv2-3]{Figure \ref*{fig:stimuli_fv2-3}\textbf{(c)}}), corresponding respectively to a bar length of $1$ to \SI{3.6}{\deg} deg in the horizontal axis.
%
%

\paragraph{Perceptual organization} \label{stimuli_fv4-5}

Perceptual organization has been previously investigated and promoted by Gestalt principles, guided by proximity, similarity, continuity, and closure properties of objects \cite{koffka1935principles}\cite{wertheimer_laws_1938}\cite{Field1993}\cite{Caputo1997}\cite{Rubin2001}. Here are described two effects related to perceptual organization, parametrized upon the aforementioned principles.

\subparagraph{Contour Integration (4)} \label{stimuli_fv4}

Continuity within set of features in a scene is able to generate the perception of a contour \cite{Hess2003}, considering that a larger set of collinear bars facilitate its detection. Accounting for saliency being influenced by contour integration \cite{Li2002}\cite{Dakin2009}\cite{Spratling2012}, a set of stimuli was created with a grid of randomly oriented and equidistant bars and a collinear contour (\hyperref[fig:stimuli_fv4-5]{Figure \ref*{fig:stimuli_fv4-5}\textbf{(a)}}). Contours were generated with a length of $3, 5, 7, 8, 9$ and $10$ collinear bars, corresponding to $7.5$ to \SI{25}{\deg}.
%
%

\subparagraph{Perceptual Grouping (5)} \label{stimuli_fv5}

Adding up to the basis of the previous section, we have also studied the relation between perceptual grouping principles and visual attention. According to the literature, the spatial layout can facilitate or prevent contextual cueing \cite{Bhatt2007}\cite{Conci2013}, in particular, the lower the proximity between a number of randomly distributed objects and a group, the higher the saliency on the grouping region \cite{Nothdurft1985}\cite{BenAv1992}\cite{BenAv1995}\cite{Rideaux2016}. Given that, here the analysis is on the influence of proximity and similarity among objects in a specific spatial organization. To do so, there were generated a set of shapes, randomly distributed and located at specific distances to a group, with similar and dissimilar shapes \hyperref[fig:stimuli_fv4-5]{Figure \ref*{fig:stimuli_fv4-5}\textbf{(b,c)}}. The proximity parameter was the euclidean distance between the group centroid and the rest of distractors, forming a wider gap between the distractors and the group as we increase distance, parametrized as $\Psi(1...N,2.5,7.5)$. Stimulus shapes were set to be symmetric in order to prevent orientation-variant guidance, with squares as the main shape for both group and distractors in the case of similar object condition, whereas in the case of dissimilar condition were selected triangle shapes for the distractors and squares for the group.

\begin{figure}[H]
 \centering
\begin{subfigure}{0.28\linewidth}
\fbox{\includegraphics[width=1\linewidth]{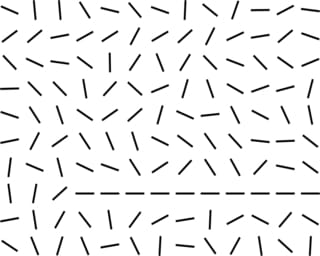}}
\caption{\centering}
\label{fig:subim1}
\end{subfigure}
\quad
\begin{subfigure}{0.28\linewidth}
\fbox{\includegraphics[width=1\linewidth]{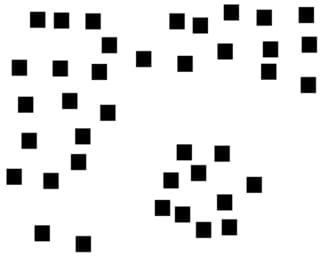}}
\caption{\centering}
\end{subfigure}
\quad
\begin{subfigure}{0.28\linewidth}
\fbox{\includegraphics[width=1\linewidth]{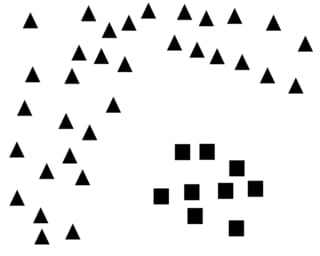}}
\caption{\centering}
\end{subfigure}
\caption{Examples of distinct perceptual organization effects, eliciting Contour Integration \textbf{(a)} formed by 10 collinear bars (corresponding to \SI{25}{\deg}) and Perceptual Grouping for similar \textbf{(b)} and dissimilar shapes \textbf{(c)} with respect a group set at a distance of \SI{7.5}{\deg} from the rest of distractors}
\label{fig:stimuli_fv4-5}
\end{figure}

\subsubsection{Visual Search task stimuli} \label{stimuli_vs}

Visual search tasks were performed with another set of stimuli. In this case, a unique target item with a specific size was used. In that manner, it is possible to change either the amount of distractors, their spatial configuration, the target-distractor feature contrast and background properties of the stimulus. Most target elements overall occupied a small area of interest in order to able to be preserve same fovea-dependent capabilities for each type of stimuli used with this type of task. Hence, using this experimentation we can observe which features pop-out faster and more often (in parallel or "effortlessly"). As small regions away from central vision cannot be detected as in the case of bigger regions shown on \hyperref[stimuli_fv]{Section \ref*{stimuli_fv}}, the guidance towards the salient target (distinctive from the rest of distractors) minimizes other type of guidance promoted from endogenous factors.

\begin{table*}[!htpb]                       
\caption{Description of the generated stimuli for the experiment using the Visual Search task. The total number of elements has been selected according to the amount of distractors presented on the scene acknowledging that one of the elements is presented to be the search target.}
\centering                         
\resizebox{\linewidth}{!}{                                                                                               
\begin{tabular}{|c|c|c|c|c|}                                                                                               
\hline                                                                                                                     
\# of stimuli & Stimulus type & Stimulus subtypes & Parametrized Feature Contrast & Total \# of elements \\
\hline                                                                                                                     
\multirow{4}*{28} & \multirow{4}*{\hyperref[stimuli_vs1]{Feature Search (6)}} & Feature & \multirow{4}*{Distractor number} & \multirow{4}*{3 to 35}\\                                                                                                                        
 &  & Conjunctive &  &  \\                                                                                                  
 &  & Feature-absent &  &  \\                                                                                              
 &  & Conjunctive-absent &  &  \\                                                                                          
\hline                                                                                                                     
\multirow{2}*{14} & \hyperref[stimuli_vs2]{Search Asymmetries (7)} & Bar presence & \multirow{2}*{Scale and Distractor number} & \multirow{2}*{35 to 520} \\                                                                                          
 &  & Bar absence &  &  \\                                                                                                  
\hline                                                                                                                     
\multirow{2}*{14} & \multirow{2}*{\hyperref[stimuli_vs3]{Noise/Roughness (8)}} & Higher deviation & \multirow{2}*{Surface Roughness} & \multirow{2}*{1} \\                                                                                                           
 &  & Lower deviation &  &  \\                                                                                             
\hline                                                                                                                     
\multirow{4}*{28} & \multirow{4}*{\hyperref[stimuli_vs4]{Color Contrast (9)}} & Red target and Unsaturated Background & \multirow{4}*{Distractor Saturation} & \multirow{4}*{34} \\                                                                                              
 &  & Red target and Oversaturated Background &  &  \\                                                                     
 &  & Red target and Unsaturated Background &  &  \\                                                                       
 &  & Blue target and Oversaturated Background &  &  \\                                                                    
\hline                                                                                                                     
\multirow{2}*{14} & \multirow{2}*{\hyperref[stimuli_vs5]{Brightness Contrast (10)}} & Light Background & \multirow{2}*{Distractor lightness} & \multirow{2}*{34} \\                                                                                                                     
 &  & Dark Background &  &  \\                                                                                             
\hline                                                                                                                     
7 & \hyperref[stimuli_vs6]{Size Contrast (11)} &  & Target Size & 34 \\                                                                                          
\hline                                                                                                                     
7 & \hyperref[stimuli_vs7]{Orientation Contrast (12)} &  & Target Orientation & 34 \\                                                                            
\hline                                                                                                                     
\multirow{3}*{21} & \multirow{3}*{\hyperref[stimuli_vs8]{Distractor Heterogeneity (13)}} & Homogeneous & \multirow{3}*{Target Orientation} & \multirow{3}*{$10\times 13$ (130)} \\                                                                                                             
 &  & Tilted-right &  &  \\                                                                                               
 &  & Flanking &  &  \\                                                                                                    
\hline                                                                                                                     
\multirow{4}*{28} & \multirow{4}*{\hyperref[stimuli_vs9]{Distractor Linearity (14)}} & Linear & \multirow{4}*{Target Orientation} & \multirow{4}*{$10\times 13$ (130)} \\                                                                                                                       
 &  & Nonlinear at 10º of slope &  &  \\                                                                                  
 &  & Nonlinear at 20º of slope &  &  \\                                                                                   
 &  & Nonlinear at 90º of slope &  &  \\                                                                                   
\hline                                                                                                                     
\multirow{3}*{21} & \multirow{3}*{\hyperref[stimuli_vs10]{Distractor Categorization (15)}} & Steep & \multirow{3}*{Target Orientation} & \multirow{3}*{$10\times 13$ (130)} \\                                                                                                                   
 &  & Steepest &  &  \\                                                                                                   
 &  & Steep-right &  &  \\                                                                                                 
\hline                                                                                                                     
\end{tabular}                                                                                                              
}                                                                                                                          
\label{table:stimuli_vs}                                                                                                 
\end{table*}

\paragraph{Feature and Conjunctive Search (6)} \label{stimuli_vs1}

Feature search increases probability and efficiency of saccading towards a specific search target on scene observation due to its unique distinctiveness. The information span processed by the HVS  varies depending on the amount of feature distractors to be processed \cite{Treisman1980}\cite{Pashler1988}\cite{Wolfe1989}\cite{Palmer1995}\cite{WOLFE1997}\cite{Hayward2000}\cite{Nothdurft2000}. Given that premise, the amount of objects in a scene would imply a variation of the difficulty towards searching a specific target for the case of serial search (distorting human's sustained attention), but not for the case of parallel search. Previous experiments show that difficulty on visual search is higher with a conjunction of distractors with different image features (such as size, color or orientation). In case distractors vary only by a unique feature, the difficulty of the task would not be as evident as the other case \cite{Nakayama1986}\cite{Treisman1988b}\cite{Wolfe1994}\cite{Flombaum2015}. In order to reproduce feature and conjunctive search, target was a red bar oriented at \SI{45}{\degree}. For the feature search case, distractors were green and set at \SI{45}{\degree} (\hyperref[fig:stimuli_vs1]{Figure \ref*{fig:stimuli_vs1}}). On the case of conjunctive search, half of distractors were green and set at \SI{45}{\degree} and the other half were red and oriented at \SI{-45}{\degree}.

\begin{figure}[H]
 \centering
\begin{subfigure}{0.35\linewidth}
\fbox{\includegraphics[width=1\linewidth]{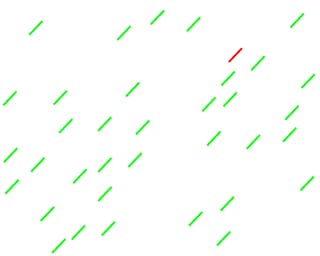}}
\caption{\centering}
\end{subfigure}
\quad
\begin{subfigure}{0.35\linewidth}
\fbox{\includegraphics[width=1\linewidth]{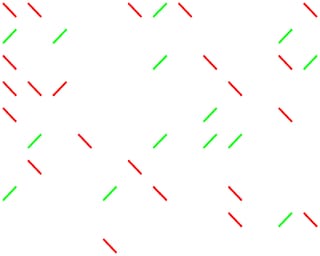}}
\caption{\centering}
\end{subfigure}
\\
\begin{subfigure}{0.35\linewidth}
\fbox{\includegraphics[width=1\linewidth]{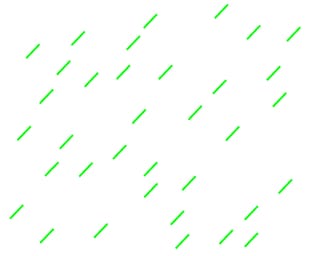}}
\caption{\centering}
\end{subfigure}
\quad
\begin{subfigure}{0.35\linewidth}
\fbox{\includegraphics[width=1\linewidth]{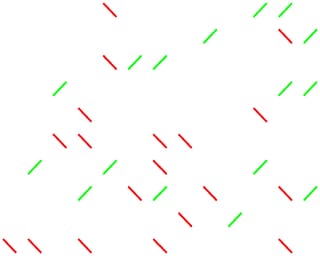}}
\caption{\centering}
\end{subfigure}
\caption{Examples used for feature and conjunctive search. Here are presented the cases of having a red target oriented at \SI{45}{\degree} and 34 distractors randomly displaced around the scene. For the feature search case \textbf{(a)}, all of the distractors are distinct in color (green). For the conjunctive search case \textbf{(b)}, \SI{50}{\percent} of the distractors are distinct in color (green) and the rest are distinctive in orientation (at an orientation of \SI{-45}{\degree}). The same cases \textbf{(c)} and \textbf{(d)} are shown but without the presence of the target.}
\label{fig:stimuli_vs1}
\end{figure}

The position of the items was randomised, with a set size of $\Psi(1...N,2,34)$, the amount of distractors ranged from 2 to 34 distractors. Both search conditions without the presence of the target was introduced in order to see if the effects are also reproduced for the case of reporting absence of target. The design of feature and conjunction search has been defined as keeping similar difficulty between the two conditions, preserving identical targets and displaying each conjunction of distractors maximally dissimilar from each other \cite{Quinlan2003}.

\paragraph{Search Asymmetries (7)} \label{stimuli_vs2}

Search asymmetries between two different type of stimuli happen when a specific target of type "a" is found efficiently among distractors of type "b", but not in the opposite case (searching for "b" among distractors of type "a") \cite{Treisman1985}\cite{Treisman1988}\cite{Hayward2000}\cite{Wolfe2001}. Clear evidence was found for plain circles crossed by a vertical bar (\hyperref[fig:stimuli_vs2]{Figure \ref*{fig:stimuli_vs2}}) at a scale of \SI{5}{\deg}), showing that it was easier to find a circle with a vertical bar among plain circles than vice versa \cite{Wolfe2001}\cite{Spratling2012}. 

\begin{figure}[H]
 \centering
\begin{subfigure}{0.30\linewidth}
\fbox{\includegraphics[width=1\linewidth]{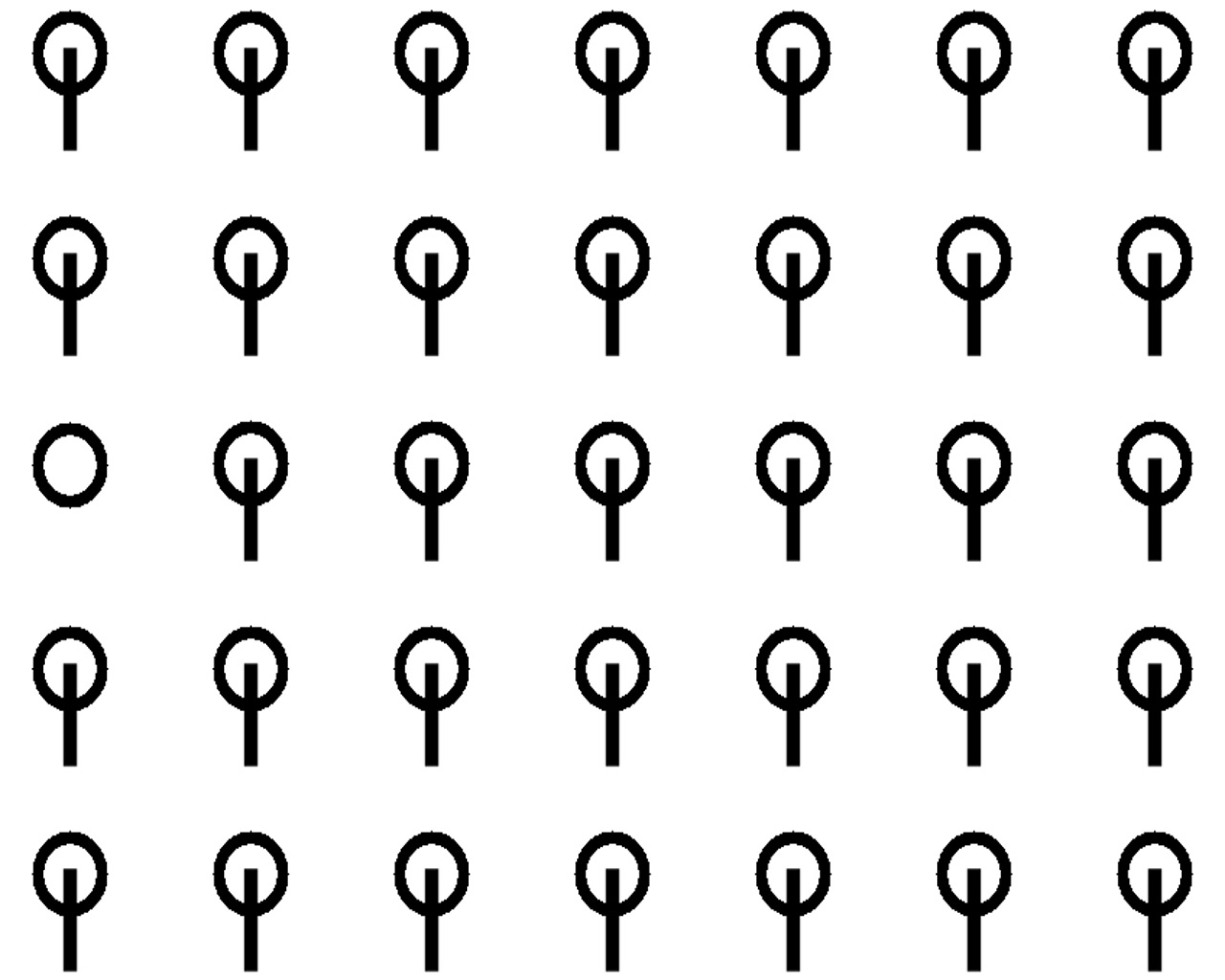}}
\caption{\centering}
\end{subfigure}
\quad
\begin{subfigure}{0.30\linewidth}
\fbox{\includegraphics[width=1\linewidth]{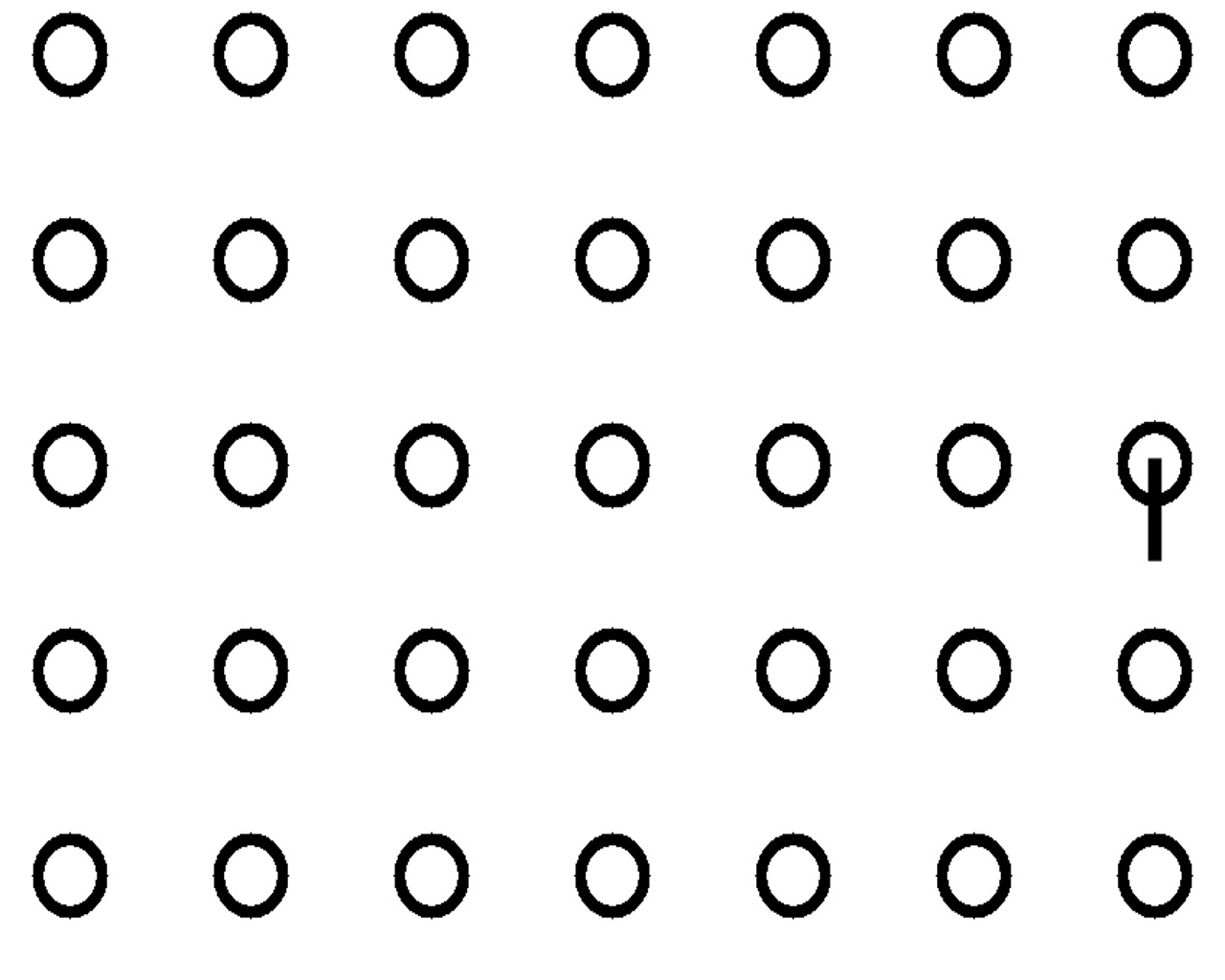}}
\caption{\centering}
\end{subfigure}
\caption{Example of stimulus types in which search asymmetries can apply. The context of circles \textbf{(b)} facilitates the search of a superimposed bar compared to the reverse case \textbf{(a)}.}
\label{fig:stimuli_vs2}
\end{figure}

The same type of stimuli was selected (with both conditions: searching a circle crossed by vertical bar among plain circles and searching a plain circle among circles crossed by a vertical bar) filling a grid of distractors according to a specific scale and randomizing the position of the target. The scale values were $\Psi(1...N,1.25,5)$, between $1.25$ and \SI{5}{\deg}, changing the amount of items to be presented, being in each case from 35 to 520 elements corresponding to arrays of $5\times 7$, $6\times 8$, $8\times 10$, $10\times 13$, $15\times 20$ and $20\times 26$ objects.

\paragraph{Noise/Roughness (8)} \label{stimuli_vs3}

For most synthetic stimuli we have considered uniform and plain backgrounds with homogeneous illumination, but in this case the influence of continuous textured background would increase or reduce search time required to detect a specific target depending on the amount of noise present in the scene. Clarke et. al. \cite{Clarke2008}\cite{Padilla2008}\cite{Clarke2009} showed that the higher the level of background texture noise of a scene, the higher the level of difficulty of the search task. They represented the background surface as a height map by parameterizing an isotropic and random-phase noise $1/f^\beta$  (being "$\beta$" the frequency roll-off magnitude factor of the inverse discrete Fourier transform of the height map and "$\sigma_{RMS}$" the deviation of the roughness noise height). The surface was obtained by rendering the height map according to the Lambert's Cosine Law model using a constant light source with slant of \SI{60}{\degree} and tilt  equal to \SI{90}{\degree}. Given these previous experiments,  each stimulus was a rough surface considering $\beta$ as the contrast value $\Psi(1...N,1.5,1.8)$ with two distinct conditions by using deviations of $\sigma_{RMS}$= $0.9$ and $1.1$. A similar target of Clarke's experimentation was used (\hyperref[fig:stimuli_vs3]{Figure \ref*{fig:stimuli_vs3}}) with a circular shape and a vertical gradient background corresponding to the height of the surface and a diameter of \SI{0.78}{\deg} (half of size corresponding to the rest of target items of this study, adjusted for preventing too low RT differences between distinct contrasts).

\begin{figure}[H]
 \centering
\begin{subfigure}{0.60\linewidth}
\includegraphics[width=1\linewidth]{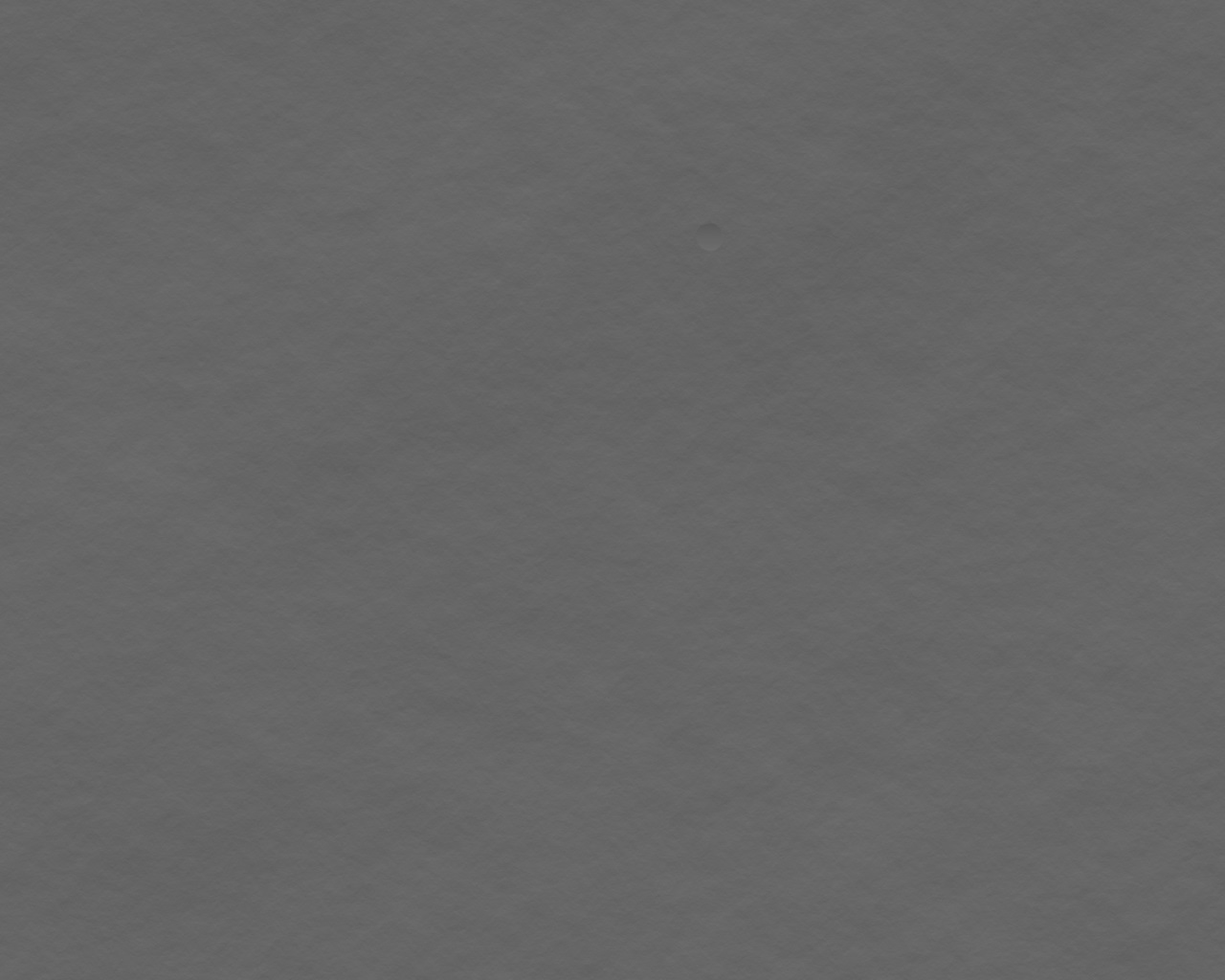} 
\caption{\centering}
\end{subfigure}
\begin{subfigure}{0.60\linewidth}
\includegraphics[width=1\linewidth]{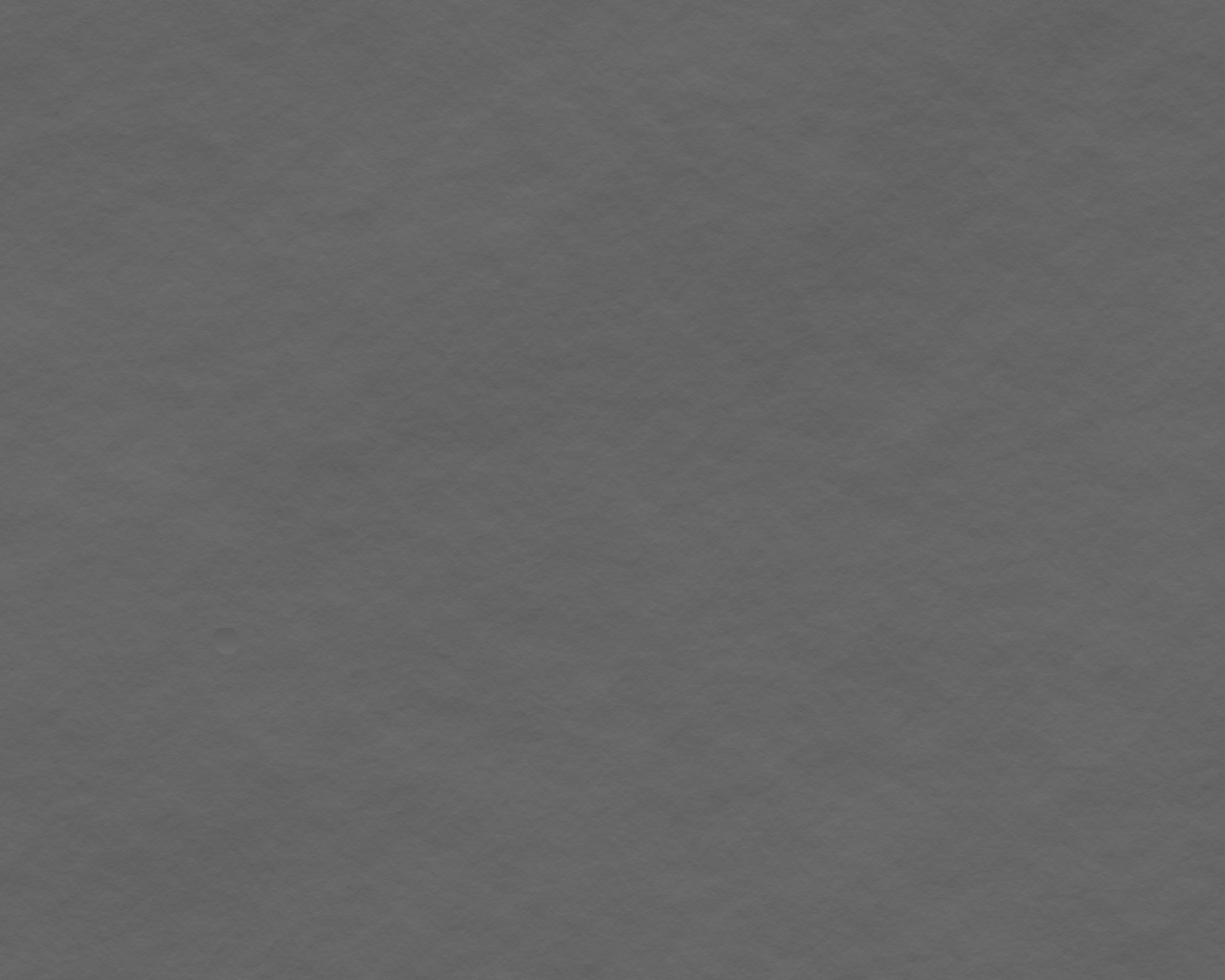}
\caption{\centering}
\end{subfigure}
\caption{Two examples of a rough surface with $\beta=1.8$ using height deviations of \textbf{(a)} $\sigma_{RMS}=0.9$ and \textbf{(b)} $\sigma_{RMS}=1.1$}
\label{fig:stimuli_vs3}
\end{figure}

\paragraph{Distractor similarity} \label{stimuli_vs3-7}

When an object is dissimilar to the rest of objects in a scene, the search of that object is more efficient. That phenomenon is called target-distractor similarity, and has been found to occur when parameterizing specific features such as color, shape or size \cite{Duncan1989}\cite{Wolfe2010a}.

\paragraph{Color Contrast (9)} \label{stimuli_vs4}
In this section the chromatic properties of distractors are changed, as well as the background of the stimuli \hyperref[fig:stimuli_vs4]{Figure \ref*{fig:stimuli_vs4}}. As shown in previous experiments \cite{DZmura1991}\cite{Bauer1996}\cite{Amano2014}\cite{Danilova2014}, color varies spatial and temporal patterns of eye movements, affecting both localization and discrimination of objects. Besides, search asymmetries happen at different background conditions \cite{Nagy1999}\cite{Rosenholtz2004}.

\begin{figure}[H]
 \centering
\begin{subfigure}{0.35\linewidth}
\includegraphics[width=1\linewidth]{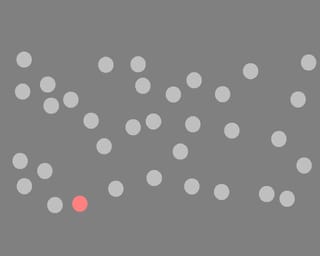} 
\end{subfigure}
\hspace{4ex}
\begin{subfigure}{0.35\linewidth}
\includegraphics[width=1\linewidth]{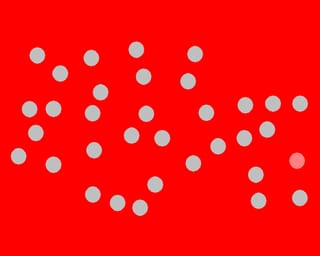}
\end{subfigure}
\\
\begin{subfigure}{0.45\linewidth}
\includegraphics[width=1\linewidth]{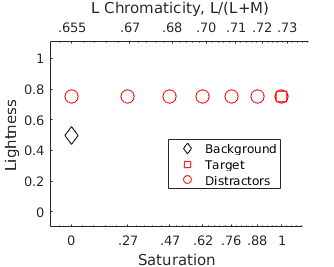} 
\caption{\centering}
\end{subfigure}
\begin{subfigure}{0.45\linewidth}
\includegraphics[width=1\linewidth]{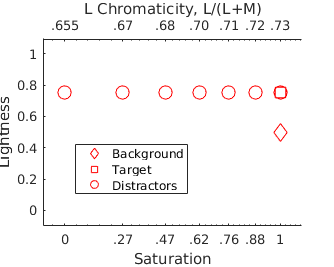} 
\caption{\centering}
\end{subfigure}
\\
\begin{subfigure}{0.35\linewidth}
\includegraphics[width=1\linewidth]{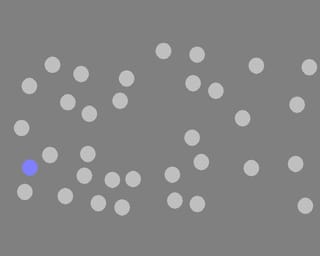} 
\end{subfigure}
\hspace{4ex}
\begin{subfigure}{0.35\linewidth}
\includegraphics[width=1\linewidth]{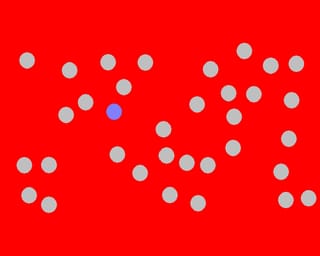} 
\end{subfigure}
\\
\begin{subfigure}{0.45\linewidth}
\includegraphics[width=1\linewidth]{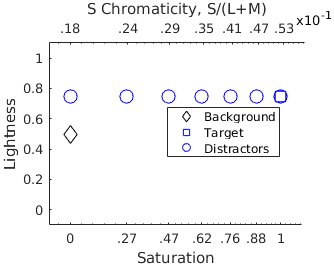} 
\caption{\centering}
\end{subfigure}
\begin{subfigure}{0.45\linewidth}
\includegraphics[width=1\linewidth]{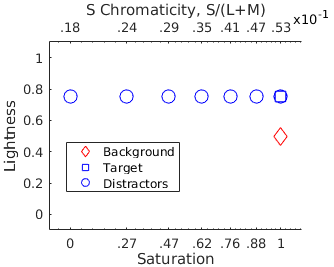} 
\caption{\centering \\ }
\end{subfigure}
\caption{Examples of the 4 conditions at maximum contrast of $\Delta S_{D,T}$=$1$, representing the values of Saturation and Lightness on each particular stimuli for each target and background configuration (showing as well the "l" and "s" chromaticities in the lsY space \cite{MacLeod1979} at \SI{400}{\nano\meter}). In \textbf{(a)} and \textbf{(b)} there are represented the stimulus for grey (unsaturated) and red (oversaturated) background respectively. Similarly, but for blue targets, are represented the cases for both background conditions in \textbf{(c)} and \textbf{(d)}.}
\label{fig:stimuli_vs4}
\end{figure}

\begin{figure*}[htpb!]

    \begin{subfigure}[b]{.4\linewidth}
    \centering
    \includegraphics[width=0.8\linewidth]{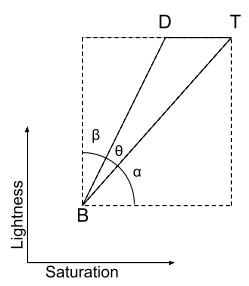}
    \end{subfigure}
    \begin{subfigure}[b]{0.5\linewidth}
    \centering
    	\begin{equation}
        \begin{aligned}
        \Delta S_{1,2} =| S_1 - S_2 |, \\ 
        \Delta L_{1,2} =| L_1 - L_2 |, \\
        \end{aligned}
        \label{eq:DeltaSDeltaL}
        \end{equation}
        \begin{equation}
        \alpha =\arctan(\frac{\Delta L_{T,B}}{\Delta S_{T,B}}), 
        \label{eq:alpha}
        \end{equation}
        \begin{equation}
        \theta =(90-\alpha)(\Psi(1...N)), 
        \label{eq:theta}
        \end{equation}
        \begin{equation}
        \begin{aligned}
        \beta = (90-\alpha)(1-\Psi(1...N))  \\ 
         \equiv 90 - \theta - \alpha  \\ 
         \equiv \arctan(\frac{\Delta S_{B,D}}{\Delta L_{B,D}}), \\
        \end{aligned}
        \label{eq:beta}
        \end{equation}
        \begin{equation}
        \begin{aligned}
        \Delta S_{D,T} =| S_T - S_D | \\ 
        \equiv  |\Delta S_{B,T} - \Delta S_{B,D} |  \\
        \equiv |\Delta S_{B,T} - (\Delta S_{B,D} \cdot tan(\beta))|, \\
        \end{aligned}
        \label{eq:DeltaS}
        \end{equation}
        \begin{equation}
        \mathcal{S_D} =
        \begin{cases}
        S_T - \Delta S_{D,T} & \text{if $S_T > S_B$} \\
       S_B - \Delta S_{D,T} & \text{otherwise}
        \end{cases}
    	\label{eq:S_DT}
        \end{equation}
	
    \end{subfigure}
    \caption{Representation of HSL values \\ for distinct distractors (D), background (B) \\ and search target (T).}
    \label{fig:HSL_BTD}
\end{figure*}

Taking into account these experiments, we wanted to analyze if these search asymmetries are present when varying saturation of distractors with respect to a search target. We will see if these differences between distractor and search target are affected by changing background saturation at distinct target and distractor hue (using the HSL color space). A set of stimulus was generated with circular shaped items with a similar displacement to Rosenholtz experiment. Stimulus was sorrounded with a vertical padding equal to the presented background in order to prevent monitor-related luminance gradients. Contrast values can be calculated according to the saturation differences between the search target (T) and distractors (D). Two background (B) conditions were defined, corresponding to Grey (achromatic and unsaturated), and Red (chromatic and oversaturated) colors. At isoluminant ($L_{D,T}$=0.75) and isohue conditions ($H_{D,T}$=\SI{0}{\degree} for red and $H_{D,T}$=\SI{240}{\degree} for blue distractors), a representative measure of color contrast between the target and distractors can be computed. This measure was named $\theta$, being the angle between the search target and distractors, with the background as the vertex of the intersection.  Same trigonometrical properties can apply using the same diagram plotting B,T and D relationships at distinct quadrants (acknowledging that in our case T is oversaturated for both conditions). 

In \hyperref[eq:DeltaSDeltaL]{Equation \ref*{eq:DeltaSDeltaL}}, is represented the absolute difference in lightness and saturation between two distinct conditions. In \hyperref[fig:HSL_BTD]{Figure \ref*{fig:HSL_BTD}} there are the angles that comprise the saturation and lightness contrast between our stimulus objects. Each of these angles represent respectively to the triangles formed by B-T ($\alpha$), B-D ($\beta$) and D-T ($\theta$), being $\alpha$ constant for constant background and target (\hyperref[eq:alpha]{Equations \ref*{eq:alpha}, \ref*{eq:theta} and \ref*{eq:beta}}). Most importantly, $\theta$ represents the angle comprising the available contrast between the distractor and the target. Given these angle calculations, it is possible to represent the $\Delta S_{D,T}$ as the absolute saturation difference between D and T (\hyperref[eq:DeltaS]{Equation \ref*{eq:DeltaS}}), calculated by the parametrization of $\beta$  using our psychometric function $\Psi(1...N)$. That absolute saturation difference will define the criterion for our distractor saturation $S_D$ as shown in \hyperref[eq:S_DT]{Equation \ref*{eq:S_DT}}. There were generated 4 experimental conditions corresponding to unsaturated and saturated background and red or blue hue. The value of $\theta$ is equivalent for saturated and unsaturated background, corresponding to values of $0, 9, 18, 35, 44$ and \SI{53}{\degree}, producing saturation differences ($\Delta S_{D,T}$) of $0, 0.121, 0.246, 0.528, 0.728$ and $1$.

\paragraph{Brightness Contrast (10)} \label{stimuli_vs5}

According to previous studies, searching a bright target is harder as luminance of distractors increase \cite{Pashler2004}\cite{Nothdurft2006a}\cite{Spratling2012}, with a distinct response with respect to chromatic stimuli \cite{Nagy1999}. Conversely, salience increases for a dark target when luminance of distractors is increased. It was parametrized as the lightness contrast and stimuli was modeled using the HSL color space, considering an achromatic (unsaturated) and isohue relationship between search target, distractors and background, using the same type of stimuli as in \hyperref[stimuli_vs4]{Color Contrast (9)}. Here the target is gray ($L_T=0.5$) and background is bright ($L_B=1$) or dark ($L_B=0$). In order to parametrize the contrast for this stimuli, we used the absolute lightness difference between search target and distractors. In \hyperref[fig:stimuli_vs5]{Figure \ref*{fig:stimuli_vs5}} theta value is \SI{0}{\degree} for all cases, at $0$ saturation, the lighness axis is parametrized.

\begin{equation}
\Delta L_{D,T}=|L_D - L_T|=
  \begin{cases}
  \Delta L_{B,T}(1- \Psi(1...N)) & \text{if $L_T > L_B$} \\
  \Delta L_{B,T}(\Psi(1...N)) & \text{otherwise}
  \end{cases}
  \label{eq:DeltaL}
\end{equation}
\begin{equation}
L_D=
  \begin{cases}
  L_T - \Delta L_{D,T} & \text{if $L_T > L_B$} \\
  L_B - \Delta L_{D,T} & \text{otherwise}
  \end{cases}
  \label{eq:L_D}
\end{equation}

Lightness differences \hyperref[eq:DeltaL]{Equation \ref*{eq:DeltaL}} are calculated by $\Delta L_{D,T}$, corresponding to the absolute difference between target and background lightness ($|\Delta L_{B,T}|$) and adjusted by our psychometric function $\Psi(1...N)$ with distinct distractor lightness values of $L_D$, depending on the absolute background lightness with respect the target.

\begin{figure}[H]
 \centering
\begin{subfigure}{0.35\linewidth}
\fbox{\includegraphics[width=1\linewidth]{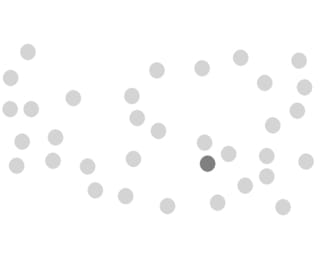}}
\caption{\centering}
\end{subfigure}
\quad
\begin{subfigure}{0.35\linewidth}
\includegraphics[width=1\linewidth]{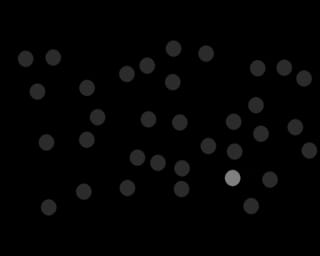} 
\caption{\centering}
\end{subfigure}
\caption{Examples of distinct background conditions, \textbf{(a)} lighter background and distractors at $L_D$=$0.66$. \textbf{(b)} Dark background with distractors at $L_D$=$0.17$. For both conditions, the lightness of the search target is grey ($L_T$=$0.5$). Absolute contrast for these cases is $\Delta L_{D,T}$=$0.33$.}
\label{fig:stimuli_vs5}
\end{figure}

\paragraph{Size contrast (11)} \label{stimuli_vs6}

Dissimilarities in size of objects tend to drive increase or decrease search speed when detecting and discriminating salient regions \cite{Sagi1984}\cite{Goolkasian1997}\cite{Tavassoli2009}\cite{Proulx2010}. Here is presented size similarity between symmetric objects (circles, without loss of generality). Each stimuli was generated with a set of 34 objects randomly around the scene (\hyperref[fig:stimuli_vs6-7]{Figure \ref*{fig:stimuli_vs6-7}\textbf{(a)}}). The search target has a distinct size with respect to the distractors, with both cases of smaller and bigger sizes with $\Psi(1...N, 1.25, 5)$=$[1.25, 1.67, 2.08, 2.5, 3.34, 4.17, 5]\deg$, being the size as the parameter that defines the similarity contrast for this case, corresponding to a scaling factor of $0.5$ to $2$ with respect to the baseline (\SI{2.5}{\deg}).

\paragraph{Orientation contrast (12)} \label{stimuli_vs7}

For this setting, varying angle of objects is found to increase search efficiency when angle contrast is increased \cite{Duncan1989}\cite{Nothdurft1993a}\cite{Nothdurft1993b}\cite{Kong2017}. A set of 34 bars were randomly displaced around the scene and oriented at \SI{0}{\degree}, in which the search target is an equally-shaped bar oriented at a distinct angle (\hyperref[fig:stimuli_vs6-7]{Figure \ref*{fig:stimuli_vs6-7}\textbf{(b)}}). Angle contrast between the search target and the set of distractors was $\Delta\Phi(1,0)$=$[0, 10, 20, 30, 42, 56, 90]$º \hyperref[eq:phi]{Equation \ref*{eq:phi}}.

\begin{figure}[H]
 \centering
\begin{subfigure}{0.35\linewidth}
\fbox{\includegraphics[width=1\linewidth]{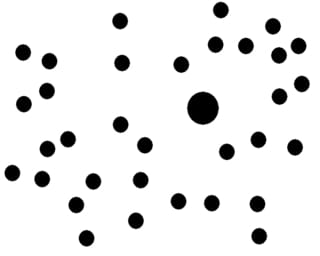}  }
\caption{\centering}
\end{subfigure}
\quad
\begin{subfigure}{0.35\linewidth}
\fbox{\includegraphics[width=1\linewidth]{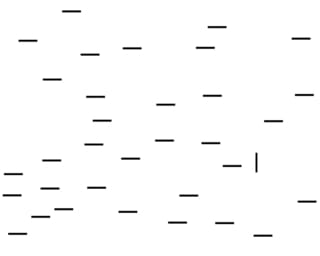}}
\caption{\centering} 
\end{subfigure}
\caption{Examples for salient targets with dissimilar size \textbf{(a)} and orientation \textbf{(b)}. For \textbf{(a)}, the search target has a diameter of \SI{5}{\deg} (a factor of $2$ with respect to the rest of distractors). For \textbf{(b)}, the orientation of the target is \SI{90}{\degree} with distractors at \SI{0}{\degree}, forming an orientation contrast of $\Delta \Phi$=\SI{90}{\degree}.}
\label{fig:stimuli_vs6-7}
\end{figure}

\paragraph{Distractor Heterogeneity (13)} \label{stimuli_vs8}

Previous design was to evaluate orientation similarity, given a unique orientation for non-target distractors. Here is presented the phenomenon of distractor heterogeneity. When several sets of distractors are dissimilar with respect to the search target, mutual information between the target and distractors is said to be heterogeneous. In the heterogeneous case, search efficiency is lower, in other terms, target search is harder \cite{Treisman1988}\cite{Duncan1989}\cite{Nothdurft1993b}\cite{Bauer1996}\cite{Rosenholtz2001}\cite{Gao2008}\cite{Wolfe2010a}. Distractor orientation heterogeneity, however, can be represented through distinct configurations, either if the set of distractors are tilted to the same direction or towards distinct directions. In this experiment, there is an array of bars oriented at \SI{75}{\degree} (with a slope of \SI{15}{\degree} with respect to the vertical quadrant). From two different sets of distractors, are defined three conditions according to the distinct orientation configurations: homogeneous, tilted-right and flanking (\hyperref[fig:stimuli_vs8]{Figure \ref*{fig:stimuli_vs8}}). For the case of homogeneous distractors, both set of distractors have a unique angle contrast with respect to the target bar. For the case of tilted-right, both set of distractors have an angle tilt of $15$ and \SI{30}{\degree}, $\phi(1,90,15,30)$. For the case of flanking, both sets of distractors have an angle tilt of \SI{15}{\degree} and \SI{-30}{\degree} respectively, having both positive and negative tilt with respect the search target, $\phi(1,90,15,-30)$.

\begin{equation}
\phi(v,a,c_1,c_2)=\{\Phi(v,a,c_1),\Phi(v,a,c_2)\},
\label{eq:subphi}
\end{equation}
\begin{equation}
\Delta\phi(v,a,c_1,c_2)=\{\Delta\Phi(v,a,c_1),\Delta\Phi(v,a,c_2)\}.
\label{eq:Deltasubphi}
\end{equation}

For this type of stimuli, is defined the angle contrast from the search target to two set of distractors, represented as contrasts from the first set "$c_1$" and the second set "$c_2$" in \hyperref[eq:Deltasubphi]{Equation \ref*{eq:Deltasubphi}}, being the maximum angle between distractors and search target (considering that bars have angle values on two quadrants for each case) as \SI{90}{\degree}. Our target angle will have values taken from $\Phi(1,90)$ being parameterized with contrast values ranging from $0$ to \SI{90}{\degree}, in order to reveal higher angle contrast values, as heterogeneous distractors are harder to be identified.

\begin{figure}[H]
 \centering
\begin{subfigure}{0.28\linewidth}
\fbox{\includegraphics[width=1\linewidth]{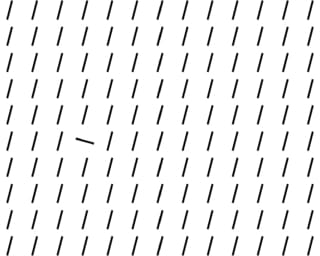}}
\caption{\centering}
\end{subfigure}
\quad
\begin{subfigure}{0.28\linewidth}
\fbox{\includegraphics[width=1\linewidth]{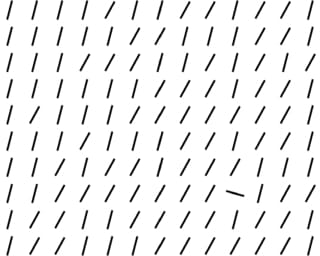}}
\caption{\centering}
\end{subfigure}
\quad
\begin{subfigure}{0.28\linewidth}
\fbox{\includegraphics[width=1\linewidth]{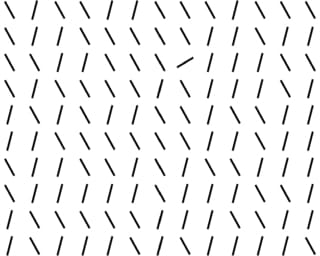}}
\caption{\centering}
\end{subfigure}
\caption{Examples of distinct distractor angle configurations, corresponding to \textbf{(a)} Homogeneous, \textbf{(b)} Tilted-right and \textbf{(c)} Flanking.}
\label{fig:stimuli_vs8}
\end{figure}

\paragraph{Distractor Linearity (14)} \label{stimuli_vs9}

Orientation collinearity facilitates visual guidance when orientation of target differs from its neighbors, making search efficient and in parallel \cite{Nothdurft1993a}\cite{Nothdurft1993b}\cite{Wolfe2010a}. Visual guidance is induced by orientation linearity given an array of bars as defined from the previous stimulus type. Each bar has been oriented with a specific angle, creating a nonlinear pattern for the whole search array (\hyperref[fig:stimuli_vs9]{Figure \ref*{fig:stimuli_vs9}}).  A linear case has also been presented to compare the conspicuity baseline from the other cases.

\begin{figure}[H]
 \centering
\begin{subfigure}{0.30\linewidth}
\fbox{\includegraphics[width=1\linewidth]{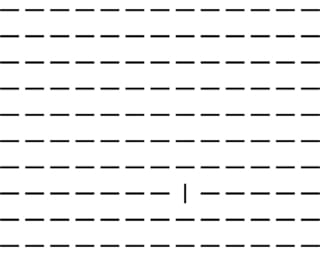}}
\caption{\centering}
\end{subfigure}
\quad
\begin{subfigure}{0.30\linewidth}
\fbox{\includegraphics[width=1\linewidth]{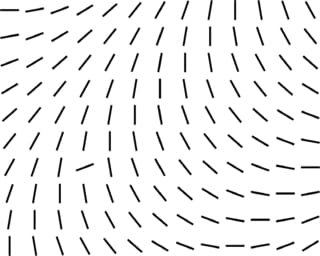}}
\caption{\centering}
\end{subfigure}
\quad
\begin{subfigure}{0.30\linewidth}
\fbox{\includegraphics[width=1\linewidth]{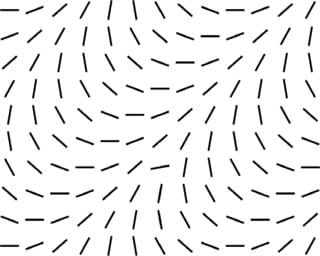}}
\caption{\centering}
\end{subfigure}
\quad
\begin{subfigure}{0.30\linewidth}
\fbox{\includegraphics[width=1\linewidth]{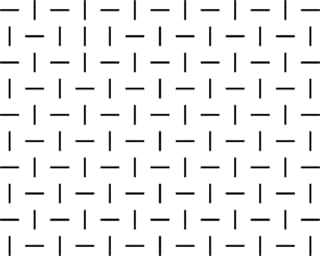}}
\caption{\centering}
\end{subfigure}
\caption{Examples of elicited guidance according to distinct linearity of the distractors. In \textbf{(a)}, distractors are set using the same angle contrast of \SI{90}{\degree} with respect to the target. Conversely, in \textbf{(b)},\textbf{(c)} and \textbf{(d)}, nonlinear patterns are set at an accumulative slope of $10$, $20$ and \SI{90}{\degree} respectively.}
\label{fig:stimuli_vs9}
\end{figure}

\begin{equation}
\varphi(u,row,col)=u \cdot row + u \cdot col,
\label{eq:varphi}
\end{equation}
\begin{equation}
\Delta\varphi(v,u,row,col)=\Phi(v,0)+\varphi(u,row,col).
\label{eq:Deltavarphi}
\end{equation}

Angle contrast is calculated as the orientation difference from the corresponding value of nonlinearity pattern "u" at a certain position on the array "row, col" with a maximum angle contrast with respect to search target (\hyperref[eq:varphi]{Equations \ref*{eq:varphi} and \ref*{eq:Deltavarphi}}).

\paragraph{Distractor Categorization (15)} \label{stimuli_vs10}

Visual search for an oriented bar can be inefficient with distractors at 2 different orientations \cite{Gao2008}\cite{Wolfe2010a}. However, it was found that not all orientations present on an image are equally coded in pre-attentive vision, different configurations of the orientations of the target and the distractions lead to different discriminability \cite{Treisman1991}. Some of these orientation configurations were categorized as "steep", in which search target is identified more efficiently. Other categories of heterogeneous distractors presented harder target search and were dependent on set size. The three categories were modeled, corresponding to "steep", "steepest", "steep-right" as defined by Wolfe et. al. \cite{Wolfe1992}. By considering the same orientation contrast between the two sets of angles, target orientation was parametrized in order to reveal at which orientation contrast is the target to both types of distractors that form these categories. We have modeled these three orientation configurations for each distractor pair, corresponding here to $-50$,\SI{50}{\degree} for steep, $-30$,\SI{70}{\degree} for steepest, $20$,\SI{80}{\degree} for steep-right (\hyperref[fig:stimuli_vs10]{Figure \ref*{fig:stimuli_vs10}}). There was the same amount of distractors for each condition as shown for search on \hyperref[stimuli_vs8]{Distractor Heterogeneity (13)} and \hyperref[stimuli_vs9]{Distractor Linearity (14)} in order to uniquely analyze orientation contrast and preserving similar stimulus type conditions. As shown in section \hyperref[stimuli_vs8]{Distractor Heterogeneity (13)}, the orientation values for each set are computed with \hyperref[eq:subphi]{Equation \ref*{eq:subphi}} and the contrast with respect to the target as \hyperref[eq:Deltasubphi]{Equation \ref*{eq:Deltasubphi}}. The maximum orientation contrast was calculated for all conditions at \SI{40}{\degree} ($v=90/40$) considering the interference of bar orientation contrast in all quadrants (between the target and both distractor orientations). Target angle had psychometric values of $\phi(v,90,-50,50)$ for steep, $\phi(v,90,-30,70)$ for steepest and $\phi(v,90,20,80)$ for steep-right.

\begin{figure}[H]
 \centering
\begin{subfigure}{0.28\linewidth}
\fbox{\includegraphics[width=1\linewidth]{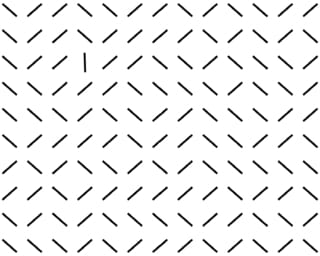}}
\caption{\centering}
\label{fig:subim1}
\end{subfigure}
\quad
\begin{subfigure}{0.28\linewidth}
\fbox{\includegraphics[width=1\linewidth]{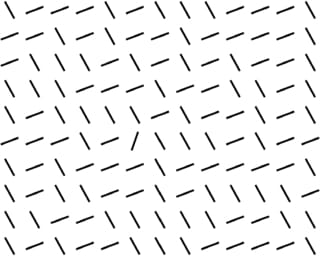}}
\caption{\centering}
\end{subfigure}
\quad
\begin{subfigure}{0.28\linewidth}
\fbox{\includegraphics[width=1\linewidth]{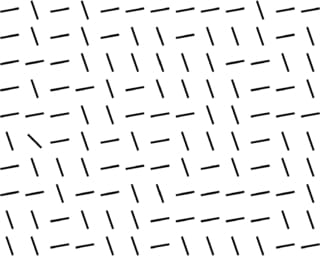}}
\caption{\centering}
\label{fig:subim2}
\end{subfigure}
\caption{Examples of distinct distractor angle configurations, corresponding to \textbf{(a)} Steep, \textbf{(b)} Steepest and \textbf{(c)} Steep-right.}
\label{fig:stimuli_vs10}
\end{figure}

\subsection{Data Analysis} \label{data_analysis}

In order to get the spatial relevance of participant's eye movements we generated binary maps from fixation coordinates. Fixation density maps are computed with a symmetric Gaussian low-pass filtering (with a window size of [$6\sigma$ x $6\sigma$]) of the respective binary maps. A value of $\sigma=1$ $deg$ was used, as recommended by LeMeur and Baccino \cite{LeMeur2012}, corresponding in our case to 40 pixels. The saliency index (SI) is a measure that relates the energy inside a specific region (that can be manually selected, such as a pop-out region) and the one outside that region.


\begin{equation}
SI(S_t,S_b)=\frac{S_t-S_b}{S_b}.
\label{eq:sindex}
\end{equation}

We adapted the metric from Spratling's work \cite{Spratling2012} in order to present positive values as a better representation of the SI (\hyperref[eq:sindex]{Equation \ref*{eq:sindex}}). The distribution of fixations inside ($S_t$) and outside ($S_b$) the area of interest (AOI) will be extracted by cropping the fixation density map using the mask presented on \hyperref[fig:example_qualitative]{Figure \ref*{fig:example_qualitative}}.

\begin{figure}[H]
\centering
\begin{subfigure}{0.28\linewidth} \quad
\includegraphics[width=1\linewidth,height=20mm]{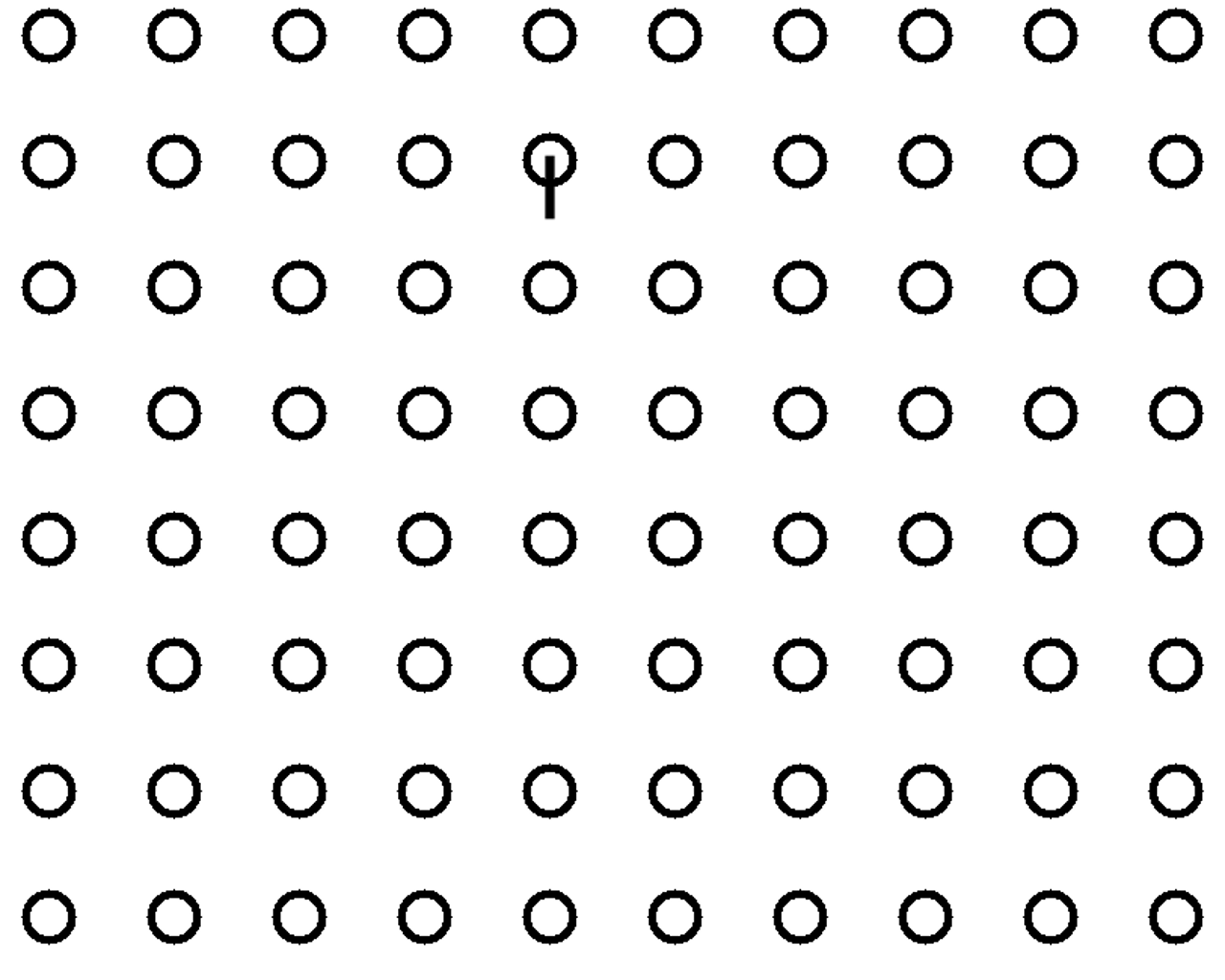} 
\end{subfigure}
\begin{subfigure}{0.28\linewidth} \quad
\includegraphics[width=1\linewidth,height=20mm]{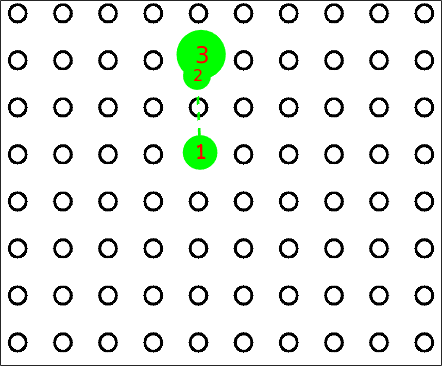} 
\end{subfigure}
\begin{subfigure}{0.28\linewidth} \quad
\includegraphics[width=1\linewidth,height=20mm]{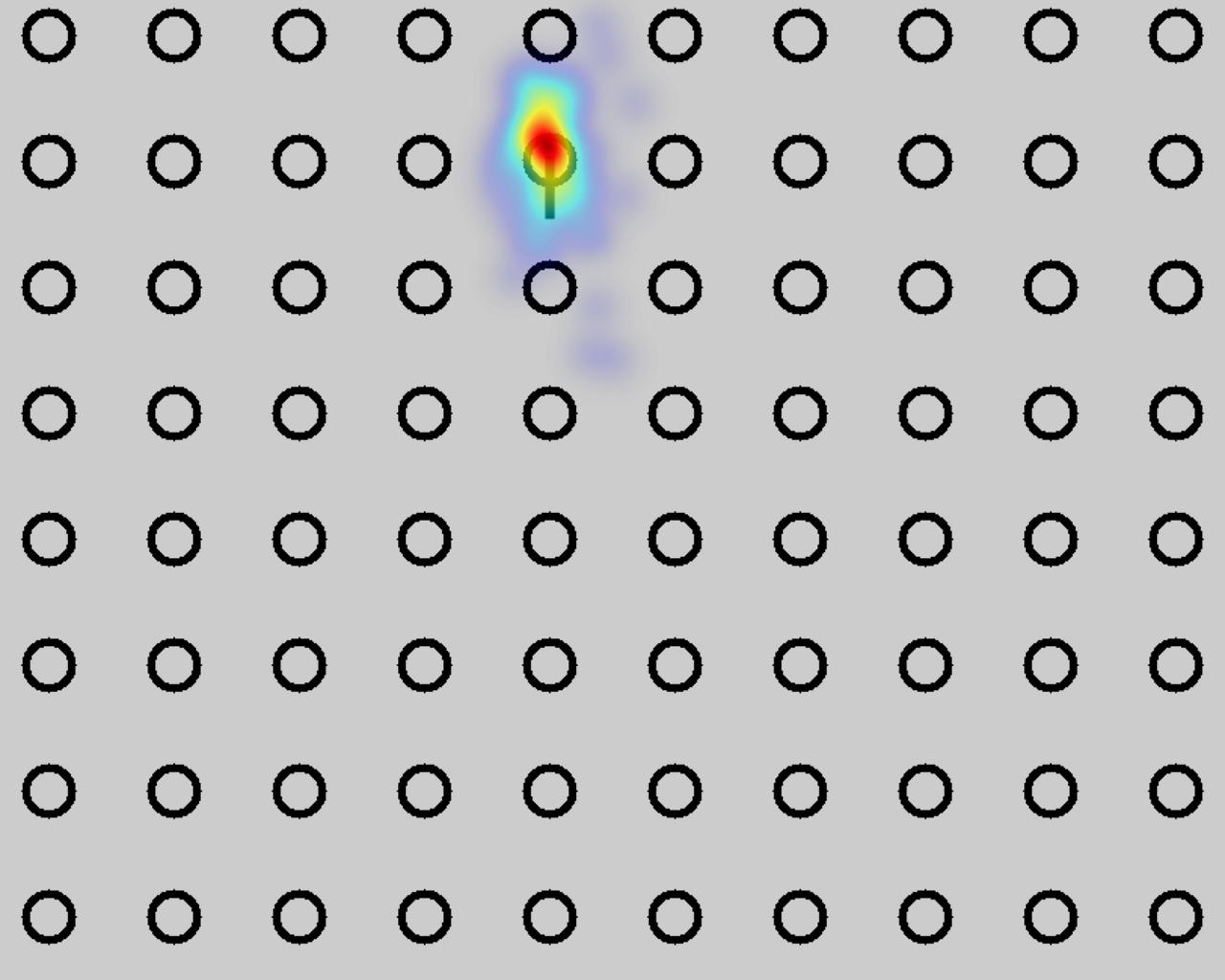} 
\end{subfigure} 
\begin{subfigure}{0.05\linewidth} \quad
\includegraphics[width=0.5\linewidth,height=20mm]{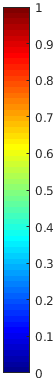} 
\end{subfigure} 
\\\vspace{2mm}
\begin{subfigure}{0.28\linewidth} \quad
\includegraphics[width=1\linewidth,height=20mm]{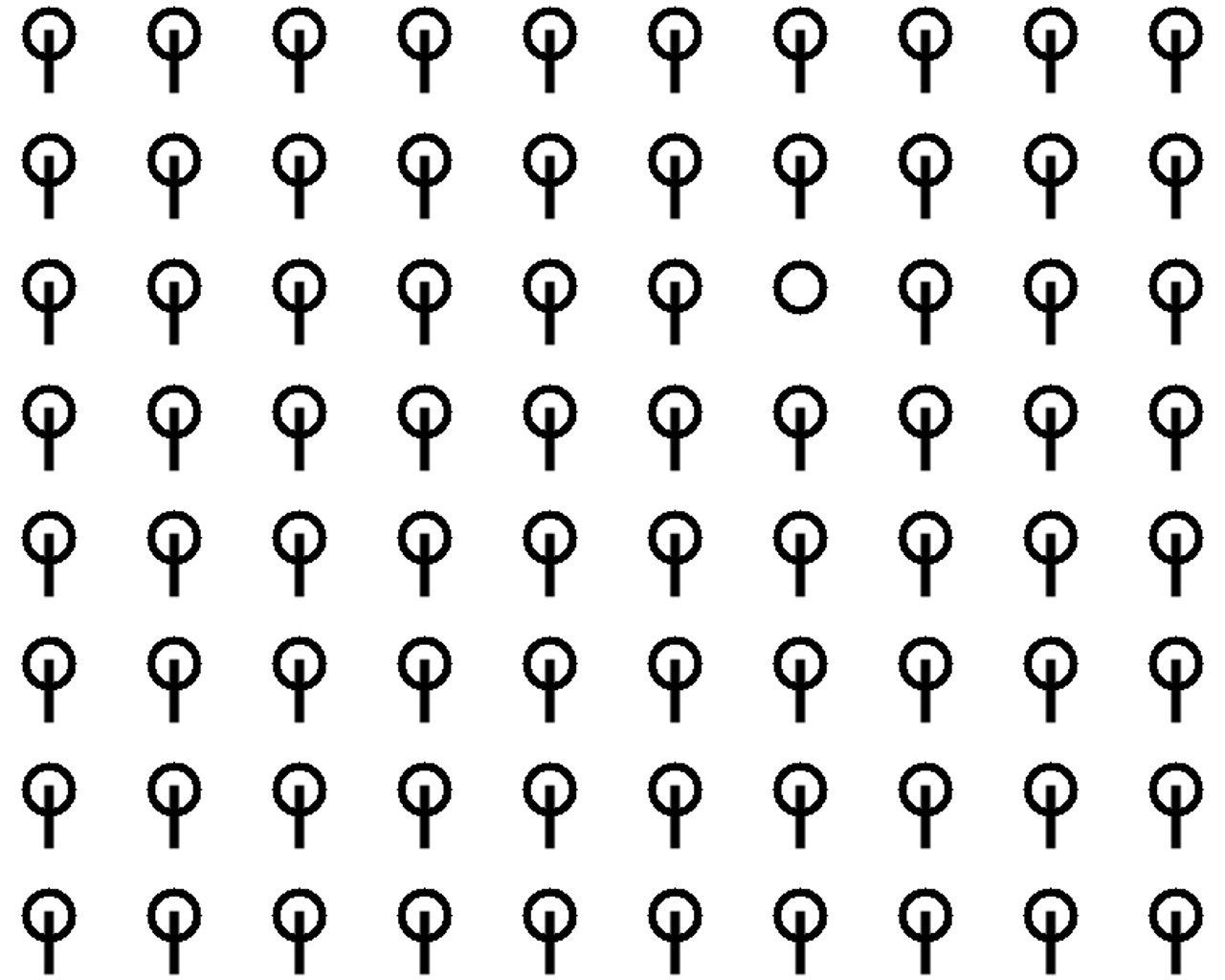} 
\caption{\centering} 
\end{subfigure}
\begin{subfigure}{0.28\linewidth} \quad
\includegraphics[width=1\linewidth,height=20mm]{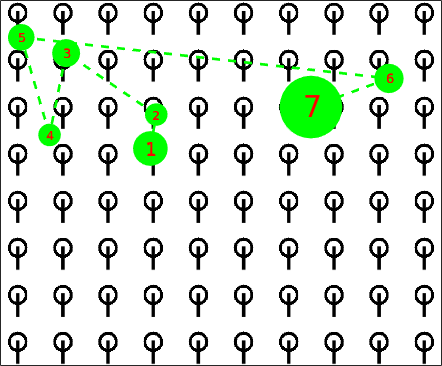} 
\caption{\centering} 
\end{subfigure}
\begin{subfigure}{0.28\linewidth} \quad
\includegraphics[width=1\linewidth,height=20mm]{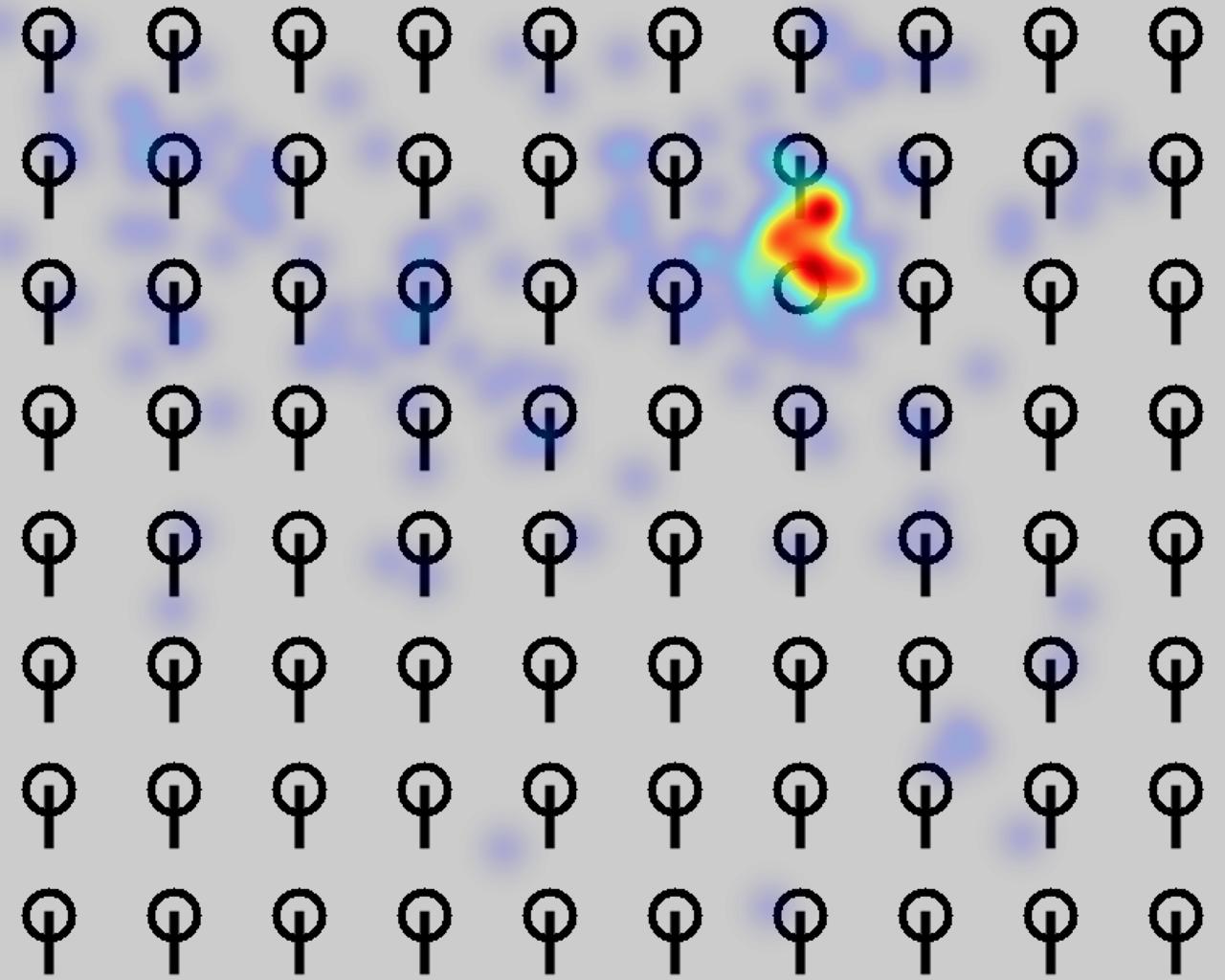} 
\caption{\centering} 
\end{subfigure} 
\begin{subfigure}{0.05\linewidth} \quad
\includegraphics[width=0.5\linewidth,height=20mm]{figures/colormap_jet.png} 
\caption*{\centering} 
\end{subfigure} 
\caption{First row shows a grid with $8\times 10$ circles, in which one of them becomes salient because of having a superimposed bar. On the second row, the superimposed bar is located instead on the rest of circles (distractors). \textbf{(a)} Representation of the mask, corresponding to the AOI of the search target in green $(S_t)$ and the background in red $(S_b)$. \textbf{(b)} Example of a scanpath of a single participant, representing each saccade with a green dashed line and each fixation number with a diameter corresponding to its fixation duration. \textbf{(c)} Superposed density map from the accumulation of fixations for all participants for such stimuli. The colorbar represents the probability of the density distribution.}
\label{fig:example_qualitative}
\end{figure}

For evaluating the SI for a specific sample, a binary visual mask of the salient region (or AOI) needs to be manually created. Given samples at distinct fixation or saccade number, it is possible to compute gaze-wise SI in order to evaluate the temporal evolution of that measure. Such metric can provide a gold standard of spatial performance in terms of how a region pops out with respect to the rest using fixation density maps from recorded eye movements. In other words, measuring the SI using the fixation density maps is the same as measuring the distribution of fixations that have been recorded inside a particular region of an image. Same parameters of the SI metric are preserved from previous studies \cite{Soltani2010}. Although other parameters (such as mask area) could be included to better represent the data, this could be a metric to be exploited in future studies.

In order to get the performance of participants on salient region localization, we recorded the reaction time (RT) on landing inside the AOI. For the free-viewing tasks, we recorded this RT from the initial fixation until the gaze landed inside the AOI. Once a fixation was outside the AOI, we recorded the time until the gaze returned to the AOI, being in this case produced by inhibition of return (IOR) mechanisms. For the visual search tasks, we recorded in a similar way the first fixation inside the AOI as well as visual discrimination. For this latter case, dwell fixations were pinpointed as being inside the AOI during 1000 ms in order to report identification of search targets. For the cases in which participants could not find the stimulus target, the RT corresponded to key pressing. We used fixation data for reporting target localization on both free-viewing and search tasks and the dwelling method for reporting target identification for visual search tasks. In that way, it is possible to discard non-representative fixations and saccades that could be present by other methods such as key trigger, that could imply spatial and temporal deviations with respect to both visual localization and identification. Given an image where salient regions are known, if the SI and the RT reproduce similar results at distinct tasks, feature contrasts and stimulus types, the SI could provide a way to spatially measure how salient is an object, considering specific regions as pop-out instead of using fixations across the whole scene as ground truth. The usage of eye tracking experiments and regions of interest for calculating localization RT instead of keyboard triggers reveals a more accurate way for evaluating visual attention, as no temporal delays are presented from the time since the participants see the search target to report that they have seen it. That method also allows to prevent them to attend to other regions outside the experimental source over time, such as looking towards the keyboard, which can impair their perceptual adaptability (in terms of light sensitivity and foveation).

\section{Results} \label{results}

A total of $90,100$ fixations were recorded over approximately 30 hours of viewing time. The mean number of fixations per stimulus was $M=\num{12(1)}$, corresponding to $M=\num{15(1)}$  for free-viewing (given from \SI{5000}{\milli\second} of viewing time) and $M=\num{11(1)}$ for visual search task stimulus (given from the total viewing time until the stimulus trigger, corresponding to target identification). 
Mean fixation duration was $M=\num{240(1)} \quad ms$ and it was not presenting significant differences from the two types of tasks. See that both distributions of Fixation Duration (FD) and Saccade Amplitude (SA) (\hyperref[fig:results1_hist]{Figure \ref*{fig:results1_hist}}) were skewed to lower values with their upper and lower quartiles at approximately $100$ and \SI{300}{\milli\second} for FD and $2$ and \SI{5}{\deg} of SA. We have also plotted the CDF for both variables and results show that most eye movements (80\%) have a FD of less than \SI{300}{\milli\second} and SA tend to be shorter than \SI{10}{\deg}.

\begin{figure}[H]
 \centering
\begin{subfigure}{0.24\textwidth}
\includegraphics[width=1\linewidth]{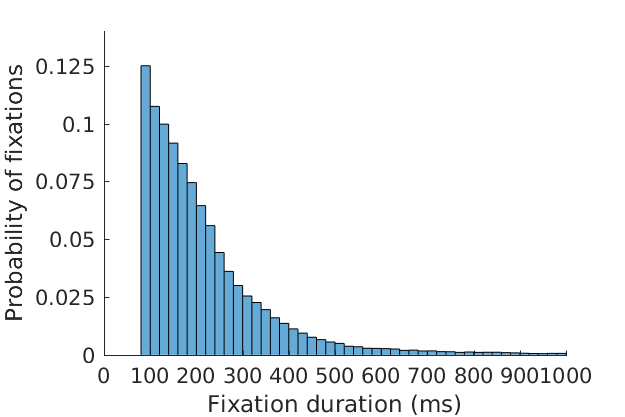} 
\caption{\centering}
\end{subfigure}
\begin{subfigure}{0.24\textwidth}
\includegraphics[width=1\linewidth]{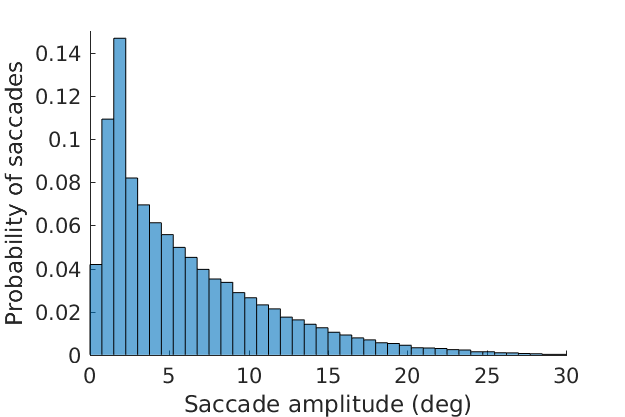} 
\label{fig:results1_histograms}
\caption{\centering}
\end{subfigure}
\begin{subfigure}{0.24\textwidth}
\includegraphics[width=1\linewidth]{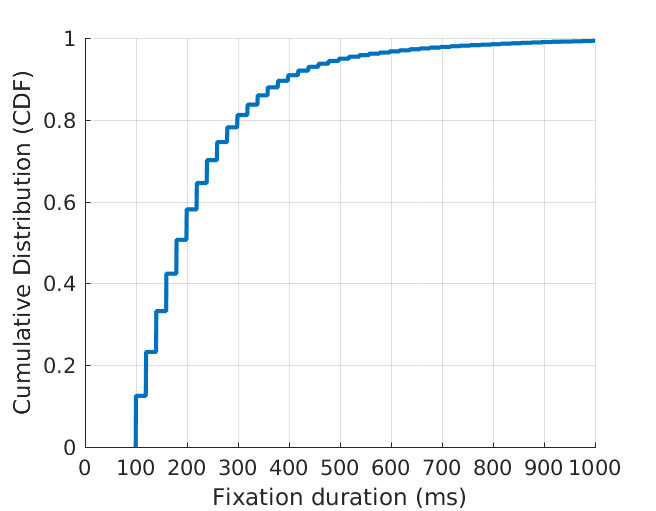} 
\caption{\centering}
\end{subfigure}
\begin{subfigure}{0.24\textwidth}
\includegraphics[width=1\linewidth]{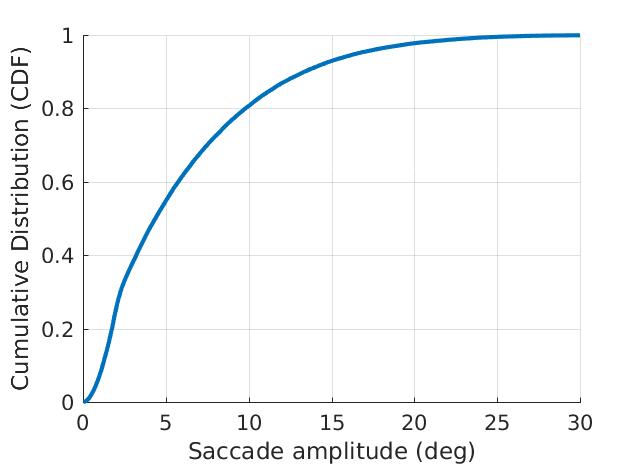} 
\caption{\centering}
\end{subfigure}
\caption{\textbf{(a)} Distribution of Fixation Duration (FD), measured as the absolute fixation time for all samples upon the probability of fixations. \textbf{(b)} Distribution of Saccade Amplitude (SA), measured as the absolute euclidean distance between saccade initiation and saccade landing all samples upon the probability of saccades. \textbf{(c,d)} Cumulative Distribution Functions for FD and SA.
}
\label{fig:results1_hist}
\end{figure}

The overall number of fixations was larger for images containing less salient regions, categorized as hard, requiring more fixations for participants in stimulus with less feature contrast. Localization probabilities were calculated, based on the scanpaths in which participants' gaze landed inside the corresponding AOI. Our results report easiest targets more probable to be localized for both free-viewing ($p$=$6.2 \times 10^{-4}$, Z=$3.4$, $P_{easy}$=$0.38$, $P_{hard}$=$0.30$) and visual search tasks ($p$=$6.9 \times 10^{-88}$, Z=$19.9$, $P_{easy}$=$0.72$, $P_{hard}$=$0.47$). After calculating the reaction times for target localization (landing inside the AOI) and identification (reporting presence of target), we discarded samples where $RT$\textgreater2$\sigma_{RT}$. In that manner we could counteract the impact from oculomotor biases in relation to the localization time with respect objects with approximately that size. As the search targets were smaller for most visual search stimuli, hence less dependent to their respective distance from the stimulus center, we did not discard the respective samples.

\subsection{Performance upon Feature Type \\ (1st Hypothesis)} \label{results1}

RTs for AOI localization are evaluated for each stimulus type and task respectively. Since overall data do not follow a normal distribution (through lilliefors test), Kruskal-Wallis tests were performed in order to evaluate task differences for each contrast difficulty (easy vs hard) as well as differences in RT between distinct type of stimuli. As feature contrasts follow distinct contrast values, we want to test if some features have similarities in RT and their interactions. For each stimulus type, the RT is different given the feature type for both free-viewing (\hyperref[fig:results3_fv]{Figure \ref*{fig:results3_fv}}) and visual search task stimuli (\hyperref[fig:results3_vs]{Figure \ref*{fig:results3_vs}}).

For the former, there were significant differences ($p$=$1.00 \times 10^{-10}$, $\tilde{\chi}^2$=$52.7$, $Mdn_{(1)}$=$523$, $Mdn_{(2)}$=$615$, $Mdn_{(3)}$=$604$, $Mdn_{(4)}$=$736$, $Mdn_{(5)}$=\SI{684}{\milli\second}) between distinct stimulus types RTs, being \hyperref[stimuli_fv1]{Corner Angle (1)} the fastest stimulus to localize the salient region and \hyperref[stimuli_fv4]{Contour Integration (4)} the slowest.

For the latter, there were significant differences ($p$=$1.11 \times 10^{74}$, $\tilde{\chi}^2$=$372$, $Mdn_{(6)}$=$782$, $Mdn_{(7)}$=$742$, $Mdn_{(8)}$=$942$, $Mdn_{(9)}$=$892$, $Mdn_{(10)}$=$593$, $Mdn_{(11)}$=$787$, $Mdn_{(12)}$=$622$, $Mdn_{(13)}$=$606$, $Mdn_{(14)}$=$676$, $Mdn_{(15)}$= \SI{952}{\milli\second}) on RTs for searching salient regions, showing highest performance for \hyperref[stimuli_vs7]{Orientation Contrast (12)} and \hyperref[stimuli_vs10]{Distractor Categorization (15)} the lowest.

By computing the saliency index from the density maps across all fixations and the stimulus masks, it is possible to spatially evaluate saliency (in terms of number of fixations inside the window, represented as a heat map isotropically distributed using a Gaussian filter) given from each stimulus types.

\begin{figure}[H]
 \centering
\begin{subfigure}{0.35\textwidth}
\includegraphics[width=.9\linewidth]{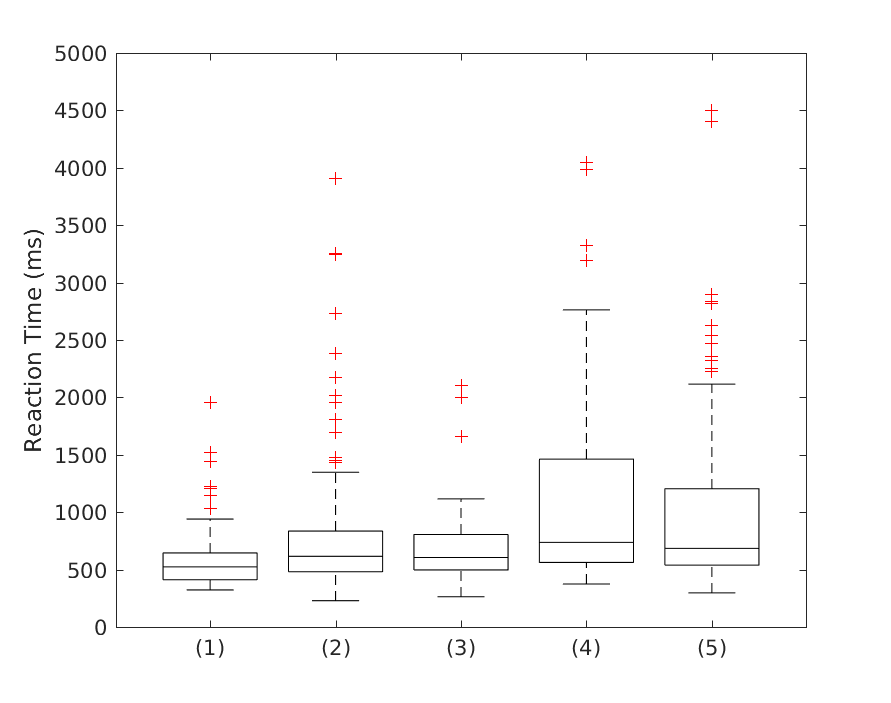} 
\caption{\centering}
\end{subfigure}
\begin{subfigure}{0.35\textwidth}
\includegraphics[width=.9\linewidth]{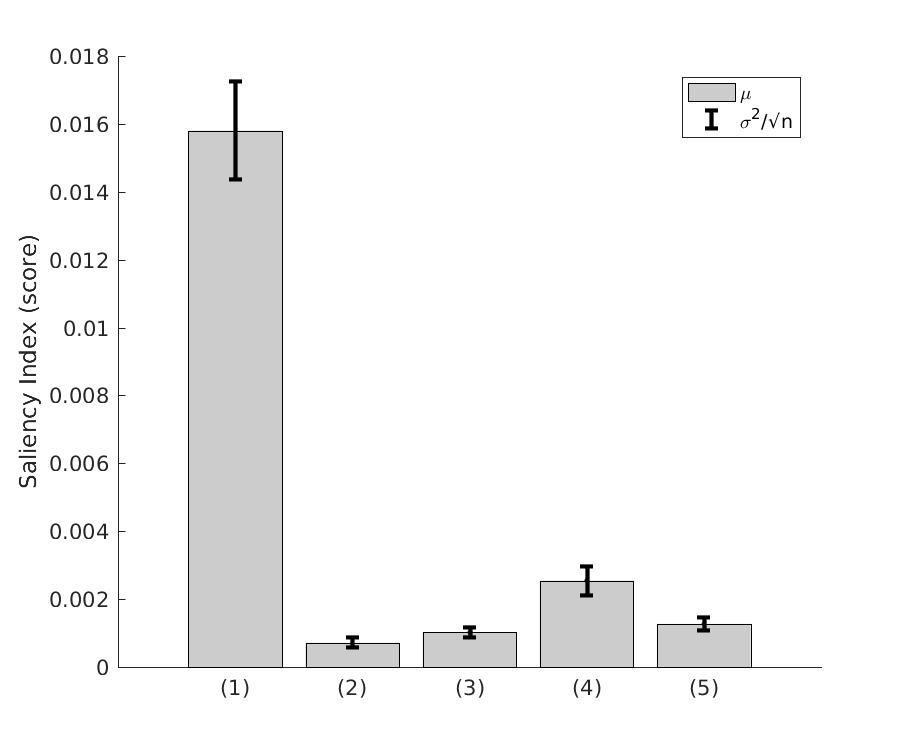} 
\caption{\centering}
\end{subfigure}
\caption{Plots for salient region localization time \textbf{(a)} and saliency index \textbf{(b)} corresponding to stimulus types of \hyperref[stimuli_fv1]{Corner Angle (1)}, \hyperref[stimuli_fv2]{Segmentation by Angle (2)}, \hyperref[stimuli_fv3]{Segmentation by Distance (3)}, \hyperref[stimuli_fv4]{Contour Integration (4)} and \hyperref[stimuli_fv5]{Perceptual Grouping (5)}}
\label{fig:results3_fv}
\end{figure}

Similarly, there were significant differences on SI depending on stimulus types for free-viewing  ($p$=$3.5 \times 10^{-7}$, $\tilde{\chi}^2$=$36$, $Mdn_{(1)}$=$1.69\times 10^{-2}$, $Mdn_{(2)}$=$6.7\times 10^{-4}$, $Mdn_{(3)}$=$1.1\times 10^{-3}$, $Mdn_{(4)}$=$2.2\times 10^{-3}$, $Mdn_{(5)}$=$1.2\times 10^{-3}$ and visual search ($p$=$4.9\times 10^{-6}$, $\tilde{\chi}^2$=$41$, $Mdn_{(6)}$=$32\times 10^{-3}$, $Mdn_{(7)}$=$1.4\times 10^{-2}$, $Mdn_{(8)}$=$8.0\times 10^{-3}$, $Mdn_{(9)}$=$1.8\times 10^{-2}$, $Mdn_{(10)}$=$4.0\times 10^{-2}$, $Mdn_{(11)}$=$1.6\times 10^{-2}$, $Mdn_{(12)}$=$4.0\times 10^{-2}$, $Mdn_{(13)}$=$3.9\times 10^{-2}$, $Mdn_{(14)}$=$3.1\times 10^{-2}$, $Mdn_{(15)}$=$13\times 10^{-3})$. Stimulus with higher SI for free-viewing task was \hyperref[stimuli_fv1]{Corner Angle (1)} and the lower was \hyperref[stimuli_fv2]{Visual Segmentation (2-3)}. For the case of visual search task stimuli, most salient targets were on stimulus presented on \hyperref[stimuli_vs6]{Size (12)} and \hyperref[stimuli_vs6]{Orientation (13)} contrast and the least ones on \hyperref[stimuli_vs2]{Noise/Roughness (8)} and \hyperref[stimuli_vs6]{Distractor categorization (12)} search.

Given the aforementioned results shown for \hyperref[fig:results3_vs]{Figures \ref*{fig:results3_fv} and \ref*{fig:results3_vs}}, RTs were lower (faster) for stimuli with higher SI. The reverse case applies for lower RTs. Target identification (when participats voluntarily report to identify the search target, as explained in \hyperref[data_analysis]{Section \ref*{data_analysis}}) was shown to be slower than target localization ($p$\textless$1.3 \times 10^{-111}$,Z=$-22.4$, $Mdn_{localization}$=$726$, $Mdn_{identification}$=\SI{1026}{\milli\second}), supporting the literature \cite{Sagi1984} \cite{Nothdurft2006b} with an absolute mean time difference of $M$=\num{415(203)} ms.

\begin{figure}[H]
 \centering
\begin{subfigure}{0.35\textwidth}
\includegraphics[width=.9\linewidth]{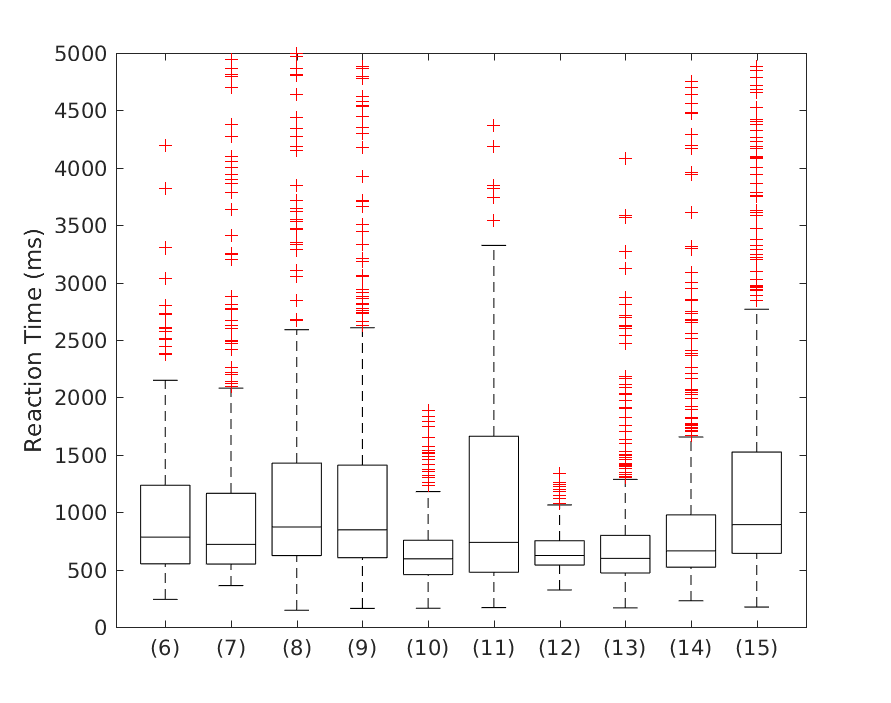} 
\caption{\centering}
\end{subfigure}
\begin{subfigure}{0.35\textwidth}
\includegraphics[width=.9\linewidth]{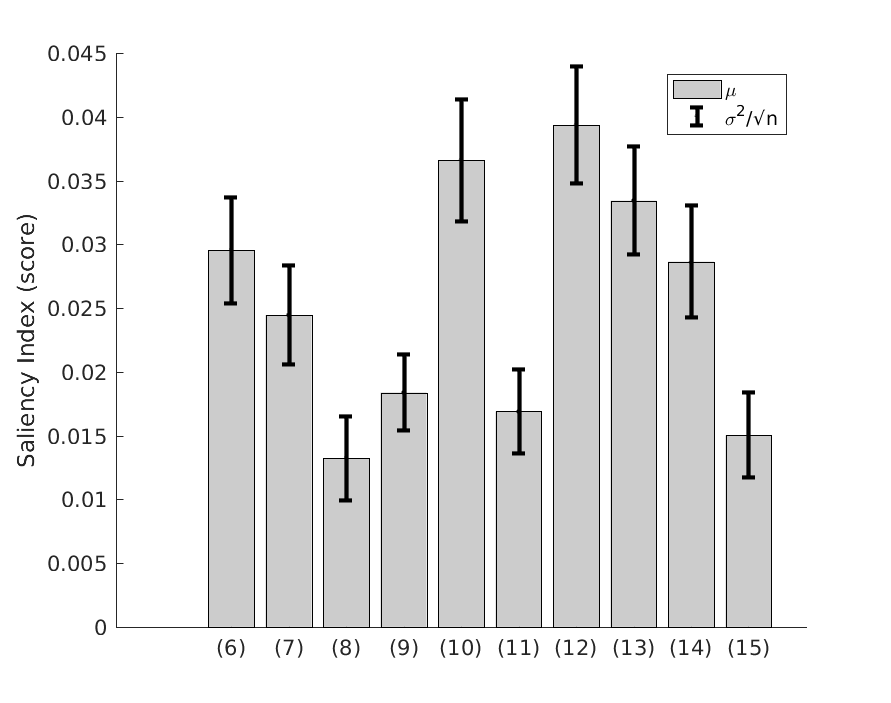} 
\caption{\centering}
\end{subfigure}
\caption{Plots for salient region localization time \textbf{(a)} and saliency index \textbf{(b)} corresponding to stimulus types of \hyperref[stimuli_vs1]{Feature and Conjunctive Search (6)}, \hyperref[stimuli_vs2]{Search Asymmetries (7)}, \hyperref[stimuli_vs3]{Noise/Roughness (8)}, \hyperref[stimuli_vs4]{Color Contrast (9)}, \hyperref[stimuli_vs5]{Brightness Contrast (10)}, \hyperref[stimuli_vs6]{Size Contrast (11)}, \hyperref[stimuli_vs7]{Orientation Contrast (12)}, \hyperref[stimuli_vs8]{Distractor Heterogeneity (13)}, \hyperref[stimuli_vs9]{Distractor Linearity (14)} and \hyperref[stimuli_vs10]{Distractor Categorization (15)}.}
\label{fig:results3_vs}
\end{figure}

\subsubsection*{Discussion}

We can observe that saliency is induced through varying distinct features of the images. Fixations from participants are shown to localize salient regions significantly with distinct performance depending on feature type and the amount of fixations are distributed or spread distinctively across these regions. These aforementioned observations might be influenced by distinct processing (and correlates) of the visual features in the HVS. 

\subsection{Performance upon Feature Contrast \\ (2nd Hypothesis)} \label{results2}


Measures of RTs and SI for salient region localization were computed for each stimulus target-distractor contrast. Overall RT data was not normally distributed, but individual data per stimulus type was normally distributed. Mean RT and error is represented according to the stimulus contrast as well as its mean SI. Spearman's rank correlation tests show that there was a significant negative correlation between RT and SI ($\rho_{RT,SI}$=$-.44$, $p_{RT,SI}$=$2.2 \times 10^{-195}$), suggesting that SI is a plausible measure for representing saliency on a particular region (higher SI and lower RT implies faster localization speed). In that respect both RT and SI were related to stimulus feature contrast (CT) measurements (shown on \hyperref[stimulus]{Section \ref*{stimuli}} and \hyperref[fig:resFV]{Figures \ref*{fig:resFV},\ref*{fig:resVSa} and \ref*{fig:resVSb}}). RT was negatively correlated with respect to CT ($\rho_{CT,RT}$=$-.14$, $p_{CT,RT}$=$7.1 \times 10^{-21}$). Conversely, SI was correlated with CT $\rho_{CT,SI}$=$.05$, $p_{CT,SI}$=$3.4 \times 10^{-3}$). These results show  that both measurements were satisfying the Weber Law (RT decreasing with higher CT and SI increasing with respect CT). Individual results for correlations between each contrast measurement satisfy for most cases the aforementioned relationships between CT and RT as well as CT and SI, presented in \hyperref[table:CT]{Table \ref*{table:CT}}.

\begin{table}[H]
\caption{Table of correlations between contrast values (3rd column) with Reaction Time (4th column), or with Saliency Index (5th column)} 
\footnotesize
\begin{tabular}{ |>{\raggedright}p{3.4cm}|c|c|c| } 
\hline
Feature type & Contrast (CT) & $\rho_{RT}$ & $\rho_{SI}$ \\ 
\hline
(1) Corner Angle & Slope$(º)$ & .23* & .53* \\ 
(2) Segment. Angle & Angle,$\Delta\Phi(º)$ & -.33* & .11 \\ 
(3) Segment. Spacing & Spacing(deg) & .65* & -.37* \\ 
(4) Contour Integration & Length(deg) & -.25* & -.35* \\ 
(5) Perc. Grouping & Distance(deg) & .29* & -.06 \\ 
(6) Feat. \& Conj. Search & Set Size(\#) & .15* & -.12* \\ 
(7) Search Asymmetries & Set Size(\#) & -.33* & -.39* \\ 
(8) Noise/Roughness & Freq., $1/f^\beta$ & -.56* & .53* \\ 
(9) Color Contrast & Sat.,$\Delta S_{D,T}$ & -.57* & .48* \\ 
(10) Brightness Contrast & Light.,$\Delta L_{D,T}$ & -.41* & .25* \\ 
(11) Size Contrast & Size(deg) & -.55* & -.29* \\ 
(12) Orientation Contrast & Angle,$\Delta\Phi(º)$ & -.18* & .05 \\ 
(13) Distr. Heterogeneity & Angle,$\Delta\Phi_{1c}(º)$ & -.04 & .17* \\ 
(14) Distr. Linearity & Angle,$\Delta\Phi(º)$ & -.07 & .01 \\ 
(15) Distr. Categorization & Angle,$\Delta\Phi_{1c}(º)$ & -.24* & .23* \\ 
\hline
\end{tabular}
\\
*: $p$\textless.05
\label{table:CT}
\end{table}

We have plotted the relationships between RT and CT as well as for SI and CT in order to see how CT varies localization performance for each stimulus feature type individually (\hyperref[fig:resFV]{Figures \ref*{fig:resFV},\ref*{fig:resVSa} and \ref*{fig:resVSb}}). In these figures we can observe (in relation to \hyperref[table:CT]{Table \ref*{table:CT}}) which feature targets are perceived in parallel or require a serial 'binding' step.

\begin{figure*}
\centering
\begin{subfigure}[b]{.195\linewidth}
\centering
\includegraphics[width=1\linewidth]{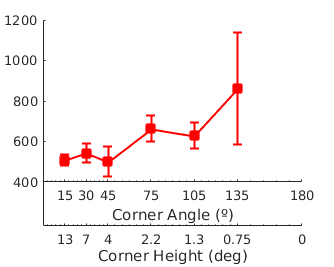} 
\end{subfigure}
\begin{subfigure}[b]{.195\linewidth}
\centering
\includegraphics[width=1\linewidth]{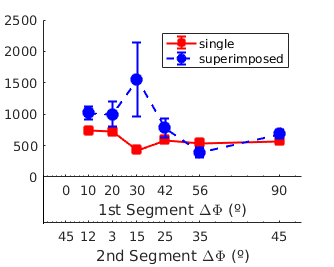} 
\end{subfigure}
\begin{subfigure}[b]{.195\linewidth}
\centering
\includegraphics[width=1\linewidth]{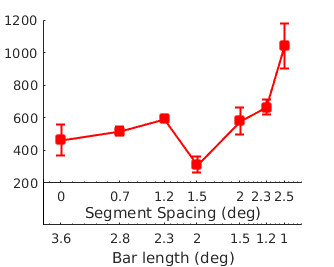} 
\end{subfigure}
\begin{subfigure}[b]{.195\linewidth}
\centering
\includegraphics[width=1\linewidth]{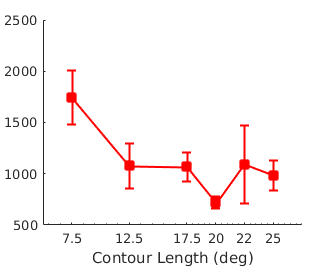} 
\end{subfigure}
\begin{subfigure}[b]{.195\linewidth}
\centering
\includegraphics[width=1\linewidth]{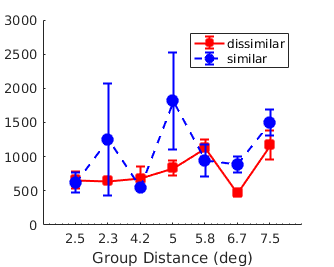} 
\end{subfigure}
\begin{subfigure}[b]{.195\linewidth}
\centering
\includegraphics[width=1\linewidth]{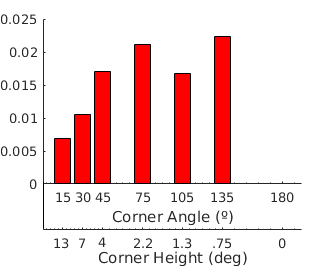} 
\caption*{\centering (1)}
\end{subfigure}
\begin{subfigure}[b]{.195\linewidth}
\centering
\includegraphics[width=1\linewidth]{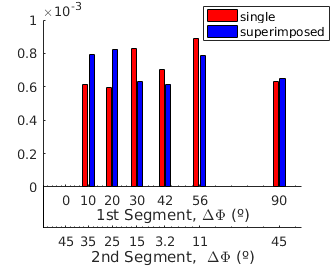} 
\caption*{\centering (2)}
\end{subfigure}
\begin{subfigure}[b]{.195\linewidth}
\centering
\includegraphics[width=1\linewidth]{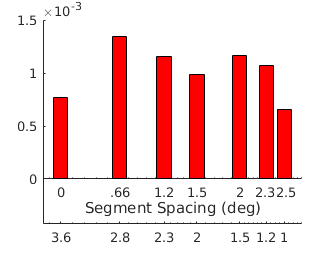} 
\caption*{\centering (3)}
\end{subfigure}
\begin{subfigure}[b]{.195\linewidth}
\centering
\includegraphics[width=1\linewidth]{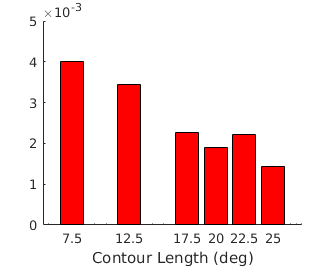} 
\caption*{\centering (4)}
\end{subfigure}
\begin{subfigure}[b]{.195\linewidth}
\centering
\includegraphics[width=1\linewidth]{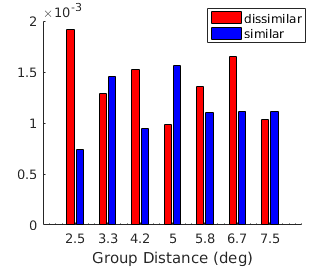} 
\caption*{\centering (5)}
\end{subfigure}
\caption{Plots of Reaction Times (top row) and Saliency Index (bottom row). Spearman's rank correlation tests were performed between RT and SI from each stimulus type and participant individually, corresponding on each case to \textbf{Corner Angle (1)}: $\rho_{(1)}$=$8.3 \times 10^{-2}$, $p_{(1)}$=$.43$, \textbf{Visual Segmentation (2,3)}: $\rho_{(2)}$=$-.22$, $p_{(2)}$=$5.6 \times 10^{-3}$; $\rho_{(3)}$=$-5.7 \times 10^{-4}$, $p_{(3)}$=$.99$, \textbf{Contour Integration (4)}: $\rho_{(4)}$=$-5.1 \times 10^{-2}$, $p_{(4)}$=$.61$ and \textbf{Perceptual Grouping (5)}: $\rho_{(5)}$=$-.13$, $p_{(5)}$=$.12$. For this cases, we have discarded samples in which participants had a fixation closer than 5 degrees of eccentricity from the search target, corresponding to the higher visual acuity of the fovea \cite{Strasburger2011}\cite{wandell1995foundations}, as the RT calculation could be impaired by center biases.}
\label{fig:resFV}
\end{figure*}

\begin{figure*}
\centering
\begin{subfigure}[b]{.195\linewidth}
\centering
\includegraphics[width=1\linewidth]{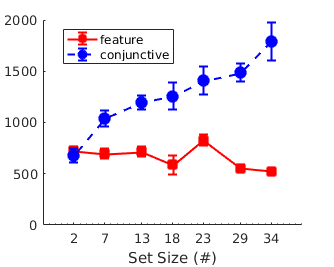} 
\end{subfigure}
\begin{subfigure}[b]{.195\linewidth}
\centering
\includegraphics[width=1\linewidth]{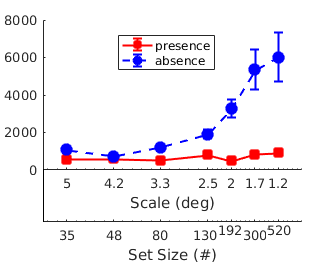} 
\end{subfigure}
\begin{subfigure}[b]{.195\linewidth}
\centering
\includegraphics[width=1\linewidth]{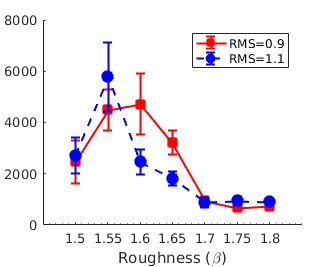}
\end{subfigure}
\begin{subfigure}[b]{.195\linewidth}
\centering
\includegraphics[width=1\linewidth]{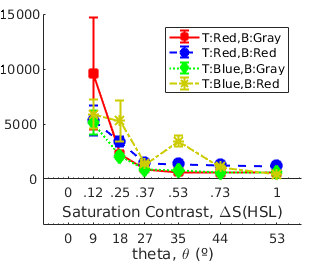} 
\end{subfigure}
\begin{subfigure}[b]{.195\linewidth}
\centering
\includegraphics[width=1\linewidth]{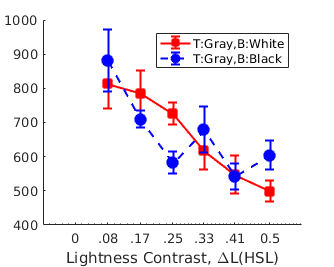} 
\end{subfigure}
\begin{subfigure}[b]{.195\linewidth}
\centering
\includegraphics[width=1\linewidth]{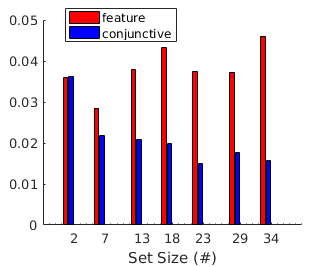} 
\caption*{\centering (6)}
\end{subfigure}
\begin{subfigure}[b]{.195\linewidth}
\centering
\includegraphics[width=1\linewidth]{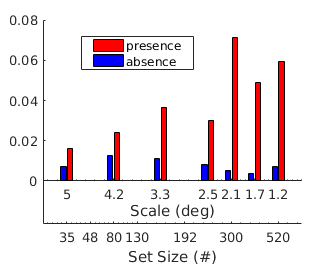} 
\caption*{\centering (7)}
\end{subfigure}
\begin{subfigure}[b]{.195\linewidth}
\centering
\includegraphics[width=1\linewidth]{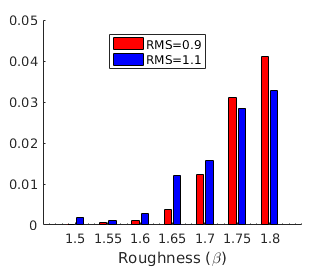} 
\caption*{\centering (8)}
\end{subfigure}
\begin{subfigure}[b]{.195\linewidth}
\centering
\includegraphics[width=1\linewidth]{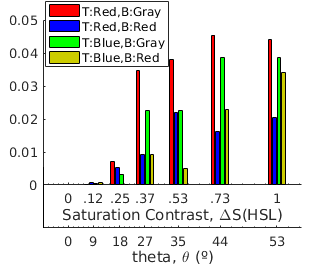} 
\caption*{\centering (9)}
\end{subfigure}
\begin{subfigure}[b]{.195\linewidth}
\centering
\includegraphics[width=1\linewidth]{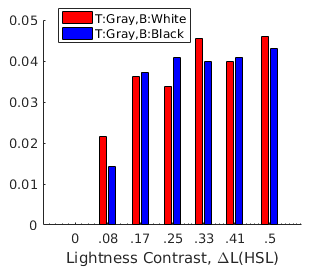} 
\caption*{\centering (10)}
\end{subfigure}
\\\vspace{2mm}
\caption{Plots of Reaction Times (top row) and Saliency Index (bottom row). Spearman's rank correlation tests were performed between RT and SI from each stimulus type and participant individually, corresponding on each case to \textbf{Feature and Conjunction search (6)}: $\rho_{(6)}$=$-.59$, $p_{(6)}$=$4.6 \times 10^{-36}$, \textbf{Search Asymmetries (7)}: $\rho_{(7)}$=$-.45$, $p_{(7)}$=$3.3 \times 10^{-9}$, \textbf{Noise/Roughness (8)}: $\rho_{(8)}$=$-.68$, $p_{(8)}$=$5.5 \times 10^{-33}$, \textbf{Color Contrast (9)}: $\rho_{(9)}$=$-.69$, $p_{(9)}$=$1.5 \times 10^{-72}$ and \textbf{Brightness Contrast (10)}: $\rho_{(10)}$=$-.51$, $p_{(10)}$=$3.4 \times 10^{-23}$.}
\label{fig:resVSa}
\end{figure*}

\begin{figure*}
\centering
\begin{subfigure}[b]{.195\linewidth}
\centering
\includegraphics[width=1\linewidth]{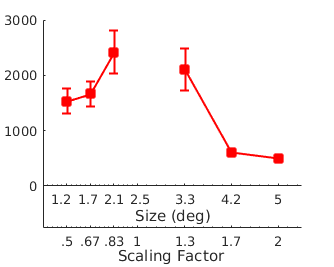} 
\end{subfigure}
\begin{subfigure}[b]{.195\linewidth}
\centering
\includegraphics[width=1\linewidth]{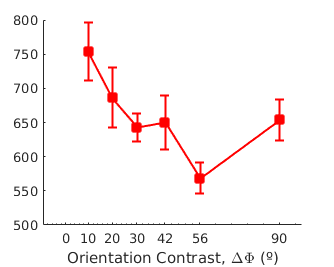} 
\end{subfigure}
\begin{subfigure}[b]{.195\linewidth}
\centering
\includegraphics[width=1\linewidth]{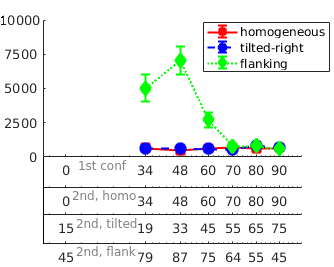} 
\end{subfigure}
\begin{subfigure}[b]{.195\linewidth}
\centering
\includegraphics[width=1\linewidth]{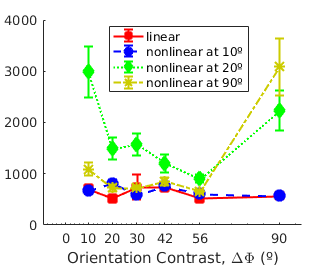} 
\end{subfigure}
\begin{subfigure}[b]{.195\linewidth}
\centering
\includegraphics[width=1\linewidth]{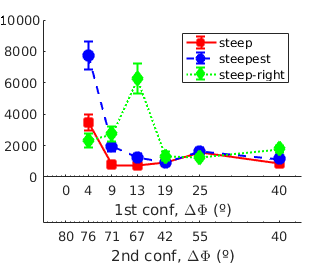} 
\end{subfigure}
\begin{subfigure}[b]{.195\linewidth}
\centering
\includegraphics[width=1\linewidth]{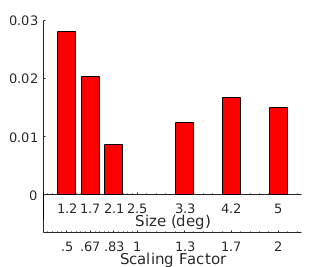} 
\caption*{\centering (11)}
\end{subfigure}
\begin{subfigure}[b]{.195\linewidth}
\centering
\includegraphics[width=1\linewidth]{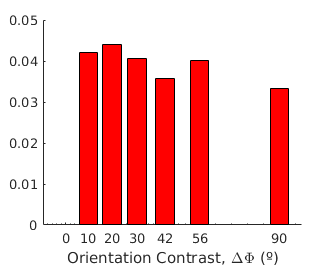} 
\caption*{\centering (12)}
\end{subfigure}
\begin{subfigure}[b]{.195\linewidth}
\centering
\includegraphics[width=1\linewidth]{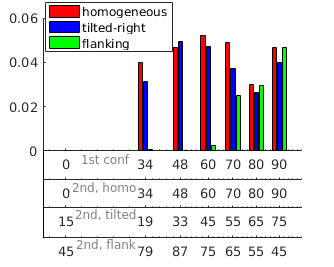} 
\caption*{\centering (13)}
\end{subfigure}
\begin{subfigure}[b]{.195\linewidth}
\centering
\includegraphics[width=1\linewidth]{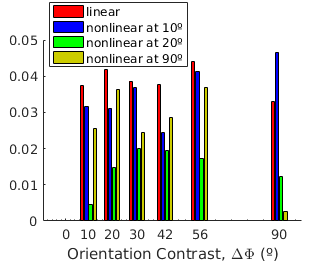} 
\caption*{\centering (14)}
\end{subfigure}
\begin{subfigure}[b]{.195\linewidth}
\centering
\includegraphics[width=1\linewidth]{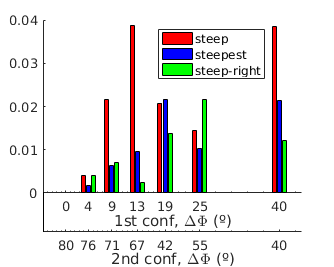} 
\caption*{\centering (15)}
\end{subfigure}
\\\vspace{2mm}
\caption{Plots of Reaction Times (top row) and Saliency Index (bottom row). Spearman's rank correlation tests were performed between RT and SI from each stimulus type and participant individually, corresponding on each case to \textbf{Size Contrast (11)}: $\rho_{(11)}$=$-.14$, $p_{(11)}$=$9.7 \times 10^{-2}$, \textbf{Orientation Contrast (12)}: $\rho_{(12)}$=$-.41$, $p_{(12)}$=$2.7 \times 10^{-8}$, \textbf{Distractor Heterogeneity (13)}: $\rho_{(13)}$=$-.57$, $p_{(13)}$=$3.5 \times 10^{-41}$ and \textbf{Distractor Linearity (14)}: $\rho_{(14)}$=$-.57$, $p_{(14)}$=$2.1 \times 10^{-53}$ and \textbf{Distractor Categorization (15)}: $\rho_{(15)}$=$-.66$, $p_{(15)}$=$2.9 \times 10^{-59}$.}
\label{fig:resVSb}
\end{figure*}

On \hyperref[results:resFV]{(1-5)} the Weber law applies for stimulus such as Corner Angle, showing slower localization on smoother corners than sharper ones. For Visual Segmentation stimuli, segment localization was faster to be localized when segment angle had a diagonal segment for both single and superimposed segments (due to its own corner angle with respect to other segment bars), being single ones with a trend to be more salient ($p$=$1.2 \times 10^{-2}$, $\tilde{\chi}^2$=$6.3$). Segments with \SI{1.5}{\deg} of segment distance and \SI{2.5}{\deg} of bar length showed faster localization rate compared to wider segments. The Weber law applied as well for contour detection, being larger contours faster to be localized. For Perceptual Grouping, similar shape distractors showed slower localization rates as grouping distance is increased (lower proximity), but it was not so evident for dissimilar distractors, being localized faster and with overall higher SI. The Weber law did not apply for this case, suggesting that at a certain proximity distance (about approximately \SI{5.5}{\deg}) participants fixated into several regions, making them similarly salient. SI results on Corner Angle and Contour Integration had positive correlations with respect RT (contradicting the general case). That would be caused by the size of the masks (from stimulus salient objects), which would be higher for higher stimulus contrasts, with decreasing absolute SI (bigger masks would require more fixations when considering the same spatial conditions). In that aspect, SI must be evaluated considering that the size of the mask is constant, which is not the case for Corner Angle (1) and Contour Integration (4). These center biases might be one of the reasons for the Weber law appliance (presenting less agreement on RT and SI continuity upon feature contrast) as endogenous visual guidance can generate higher inter-participant differences. A more continuous slope for RT and SI observed for stimulus feature contrasts could be acquired by using an onset cue and a constant distance between the initial fixation and the stimulus target, but that method could generate oculomotor biases with respect to the possible positions distinct from the center (that could also vary the temporality of the fixations with respect to the center distance). An alternative solution that would partly solve the problem (as distance from the initial fixation and the stimulus target would still not be totally constant) would be to acquire a larger amount of observations at distinct randomized regions for each stimulus contrast and stimulus type \cite{Wolfe2010b}\cite{Wolfe2011}. 

Feature search show a faster localization of the target than conjunctive search (\hyperref[fig:resVSa]{Figure \ref*{fig:resVSa}}), with an almost constant RT with respect to set size (features processed in parallel). Conjunction search reveal slower localization of stimulus targets ($p$=$2.5 \times 10^{-24}$, $\tilde{\chi}^2$=$104$) as we increase distractor number (consequently, features being shown to be processed in a serial manner), likewise with lower SI. Similarly, reporting stimulus absence presented similar response time distributions, presenting conjunctive distractors to be more uncertain for reporting absence ($p$=$1.9 \times 10^{-36}$, $\tilde{\chi}^2$=$159$) than feature search ones. Searching a target circle among circle distractors with a superimposed bar show lower performance at increasing scale and set size ($p$=$6.2 \times 10^{-33}$, $\tilde{\chi}^2$=$143$), reversely, searching a target circle with a superimposed bar among circles shows more constant performance, revealing that search asymmetries for this case apply. SI also reveals search asymmetries with respect these two types of stimuli, however, the SI is lower for the former case. 

The Weber law is present for the case of background roughness, showing a decrease in search performance and SI at low beta values (rougher surfaces). Both conditions of height deviation  ($\sigma_{RMS}$=$0.9, 1.1$) present similar performance with both metrics, with a trend of better search efficiency for higher RMS values. When searching a target with higher saturation contrast with respect distractors, both search performance and SI is higher than with lower saturation contrasts. Background conditions present a trend to drive search asymmetries, showing faster localization RTs and SI for unsaturated backgrounds. In that aspect, achromatic backgrounds presented faster localization for both red ($p$=$3.8 \times 10 ^{-9}$) and blue distractors ($p$=$2.1 \times 10^{-4}$). SI is shown to be higher for red hue in contrast to blue hue for search target and distractors. Lightness contrast also conforms with the Weber law, similarly to saturation contrast but with higher overall performance. Lighter backgrounds with darker search targets present a trend to have higher SI with respect to darker backgrounds with lighter salient objects.

Results on size similarity reveal increased search efficiency with respect to size contrast, with a tendency of perceiving bigger objects as more salient than smaller ones for both localization time and saliency index as in \hyperref[fig:resVSb]{Figure \ref*{fig:resVSb}}. Similarity on orientation also shows increased search efficiency with respect angle contrast, with diagonal angles localized faster than vertical or horizontal ones. Orientation contrast has been found to have high search efficiency, specially with diagonal angles and homogeneous angle organization for distractors. In contrast, heterogeneous set of distractor angles present a lower search efficiency with respect to the homogeneous ones. Homogeneous distractors were significantly localized faster for heterogeneity at distinct angle quadrant configurations (flanking) $p$=$1.0 \times 10^{-9}$ but not for heterogeneous distractors with angle configurations at the same quadrant (tilted-right) $p$=$.63$. Another distinct type of orientation-related guidance is distractor linearity, presenting differences depending on each slope condition ($p$=$1.3 \times 10^{-35}$, $\tilde{\chi}^2$=$165$). Non-linear orientation patterns at a slope increment of \SI{20}{\degree} present lower search efficiency than the ones at \SI{10}{\degree} and \SI{90}{\degree}. The latter case presents a slightly lower search efficiency at vertical or horizontal orientations due to its similar orientation interactions between the target and one of the distractor sets. Results suggest that both the amount of distractor sets and each of their orientation contrasts with respect to search target might be the source of overall distinctiveness for non-linear orientation patterns. Results related with orientation pattern categorization report overall higher SI and a trend for faster localization rate for steep orientation organization than steepest ($p$=$8.4 \times 10^{-2}$) and significant with respect to steep-right ($p$=$1.2 \times 10^{-5}$), confirming that search asymmetries apply for this case considering that three conditions possess the same orientation contrast between the two distractor sets.

\subsubsection*{Discussion}

In this study we show that feature contrast is correlated to saliency using distinct measures and feature types, being saliency higher at higher feature contrasts. By using visual search tasks and synthetic images, there is a better control of exogenous cues by reducing endogenously-dependent guidance. It is about to consider that the SI is a good measure for evaluating saliency for specific areas of interest.

\subsection{Attention changes nonlinearly over time (3rd Hypothesis)} \label{results3}

Values of FD and SA were grouped for each gaze as functions of viewing time. In \hyperref[fig:results1_evol]{Figure \ref*{fig:results1_evol}\textbf{(a,b)}}, during the first 1 to 2 seconds, fixations have a larger duration for visual search tasks. For the visual search task, fixations have a duration with a peak at \SI{274}{\milli\second} during the beginning of the experiment and progressively drop during the end of the stimulus view to $217$ after \SI{5000}{\milli\second} of viewing time. In free-viewing tasks, FD remains stable after the first and second fixation at approximately \SI{202}{\milli\second}. For the SA on both tasks there is a peak for the first saccade between $6.5$ and \SI{7}{\deg}. During the first and second gaze, SA drops to a value between $5.5$ and \SI{6}{\deg} and increase during 1 second to amplitudes between approximately $6$ and \SI{6.5}{\deg}. During the last gazes, after 2 seconds of viewing time, SA progressively drops during the rest of viewing time. Such behavior occurs similarly for both visual search and free-viewing cases, these patterns might also be related to endogenous factors commented previously. These distinct eye movement patterns might be related to how participants approach targets depending on task priors and show an overview of how relevant is to account for temporal properties when evaluating eye movements. 

SI was computed using the density maps across fixation number \hyperref[fig:results1_evol]{Figure \ref*{fig:results1_evol}\textbf{(c)}}, it decreases with respect to fixation number, being the first fixations (from the 1st to the 5th) the ones that have higher SI (accounting for fixations inside the salient region). Inhibition of return (IOR) mechanisms might be responsible for the aforementioned effects. IOR was present and we believe that it may have influenced both types of tasks. To know that, mean return saccade time was computed, corresponding to the time spent from the first fixation inside the AOI to the second fixation that returned inside the AOI, which was $M$=$\num[separate-uncertainty]{16.6(9)}\times 10^{2}$ ms, corresponding to $M$=$\num[separate-uncertainty]{14.1(5)}\times 10^{2}$ ms for Free-Viewing and $M$=$\num[separate-uncertainty]{18.6(15)}\times 10^{2}$ms for Visual Search tasks respectively. 


\begin{figure}[H]
 \centering
\begin{subfigure}{0.35\textwidth}
\includegraphics[width=1\linewidth]{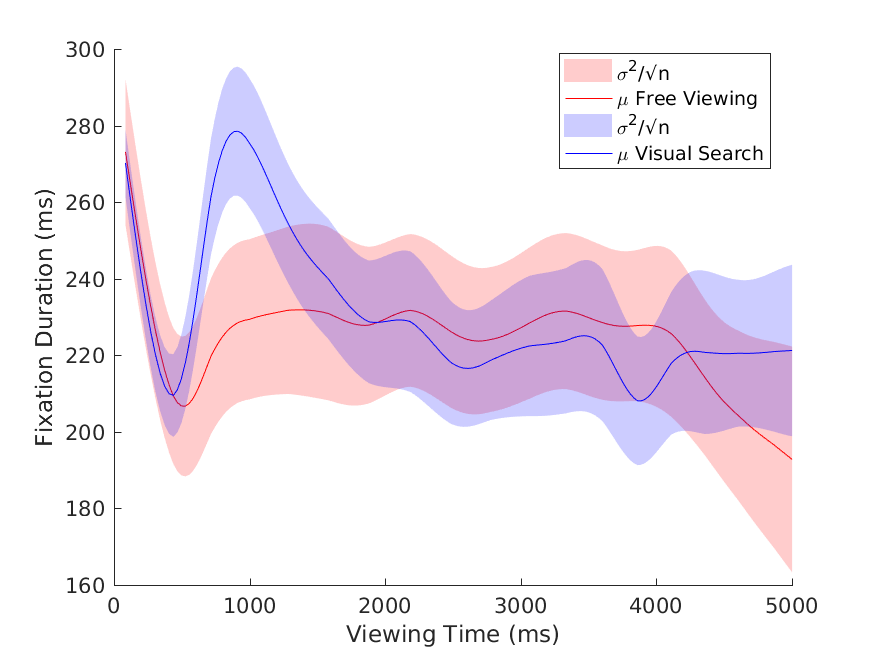} 
\caption{\centering}
\end{subfigure}
\begin{subfigure}{0.35\textwidth}
\includegraphics[width=1\linewidth]{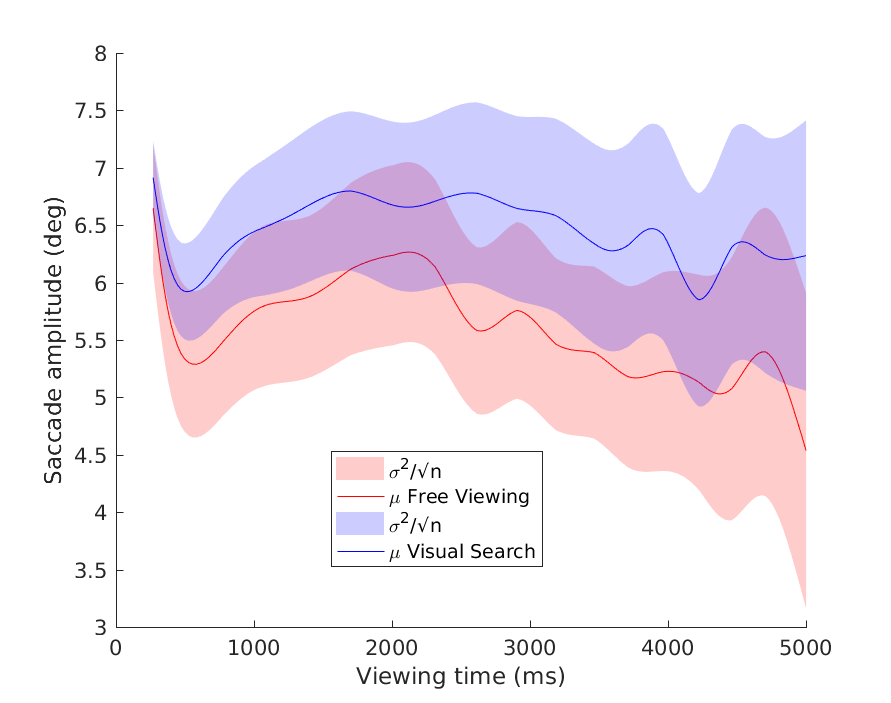} 
\caption{\centering}
\end{subfigure}
\begin{subfigure}{0.35\textwidth}
\includegraphics[width=1\linewidth]{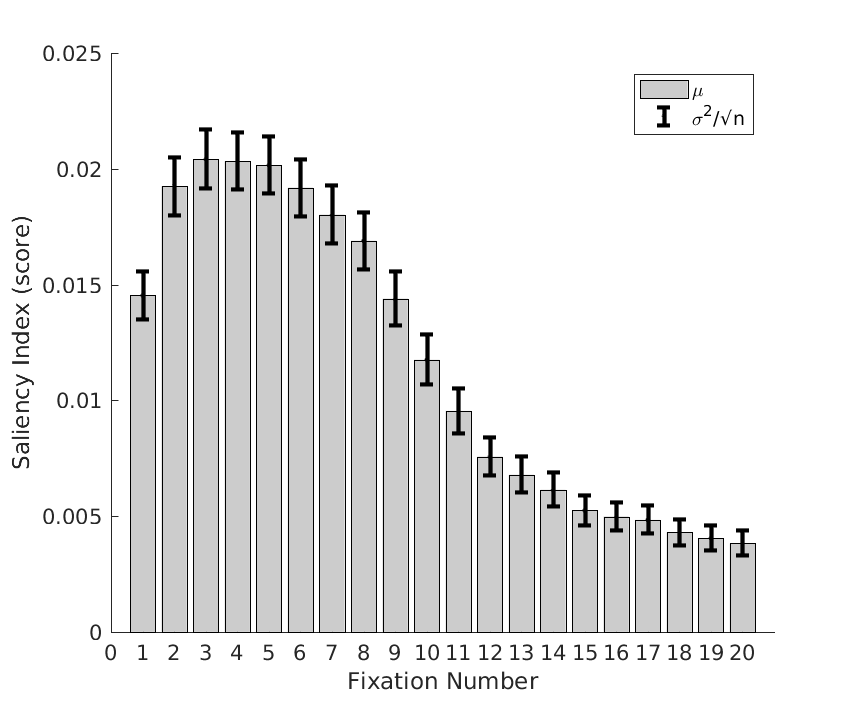} 
\caption{\centering}
\end{subfigure}
\caption{\textbf{(a)} Temporal evolution (from $0$ to \SI{5000}{\milli\second}) of fixation duration. \textbf{(b)} Temporal evolution (from $0$ to \SI{5000}{\milli\second}) of saccade amplitude. For both plots, samples corresponding to free-viewing task fixations and saccades are represented in red and blue for the case of Visual Search. \textbf{(c)} Mean saliency index upon fixation number.
}
\label{fig:results1_evol}
\end{figure}

\subsubsection*{Discussion}

The temporal evolution of fixation and saccade behavior reveal distinct patterns of eye movements upon viewing time , confirming the evidence that visual attention is an active process and its modeling involving temporality requires further investigation. Scanpath prediction could allow the reproduction of the aforementioned effects, regarding in that aspect both bottom-up and top-down processing of visual features that distinctively guide visual attention \cite{Boccignone2004}\cite{Kubota2012}\cite{Chang2014}\cite{LeMeur2015}\cite{Aboudib2015}\cite{Adeli2016}\cite{Wang2016}\cite{Wloka2017}\cite{White2017a}. In that aspect, as saliency decreases over time, saliency evaluation measures should be done in that line.

\subsection{Task influences perceived attention \\ (4th Hypothesis)} \label{results4}

Distinct eye movement behavior in terms of FD and SA was presented depending on each task type (\hyperref[fig:results3]{Section \ref*{results3}}). Task priors also influenced the localization performance in relation to feature contrast. 

First, Wilcoxon signed-rank tests were performed to evaluate the amount of fixations between easy and hard targets and was found to be lower for easy than for hard targets in the visual search task ($p$=$2.1 \times 10-147$, Z=$-26$, $Mdn_{easy}$=$4$, $Mdn_{hard}$=$7$), but there was no difference for the case of the free-viewing task (p=$.069$, Z=$-0.1$, $Mdn_{easy}$=$15$, $Mdn_{hard}$=$16$). There were differences in FD between the easy and hard targets for visual search ($p$=$3.6 \times 10^{-36}$, Z=$13$, $Mdn_{easy}$=$199$, $Mdn_{hard}$=$179$ ms), but it was not occurring for free-viewing tasks ($p$=$.57$, Z=$.57$, $Mdn_{easy}$=$199$, $Mdn_{hard}$=$199$ ms). Same phenomena was presented for SA, in which there was a significant difference depending on the stimulus contrast difficulty for visual search ($p$=$1.7 \times 10^{-56}$, Z=$-16$, $Mdn_{easy}$=$3.9$, $Mdn_{hard}$=\SI{4.7}{\deg}) but not for the case of free-viewing ($p$=$.069$, Z=$-1.8$, $Mdn_{easy}$=$3.9$, $Mdn_{hard}$=\SI{4.1}{\deg}). These results evidence less dependence from low-level feature contrasts for free-viewing tasks in contrast to visual search tasks, acknowledging that participants are not always exogenously guided to gaze towards salient regions for free-viewing tasks, namely, that endogenous factors are prevailing more in this kind of task, making saliency less accurate spatially and temporally.

Second, we observed the correlations of RT, SI and feature contrast (FC), described in \hyperref[results2]{Section \ref*{results2}}. Here we define FC as $\psi$ values for considering a generalized feature contrast, as CT values vary between blocks, but FC values do not. For visual search stimuli, RT was negatively correlated with SI ($\rho_{RT,SI}$=$-.59$, $p_{RT,SI}$=$.00$), FC was negatively correlated with RT ($\rho_{FC,FC}$=$-.08$, $p_{FC,RT}$=$6.4\times 10^{-7}
$) but positively correlated with SI ($\rho_{FC,SI}$=$.05$, $p_{FC,SI}$=$2.4\times 10^{-3}$). For free-viewing stimuli, there was a significant negative correlation between RT and SI ($\rho_{RT,SI}$=$-.16$, $p_{RT,SI}$=$1.2 \times 10^{-4}$), a negative correlation between FC and RT ($\rho_{FC,RT}$=$.26$, $p_{FC,RT}$=$6.5 \times 10^{-10}$) but the relationship with respect feature contrast and SI was non-significant ($\rho_{FC,SI}$=$-.04$, $p_{FC,SI}$=$.32$). Distinct behavior is presented on the regression lines shown in \hyperref[fig:results4comp]{Figure \ref*{fig:results4comp}} the relationships from RT and SI with respect to CT (here represented as a unique contrast value, although calculated for each contrast measurement separately as in \hyperref[table:CT]{Table \ref*{table:CT}}) for both tasks.

    
    

\begin{figure}[H]
 \centering
\begin{subfigure}{0.35\textwidth}
\includegraphics[width=1\linewidth]{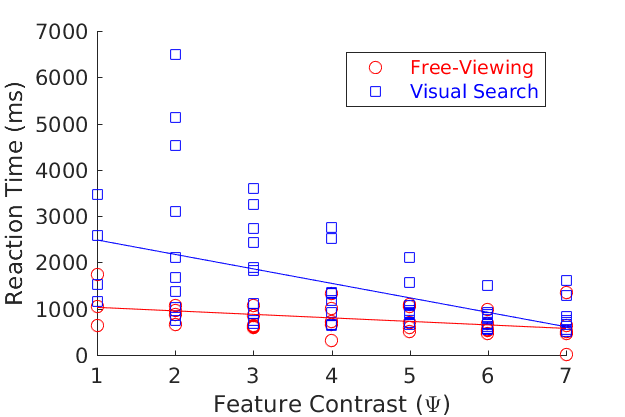} 
\caption{\centering}
\end{subfigure}
\begin{subfigure}{0.35\textwidth}
\includegraphics[width=1\linewidth]{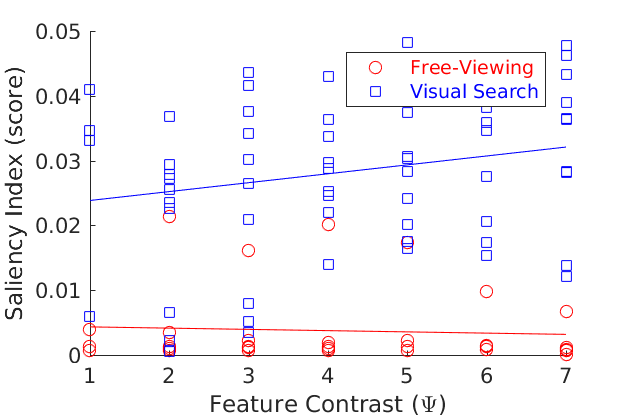} 
\caption{\centering}
\end{subfigure}
\begin{subfigure}{0.35\textwidth}
\includegraphics[width=1\linewidth]{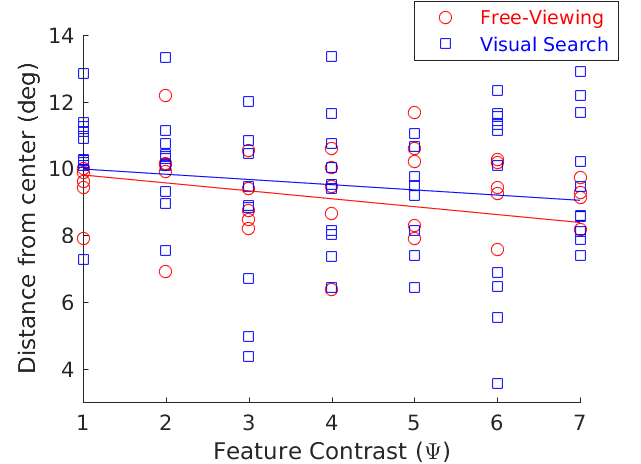} 
\caption{\centering}
\end{subfigure}
\caption{Scatter plots of Reaction Time \textbf{(a)}, Saliency Index \textbf{(b)} and Distance from center \textbf{(c)} upon feature contrast ($\Psi$). We represented the mean of each feature type separately and we have plotted the regression line for both tasks. }
\label{fig:results4comp}
\end{figure}

\subsubsection*{Discussion}

Salient region localization performance varies with respect to feature contrast depending on the task. Fixation duration and saccade amplitude are affected more by stimulus contrast on Visual Search than Free-Viewing tasks. Moreover, the center bias seems to be more present for Free-Viewing tasks. Further analysis of interest would be the evaluation of absolute task differences in localization performance. In that respect, we could present the same stimuli with several observations for each feature contrast and distinct cueing, so that to see the absolute influences from endogenous guidance for each distinct feature type and contrast.

\subsection{Center biases are endogenous \\ (5th Hypothesis)} \label{results5}

The center bias was represented by grouping fixations for all samples and representing the density map shown in \hyperref[fig:results2_centerbias]{Figure \ref*{fig:results2_centerbias}}. From such baseline, it is possible to estimate the mean euclidean distance from every fixation to the baseline center (DC). This baseline shows increasing spreadity and area with respect to fixation number and consequently with respect time. In \hyperref[fig:results2_evol]{Figure \ref*{fig:results2_evol}} there is the DC as a function of viewing time (centroid was computed as a unique point corresponding to the initial fixation baseline). From this plot, we can observe that participants move their eyes away from the center of the stimulus after the first and second fixation, between 10 and 11 deg. After 2 seconds of viewing time, mean distance from baseline center is nearly constant for the visual search case but not for the free-viewing case. For the latter, fixations get become closer to the baseline center showing increasing patterns of center bias for this task, similarly to SI (\hyperref[fig:results1_evol]{Figure \ref*{fig:results1_evol}}), which increases during the first fixations and drops on late fixations. 

\begin{figure}[H]
 \centering
\begin{subfigure}{0.4\textwidth}
\includegraphics[width=1\linewidth]{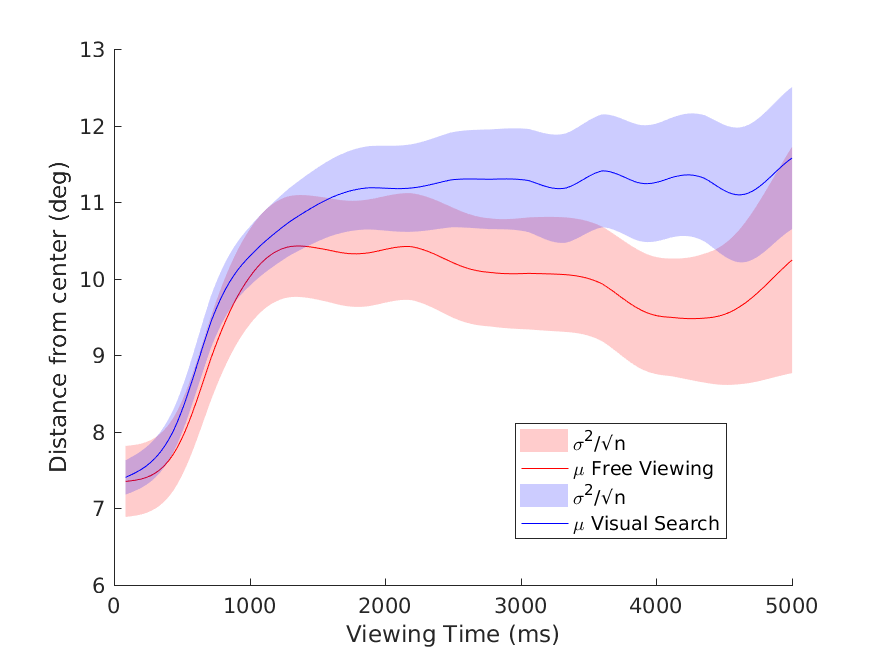} 
\end{subfigure}
\caption{Representation of the center bias as the mean euclidean distance between fixation localization and the baseline center. }
\label{fig:results2_evol}
\end{figure}

\begin{figure}[H]
 \centering
\begin{subfigure}{.5\textwidth}
 \centering
\begin{subfigure}{0.16\linewidth}
\includegraphics[width=1\linewidth]{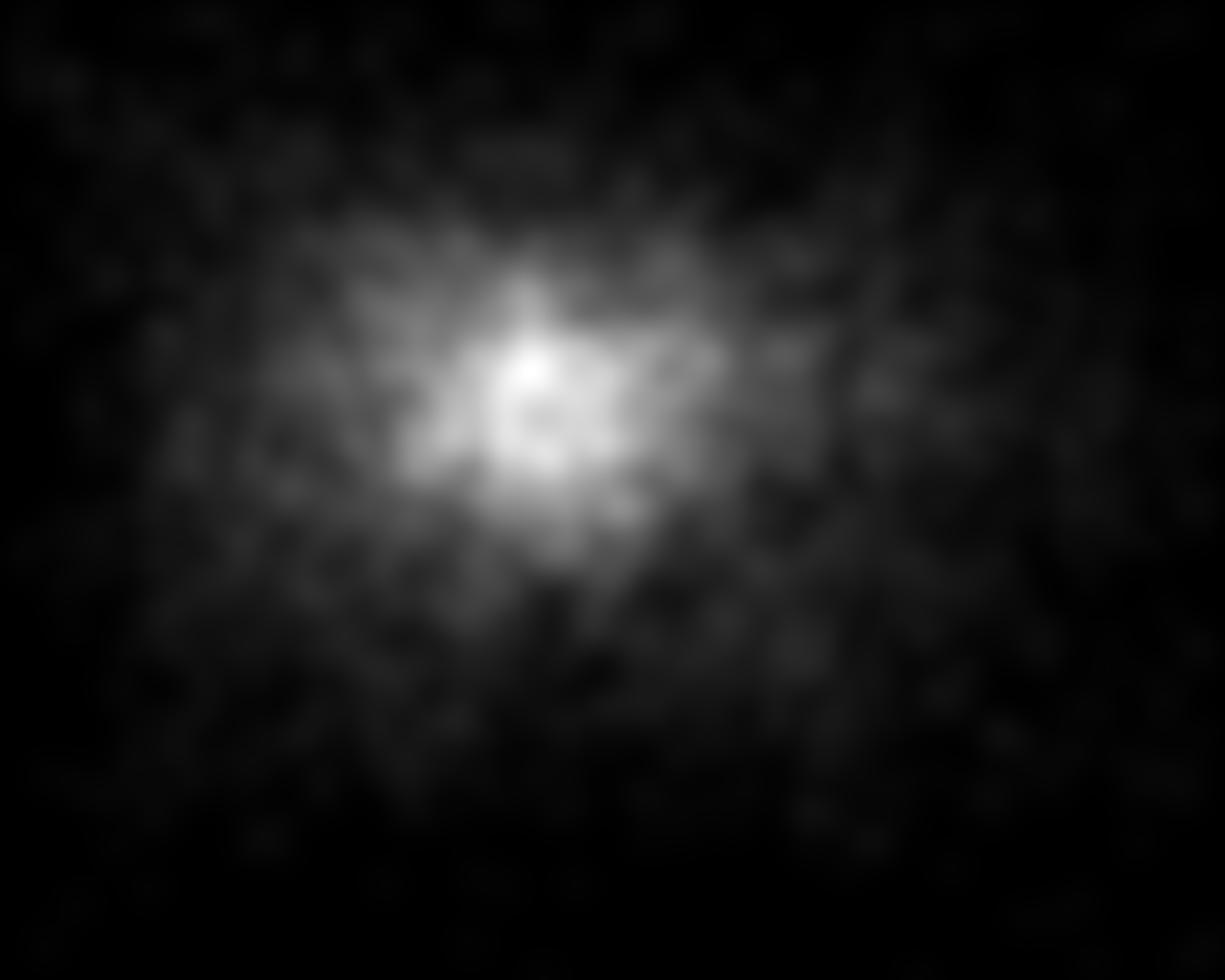} 
\caption*{\centering \textbf{Initial}}
\end{subfigure}
\begin{subfigure}{0.16\linewidth}
\includegraphics[width=1\linewidth]{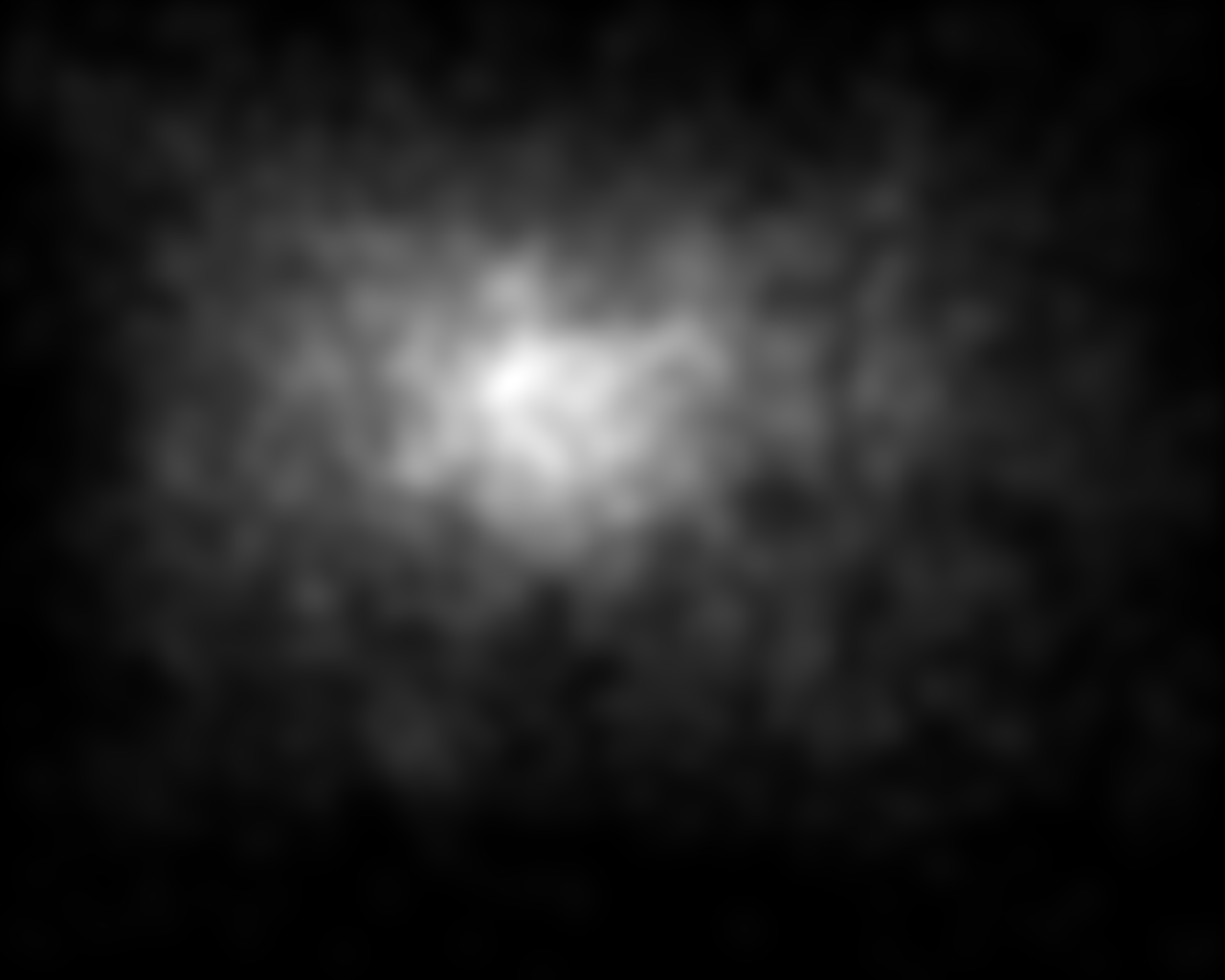} 
\caption*{\centering \textbf{1st}}
\end{subfigure}
\begin{subfigure}{0.16\linewidth}
\includegraphics[width=1\linewidth]{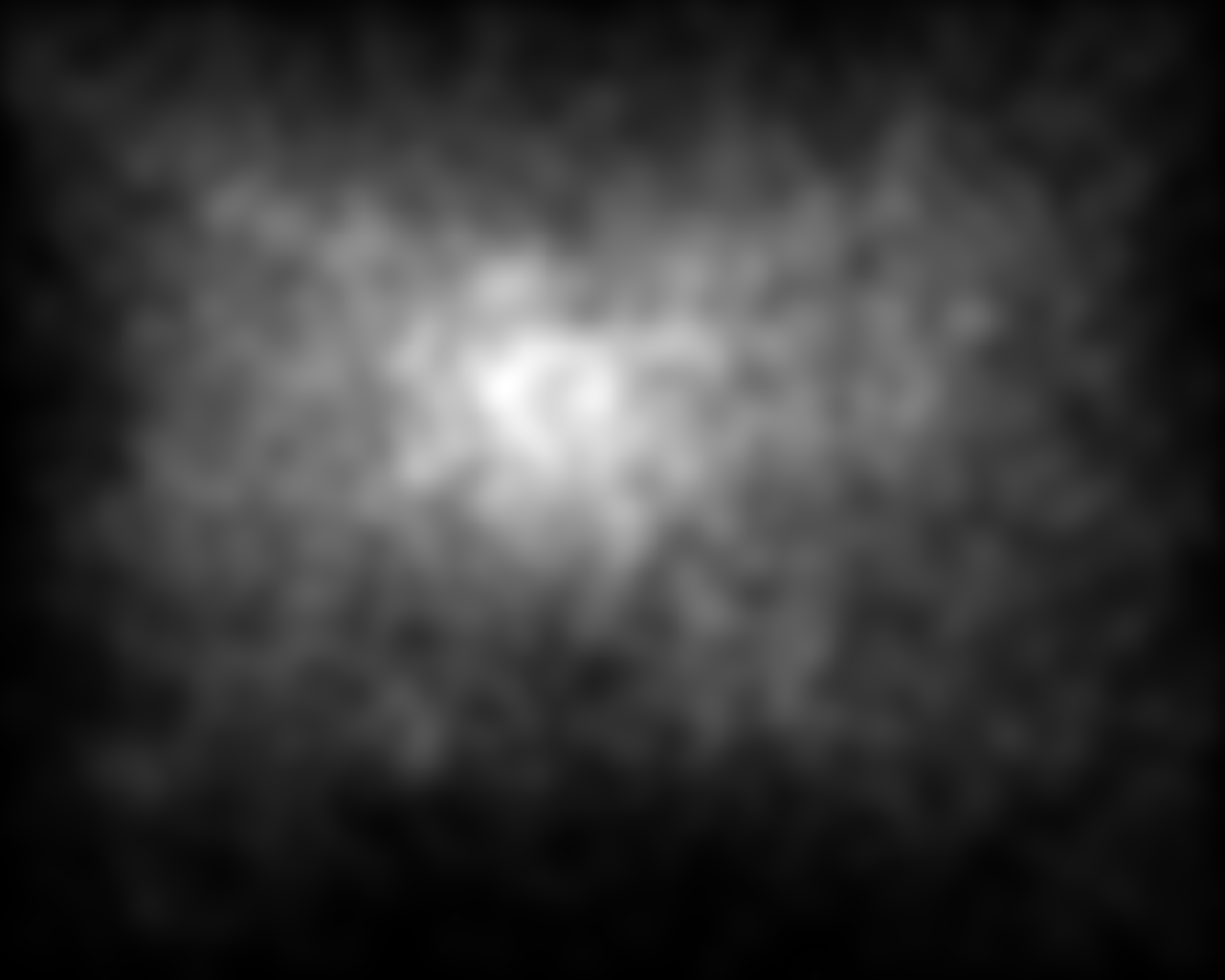} 
\caption*{\centering \textbf{5th}}
\end{subfigure}
\begin{subfigure}{0.16\linewidth}
\includegraphics[width=1\linewidth]
{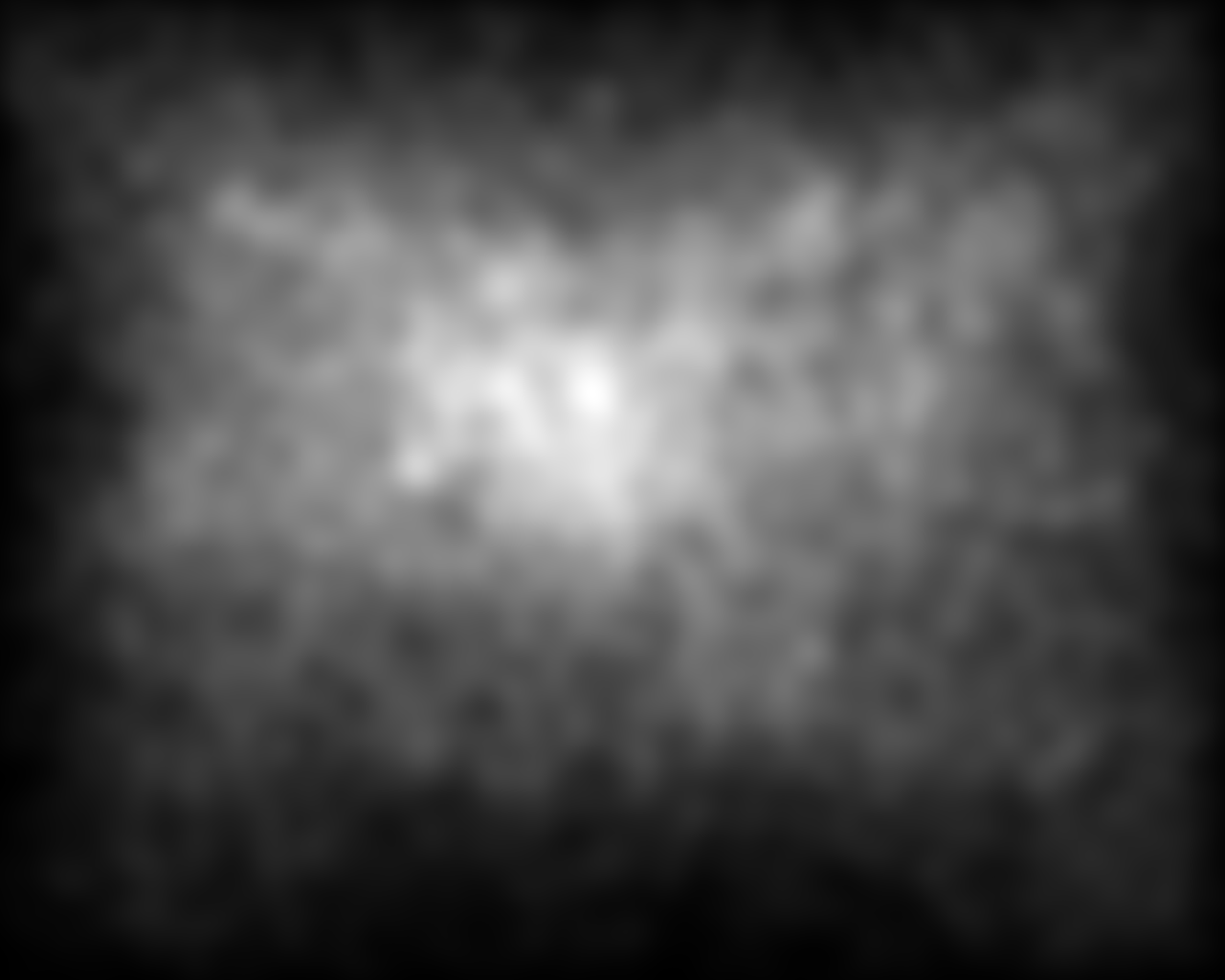} 
\caption*{\centering \textbf{10th}}
\end{subfigure}
\begin{subfigure}{0.16\linewidth}
\includegraphics[width=1\linewidth]{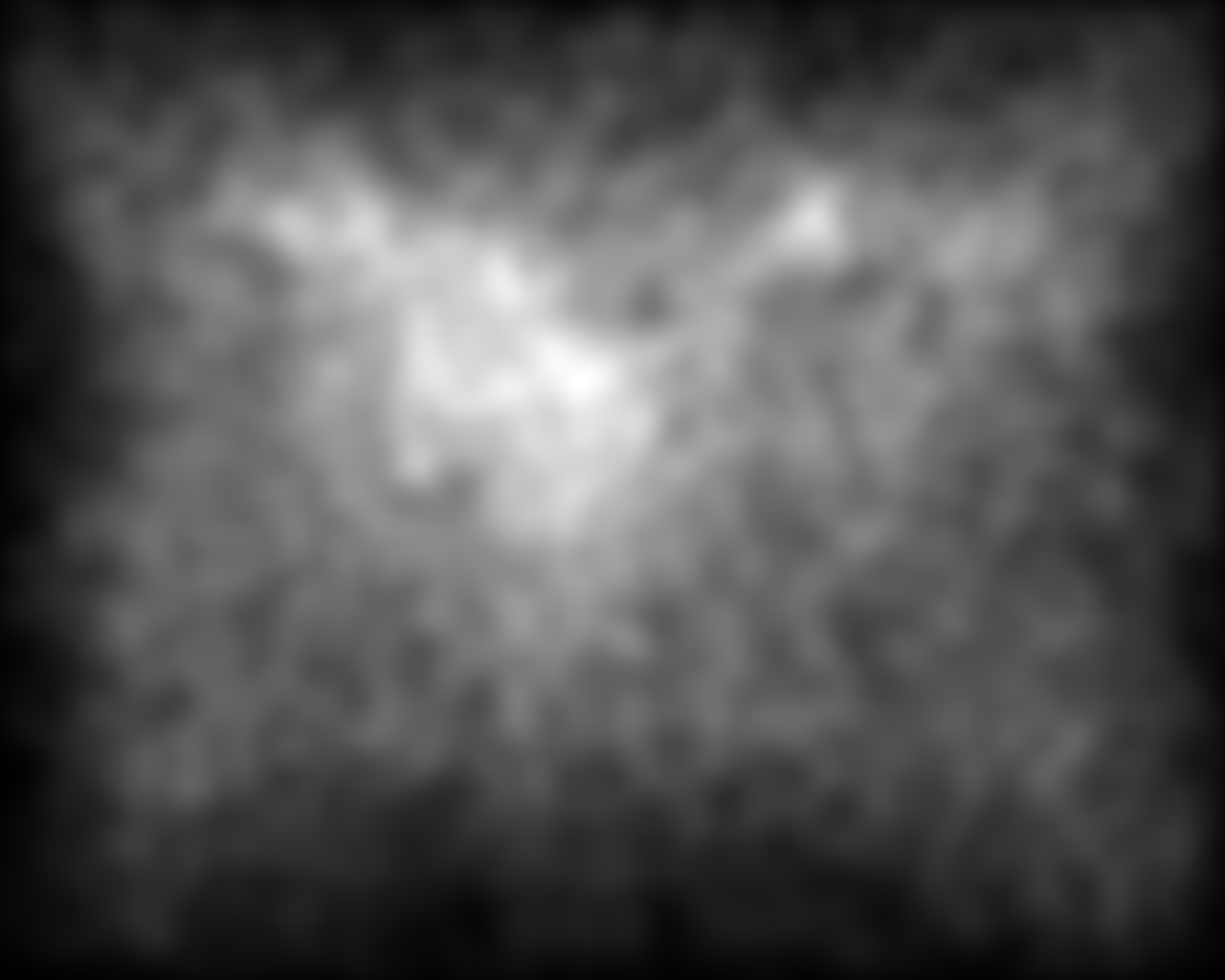} 
\caption*{\centering \textbf{20th}}
\end{subfigure}
\end{subfigure}
\caption{Representation of the density map for all fixations grouped together across all stimuli.}
\label{fig:results2_centerbias}
\end{figure}

DC was negatively correlated with FD ($\rho$=$-.08$, $p$=$3.9 \times 10^{-134}$) and positively correlated with SA ($\rho$=$.08$, $p$=$1.4 \times 10^{-148}$). Here, short fixations and large saccades might be eye movement patterns highly related to saliency (as being negatively correlated to the center bias). By computing the mean per stimulus for the case of FC, it is possible to compare how DC was affected by feature contrast \hyperref[fig:results4comp]{Figure \ref*{fig:results4comp}c}. For Free-Viewing task, DC was significantly negatively correlated with FC ($\rho$=$-.13$, $p$=$2.6 \times 10^{-3}$). For Visual Search task, DC was not significantly correlated with FC ($\rho$=$-.03$, $p$=$.07$). Acknowledging that stimulus targets were randomized, feature contrast was decreasing the center bias (increasing DC) more on Visual Search tasks than for Free-Viewing tasks, supporting the literature stated in \hyperref[biases_relevance]{Section \ref*{biases_relevance}}.




\begin{table}[H]
\centering
\caption{Table of correlations between Feature Contrast (FC) with Distance from baseline center (DC)} 
\footnotesize
\begin{tabular}{ |>{\raggedright}p{3.4cm}|c| } 
\hline
Feature type & $\rho_{\psi,DC}$ \\ 
\hline
(1) Corner Angle  & -.004 \\ 
(2) Segment. Angle  & -.32* \\ 
(3) Segment. Spacing & -.16 \\ 
(4) Contour Integration  & .012 \\ 
(5) Perc. Grouping & -.07 \\ 
(6) Feat. \& Conj. Search & -.34* \\ 
(7) Search Asymmetries  & .24* \\ 
(8) Noise/Roughness   & -.29* \\ 
(9) Color Contrast  & -.12* \\ 
(10) Brightness Contrast & .66* \\ 
(11) Size Contrast  & -.31* \\ 
(12) Orientation Contrast & -.18* \\ 
(13) Distr. Heterogeneity  & .21* \\ 
(14) Distr. Linearity  & -.28* \\ 
(15) Distr. Categorization & -.03 \\ 
\hline
\end{tabular}
\\
*: $p$\textless.05
\label{table:DC}
\end{table}

We have added in \hyperref[table:DC]{Table \ref*{table:DC}} the correlations between DC and FC for each feature individually. Most cases of singleton search (i.e. 6-15) show a significant negative correlation between DC and FC, meaning, when the feature contrast is higher, the center bias is lower.

\subsubsection*{Discussion}

Short saccades and large fixation durations are shown to be correlated with eye movement behavior related to the center bias. Temporality of fixations show a non-linear evolution of the center bias, showing more dispersion with respect to viewing time. Moreover, distance from center In that aspect, saliency would not only need to be evaluated by adjusting metric performances using metrics that account for the aforementioned center biases \cite{Zhang2008}\cite{Borji2013d}\cite{Kummerer2014}\cite{Wloka2016}\cite{Nuthmann2017}, but also upon the importance of temporality on fixation and saccade characteristics, by computing each metric upon gaze number on each stimulus fixational data. Thus, saliency metrics should account for feature contrast and minimize the contextual effects in order to accurately reproduce eye movement behavior.

\section{General Discussion} \label{discussion}

Given the presented results, we emphasize that saliency is influenced by a variety of factors when observing eye movement behavior. In this study is presented a dataset considering all the aforementioned factors, by evaluating eye movements for distinct feature types, contrasts, temporality, task and representing the center biases. First, scene context (here defined as different feature types) is known to affect attention with specific performance, significantly determining efficiency of localizing and/or identifying salient regions. Second, saliency measures are shown to be correlated to feature contrast and distinctively depending on feature type. Third, fixation and saccade characteristics are presented to evolve non-linearly over time, making saliency decrease with respect saccade number and/or viewing time. Fourth, visual search tasks show higher performance in comparison to free-viewing on our saliency measurements and they have a higher correlation with respect saliency and feature contrast. Fifth, the central bias is shown to be correlated to short saccades and long fixation durations.

Eye movements are a behavioral output that imply processing of both endogenous and exogenous factors, namely, that have both top-down tuning and bottom-up interactions at different levels of the HVS. Thus, eye movement prediction might require recurrent processing of information from the ventral and dorsal pathways of the HVS, generating a unique representation for eye movement control (visual priority) \cite{Lamme2000}\cite{Corbetta2002}\cite{FECTEAU2006}. If the unique factor to be evaluated is early saliency, stimulus in which features are processed fast and in parallel would be more relevant when evaluating eye movement prediction (showing less inter-participant differences as a consequence of higher SI), namely, the ones with salient regions that are reflectively selected and separated from the background (with higher contrasts with respect the rest of the scene).

\subsection*{Further considerations}

Current literature acknowledges that temporal patterns of saccades have been shown to be fovea-dependent and lately classified as focal and ambient, being ambient fixations responsible for early saccades (sensitive to peripheral signals) and the latter for later saccades (being these ones foveal) \cite{Unema2005}\cite{Pannasch2008}\cite{Follet2011}\cite{Eisenberg2016}. Similarly with saccade latencies, a bimodal latency distribution distinguishes regular from express saccades \cite{Saslow1967}\cite{Schiller1987}\cite{Sommer1997}\cite{vanZoest2006}. We have to acknowledge that the usage of an eye tracker with higher sampling rate (e.g. above \SI{250}{\Hz}) would improve accuracy in this type of experimentation, especially for a possible microsaccadic analysis. Distinct eye movement behavior is presented to be dependent as well for saccade length, pupil dilation and eye vergence \cite{Privitera2014}\cite{Miura2001}\cite{Fallah2012}\cite{Wang2014}\cite{SolPuig2013}. All of these factors should be considered in future visual attention modeling considering their relationships with the two-stream hypothesis \cite{VanEssen1994}\cite{Bell2013}\cite{Trevarthen1968}\cite{Sheth2016} in order to specify the experimental conditions for a better evaluation of uniquely bottom-up visual attention. 

\subsection*{Future work}

Future experimentation for low-level feature analysis in eye movements would be to explore covert attention influences varying some of the presented feature contrasts at distinct eccentricities \cite{Carrasco1998}\cite{Carrasco2006a}\cite{Carrasco2006b}. Another observation of interest would be the evaluation of task differences in localization performance. In that respect, to present the same stimuli with several observations for each feature contrast and distinct cueing would reveal absolute influences from endogenous guidance. Our study could be extended by analyzing the influence of dynamic scenes on saliency modeling \cite{Leboran2017}\cite{Riche2016b} using synthetic videos with both static or dynamic camera. In that direction, it would be able to see the interaction between low-level visual features and temporally-variant features. Another remark would be to see the impact of the target template search in comparison to the odd-one out type of tasks, in this case, but for stimuli with similar display conditions but distinct feature type. 

Physiological evidence could provide an explanation for the low-level feature processing, including both bottom-up and top-down computations reproducing the presented effects not only spatially but temporally. Computations made by the visual cortex that process these low-level features (in reference to the mechanisms that respond distinctively to color, orientation and spatial sensitivities as well as their interactions) might be responsible for most if not all of the effects presented in this study. Further analysis on mid and high-level features would require further study in terms of their relation to psychophysical effects on eye movements as well as their biological foundations \cite{Kruger2013}.


\section{Acknowledgments} \label{acknowledgements}

This  work  was  funded  by  the  Spanish Ministry of Economy and Competitivity (DPI2017-89867-C2-1-R and TIN2015-71130-REDT), Agencia de Gesti\'o d'Ajuts Universitaris i de Recerca (AGAUR) (2017-SGR-649), and CERCA Programme / Generalitat de Catalunya.

\nolinenumbers

{\footnotesize\bibliography{library} \label{references}

\bibliographystyle{apalike}
}
\ifx\multcol\true
\end{multicols}
\fi

\section{Supplementary Material}

\subsection*{Code for the Stimulus Generation} \label{stimulusCode}

\url{https://github.com/dberga/sig4vam}

\subsection*{Dataset Images, Masks and Fixation Data} \label{stimulusFiles}

\url{http://www.cvc.uab.es/neurobit/?page_id=53}

\end{document}